\newif\ifarxiv
\arxivtrue
\documentclass[10pt]{article}
\ifarxiv
\usepackage[preprint]{tmlr}
\else
\usepackage{tmlr}
\fi
\usepackage[american]{babel}
\usepackage{xurl,needspace,longtable}
\usepackage[hidelinks,hypertexnames=false]{hyperref}
\usepackage{natbib,amsmath,amssymb,amsthm,mathtools,graphicx,booktabs,array,tabularx,placeins,float}
\newcolumntype{Y}{>{\raggedleft\arraybackslash}X}
\newcolumntype{Z}{>{\raggedright\arraybackslash}X}
\theoremstyle{definition}
\newtheorem{theorem}{Theorem}
\newtheorem{proposition}[theorem]{Proposition}
\newtheorem{lemma}[theorem]{Lemma}
\newtheorem{corollary}[theorem]{Corollary}
\newtheorem{assumption}{Assumption}
\title{The First-Order Value of Free Covariates in Budgeted Risk Monitoring}
\ifarxiv
\author{\name Yanze Cao \email caoyanze426@gmail.com\\
        \addr Shaanxi University of Technology}
\hypersetup{pdfauthor={Yanze Cao}}
\else
\author{Anonymous authors}
\fi
\hypersetup{hidelinks,hypertexnames=false,pdftitle={The First-Order Value of Free Covariates in Budgeted Risk Monitoring}}
\begin{document}\maketitle
\begin{abstract}
A deployed model sees its inputs for free, but checking whether its
risk has risen requires costly labels. We study how to spend a label
budget so that a risk increase is detected quickly while false alarms
stay controlled, including under input shifts that keep the risk below
a safety threshold $\tau$. Let $A$ be the false-alarm control scale and
$h(r)=\operatorname{kl}(r,\tau)$ the Bernoulli KL divergence between
the changed risk $r$ and $\tau$. At label rate $b=c/\log A$, we prove
that every rule that chooses requests from its own label history has
worst-case (Lorden) delay at least $(1/h(r)+1/c)\log A$ to first
order: besides collecting evidence, it must wait for its next label.
The bound needs only a label budget on one finite stretch of safe
source data and false-alarm control over an initial window; a bound on
the average time to false alarm does not suffice. When a harmful change
also shifts the inputs, a rule that lets the inputs trigger labeling
attains delay $\log A/h(r)$ to first order on a fixed Gaussian model,
so free covariates save at least $1/b$ rounds. A hybrid rule adds a
label-history fallback for risk changes that leave the inputs
unchanged. Synthetic experiments and two ACS tasks, with strengthened
label-only baselines, show the predicted pattern. The
covariate trigger has a cost: because its budget is calibrated on the
source, a benign input shift can make it request nearly every label.
\end{abstract}

\section{Introduction}

An input shift need not harm a deployed model.\footnote{Large language models were used for
literature translation, proof development, experimental programming and
language editing; see Appendix~\ref{app:ai-assistance} for details.} In our ACS Income study,
DC and South Dakota have comparable input mean shifts from California,
yet error falls from 17.1\% to 14.9\% in DC and rises to 27.8\% in SD.
Conversely, error can rise without any input change.
Risk monitoring must tell these cases apart
\citep{PR22,TVNN25}, and the labels it needs may require costly review.

Our null hypothesis allows any input change that keeps the risk below
$\tau$, so an input shift alone can never justify an alarm. Inputs can
still tell us when labels are worth buying, and we use them only for
that: a covariate trigger switches on intensive labeling, and alarms
rest solely on the labels obtained. A second, label-history channel
covers risk increases that leave the inputs unchanged.

A rule that chooses requests from its label history receives no new
feedback between requests. Sampling once every 100 rounds at $b=1\%$,
for example, leaves a change that occurs just after a request waiting
99 rounds for the next label. Under Lorden's worst-case criterion this
gap costs on the order of $1/b$ rounds. We show that no adaptive
label-history rule can avoid it. Free inputs can: a harmful input shift
can trigger labeling inside the gap.

Using accumulated evidence to decide when to acquire data is well
established, from DE-CuSum \citep{BV12} to cost-constrained experiment
design \citep{LB25}. We ask a different question: can any
label-history rule avoid the waiting cost at first order, and can free
covariates? Even with every label, detection needs about
$\log A/h(r)$ rounds. At $b=c/\log A$, we prove that every
label-history rule needs at least $(1/h(r)+1/c)\log A$, while a
covariate-triggered rule attains $\log A/h(r)$ on a fixed Gaussian
harmful-input model. Covariates therefore save at least $1/b$ rounds to first order.
Whether the label-history lower bound is attained remains open.

The critical budget is where waiting and evidence are comparable. For
fixed $h(r)$, the wait $1/b$ is of lower order than $\log A$ when
$b\log A\to\infty$ and dominates when $b\log A\to0$; at
$b=c/\log A$ both enter the leading coefficient.

Our contributions are:
\begin{enumerate}
\item \textbf{A first-order value of covariates.} A converse uniform
over label-history rules (Theorem~\ref{thm:critical-budget}) and a
covariate-triggered construction (Theorem~\ref{thm:upper}) separate the
two acquisition classes at the critical budget.
\item \textbf{The right error condition.} False-alarm control over an
initial window and a budget on one source prefix suffice for the
converse. A counterexample shows that bounding only the average time to
false alarm does not (Section~\ref{sec:lower}).
\item \textbf{Hybrid acquisition.} A parallel label-history fallback
retains detection when the inputs do not change, at the price of a
split budget (Theorem~\ref{thm:upper}(b)). A growing betting grid
handles an unknown post-change risk, and we characterize when the risk
is identifiable under restricted auditing.
\item \textbf{Experiments.} On synthetic data and two ACS tasks,
label-only delay grows as the budget shrinks while the covariate
trigger's delay stays nearly flat. After benign input shifts, however,
the covariate channel requests almost every label
(Section~\ref{sec:experiments}).
\end{enumerate}

\section{Model and risk validity}\label{sec:model}

At each round $t$ the monitor sees the input $X_t$ for free and may pay
to observe the binary loss $L_t$ of the deployed model. Write
$R_t\in\{0,1\}$ for the request indicator. The observed filtration is
\[
\mathcal F_t=\sigma(\mathcal F_0,X_s,R_s,R_sL_s:s\le t).
\]
It includes independent randomization drawn at initialization, but not
audit coins for future rounds. The risk null is
\begin{equation}
\mathcal H_0=\{Q:E_Q[L_t\mid\mathcal F_{t-1}]\le\tau\ \forall t\},
\label{eq:M1}
\end{equation}
where $0<\tau<1$. The null constrains only the conditional mean loss.
Input distributions may change arbitrarily, losses may depend on $X_t$
in any way, and a safe process may even react to past requests.

A general policy chooses $p_t(\cdot)$ from past information, observes
$X_t$, and then draws $R_t$ with probability $p_t(X_t)$, independently
of $L_t$ given $(\mathcal F_{t-1},X_t)$; Appendix~\ref{app:input-kernels}
gives the corresponding augmented losses. Our constructive policies
are simpler. They decide whether to request before seeing $X_t$, so
their on/off state is determined by $\mathcal F_{t-1}$, and they bet on
requested losses with the factor
\begin{equation}
G_t(\lambda)=1+\lambda R_t(L_t-\tau),\quad 0<\lambda<1/\tau.
\label{eq:M2}
\end{equation}
Under every null this factor is nonnegative with conditional mean at
most one; unrequested rounds contribute a factor of one.

Fix $0<b<1$ and an IID nominal source $P_0$. The budget counts the
source requests of the \emph{unstopped} policy, including requests
after an alarm:
\begin{equation}
E_0N_K\le bK+c_0,\quad N_K=\sum_{t=1}^KR_t,\quad K\ge1.
\label{eq:M3}
\end{equation}
\begin{assumption}[Nominal safety]\label{ass:nominal}
$r_{\rm src}:=E_{P_0}L\le\tau$, so $P_0\in\mathcal H_0$.
\end{assumption}
The general converses require an allowance $c_0\le C_0$ common to the
policy family, including rules whose design depends on the threshold.
The critical-budget converse needs the budget at only one prefix length
in a common range. In both cases the budget is measured on safe source
data and does not limit labeling after a change;
Section~\ref{sec:experiments} shows that this cost can be large.

For delay, full-data rounds are IID $P_0$ before $\nu$ and IID $P$
afterward, independently of the past, with $\tau<r:=E_PL<1$.
Inputs and losses in the same round may be dependent. Lorden delay is
\begin{equation}
D_P(T)=\sup_{\nu\ge1}\operatorname*{ess\,sup}
 E_{\nu,P}[(T-\nu+1)^+\mid\mathcal F_{\nu-1}].
\label{eq:M4}
\end{equation}
The general converses use finite-window false-alarm control:
\begin{equation}
Q(T_A\le m)\le m/A\quad(Q\in\mathcal H_0,\ m\ge1).
\tag{FW}\label{eq:FW}
\end{equation}
Unlike a bound on the average run length, \eqref{eq:FW} bounds the
probability of an alarm by each finite horizon. Theorem~\ref{thm:critical-budget}(b)
needs it only on an initial window, and Section~\ref{sec:lower} shows
that a bound on the average run length alone does not suffice. Put
$h(r)=\operatorname{kl}(r,\tau)$ and
$\mathcal E(b,A)=1+\log(1/b)+\log\log A$, $A\ge e^2$.

Our detectors are SR and CUSUM e-detectors \citep{SRR23}. Each
component starts at 0 and follows
$V_t=G_t(1+V_{t-1})$ or $U_t=G_t\max(1,U_{t-1})$.
An alarm when a fixed convex average of components reaches $A$
satisfies both (FW) and $E_QT_A\ge A$ for every $Q\in\mathcal H_0$
(Theorem~\ref{thm:validity}, Appendix~\ref{app:validity}).
The lower bounds, however, apply to any stopping rule that meets their
error conditions.

\section{Acquisition policies}
\label{sec:controllers}

We analyze two acquisition channels and their combination. C uses the
inputs to decide when to label; L uses only its own label history.

\paragraph{Covariate blocks C.}
C watches the input mean block by block. For the constructive theory,
source inputs are $N(0,I_d)$ and changed inputs are $N(\delta u,I_d)$,
with fixed $d$, $\delta>0$ and unit $u$. On block $j$ of length $w$,
C requests every label if the previous block looked shifted:
\[
H_j^C=1\{w\|\overline X_{j-1}\|^2\ge c_b\},\qquad H_1^C=0.
\]
The threshold $c_b$ is the central chi-square quantile with source
trigger probability $b$, so $E_0N_K^C\le bK$. A block length
$w=O(\mathcal E(b,A)/\delta^2)$ gives trigger power
$\pi_A\ge1-1/\log A$ on changed blocks. The requested labels feed the
risk detector.

\paragraph{Label-history L.}
L is a sampling--sleep controller in the spirit of DE-CuSum
\citep{BV12}. Fix a design risk $r_0$ with $r_{\rm src}\le r_0<\tau$.
While active, L requests every label and adds
\[
Y(L)=L\log(\tau/r_0)+(1-L)\log((1-\tau)/(1-r_0))
\]
to its score. When the score falls below 0, L resets it to 0 and sleeps
for $W=\lceil e_0(1-b_L)/b_L\rceil$ rounds. L starts active at score 0,
has no positive cap, and uses only its own labels. With source
excursion mean $e_0<\infty$, $E_0N_K^L\le b_LK+e_0$. After a harmful
change, an excursion escapes into permanent sampling with probability
$p_{\rm esc}(r)\ge\tau-r_0>0$ (Lemma~\ref{lem:controller}).

\paragraph{Hybrid.}
Hybrid runs C and L in parallel with budgets $b_C+b_L=b$ and requests a
label when either channel asks, $R_t^H=H_t^C\vee H_t^L$. A shared
request costs one label, and every requested loss feeds one common
alarm statistic. L's score uses only L's own requests, and C's requests
do not interrupt L's sleep, so $E_0N_K^H\le bK+e_0$. Both channels
decide before seeing $X_t$ (the current input affects only a later
block state), so the factor \eqref{eq:M2} remains valid.

\section{Lower bounds}
\label{sec:lower}\label{sec:critical-budget}

We first identify the evidence cost. Reweighting $P$'s labels to
probabilities $\tau,1-\tau$ gives a safe $Q^*$ with
$D(P\Vert Q^*)=h(r)$, and no adaptive observation kernel can increase
the one-round KL. Consequently, (FW) implies the full-information
first-order lower bound $1/h(r)$ (Theorem~\ref{thm:information}).
Here the composite null matters: with a single specified null, an input
change alone would identify the alternative. For hard audit regions
$B$, the risk is identifiable if and only if $E_P[L1_B(X)]>\tau$
(Proposition~\ref{prop:identification}).

The valid class is nonempty even with $c_0=0$: requesting every
$\lceil1/b\rceil$ rounds meets the budget, and a one-factor CUSUM
using \eqref{eq:M2} satisfies (FW).

\Needspace{8\baselineskip}
\begin{theorem}[Additive waiting and evidence lower bounds]\label{thm:additive}

Assume \eqref{eq:M3} with a common \(c_0\le C_0\), Assumption~\ref{ass:nominal}, fixed model parameters, \(b\downarrow0\), and \(A\ge b^{-3}\). For any stopping rule satisfying \eqref{eq:FW}:

\textbf{(a) Label-only acquisition.} If requests use only independent initialization, past requests, and previously requested losses, with no online covariate information, then
\begin{equation}
D_P(T_A)\ge\frac{\log A}{h(r)}
+\frac{(1-b)^2}{2b}
-O(\mathcal E(b,A)).\label{eq:M5}
\end{equation}

The detector may still observe covariates.

\textbf{(b) Pure label change, including covariate-using acquisition.} If the full-data model is IID on either side and \(P_X=P_{0,X}\), the same bound holds for every admissible acquisition kernel, including kernels using current and past covariates.

\end{theorem}

\begin{figure}[t]
\centering
\includegraphics[width=0.60\linewidth]{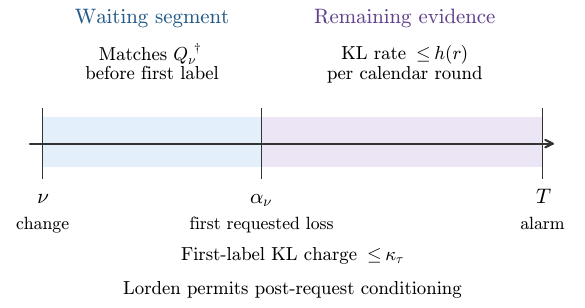}
\caption{One path, two successive costs. Before the first changed
label, the observations agree with a safe reference. The first label
costs at most $\kappa_\tau$ in stopped KL, and each later calendar
round at most $h(r)$. Lorden's criterion allows conditioning on the
history just after a request.}
\label{fig:waiting-timeline}
\end{figure}

\paragraph{Why the costs add.}
Let $\kappa_\tau=\max\{\log(1/\tau),\log(1/(1-\tau))\}$.
Let $\alpha_\nu$ be the first request at or after the change, and split
$(T-\nu)^+$ into the wait until $\min(T,\alpha_\nu)$ and the remainder
after $\alpha_\nu$. A safe reference process $Q_\nu^\dagger$ (defined
below) keeps the changed input law until that request, so validity
controls premature alarms during the wait. The reference then switches
to $Q^*$, which costs at most $\kappa_\tau$ in KL for the first label
and $h(r)$ for each later expected calendar round. A binary test at one
common horizon gives the evidence requirement. Because waiting and
evidence are averaged over the same changepoints in the same joint
experiment, their bounds add. Appendix~\ref{app:additive} gives the
finite inequality and the stopped-KL construction; the argument needs
no concentration of the observed information.

\paragraph{The finite inequality behind Theorem~\ref{thm:additive}.}
The additivity can be checked before taking limits. For integers
$K,M\ge1$ and $K<m<A$, define $\alpha_\nu=\inf\{t\ge\nu:R_t=1\}$,
\[
Y_\nu=(\min\{T,\alpha_\nu\}-\nu)^+,
\qquad Z_\nu=(T-\alpha_\nu)^+,
\]
with $Z_\nu=0$ when $\alpha_\nu=\infty$. The identity
$Y_\nu+Z_\nu=(T-\nu)^+$ also covers alarms before the change.
Set $n_* =\min\{K,bK+c_0\}$ and
\[
\begin{aligned}
\varphi_M(x)&=\tfrac12\{x^2-(x-M)_+^2\},\quad x\ge0,\\
\Phi_{K,M}(n)&=(n+1)\varphi_M\!\left(\frac{K-n}{n+1}\right),\\
B_{K,M}&=\left[\frac{\Phi_{K,M}(n_*)}{K}
              -\frac{M(K+M)}{2A}\right]_+.
\end{aligned}
\]
An audit path with $N_K$ requests splits its $K-N_K$ unaudited rounds
into $N_K+1$ gaps. The truncated forward waits in these gaps sum to at
least $\Phi_{K,M}(N_K)$. Convexity, monotonicity and the expected source
budget then give $K^{-1}\sum_{\nu\le K}E_\nu Y_\nu\ge B_{K,M}$ after
the (FW) correction for premature alarms.

For evidence, $Q_\nu^\dagger$ uses $P_0$ before $\nu$,
$P_X\otimes\operatorname{Bernoulli}(\tau)$ through the first changed
request, and $Q^*$ thereafter. Both changed laws have risk $\tau$, and
the switch depends only on past requests, so the process stays in
$\mathcal H_0$. For $a(x)=P(L=1\mid X=x)$, the KL of the first label obeys
\[
\begin{aligned}
\int p_t(x)\operatorname{kl}(a(x),\tau)P_X(dx)
 &\le\kappa_\tau\int p_t(x)P_X(dx).
\end{aligned}
\]
The stopped compensated count of a first request is at most one.
After that request, data processing bounds each calendar-round KL by
$h(r)$. Thus for $U=T\wedge m$, the chain rule for the bounded stopped
trajectory gives
\[
\operatorname{KL}(P_\nu|_{\mathcal F_U}\Vert
 Q_\nu^\dagger|_{\mathcal F_U})
 \le\kappa_\tau+h(r)E_\nu Z_\nu.
\]
Charging the first label separately handles losses that vary with $X$,
and stopping at a bounded time avoids optional-stopping arguments for
unbounded times.

Give the analyst an index $J$ uniform on $\{1,\ldots,K\}$, hidden from
the procedure, and use the corresponding joint alternative and safe
reference. Put $\mu=E(Y_J+Z_J)$, $d=m-K$, and $\ell=\log(A/m)$.
In this joint experiment the reference alarm probability is at most
$m/A$ and the stopped KL is at least $(1-\mu/d)\ell-\log2$.
Combining the two bounds gives
\begin{equation}
D_P(T_A)\ge
\left[\frac{B_{K,M}+(\ell-\kappa_\tau-\log2)/h(r)}
{1+\ell/(h(r)d)}\right]_+.
\label{eq:main-finite-interface}
\end{equation}
Appendix~\ref{app:additive} proves this inequality. Choosing
$M=\lceil b^{-1}\rceil$,
$K=\lceil(C_0+1)b^{-2}\rceil$, and
$d=\lceil(\log A+b^{-1}+1)^2\rceil$ recovers
Theorem~\ref{thm:additive}. Because the allowance $C_0$ is common to
all rules, these choices and the error terms are uniform over
acquisition rules.

\paragraph{Calendar time versus request epochs.}
Periodic sampling shows where the factor $1/2$ in
Theorem~\ref{thm:additive} comes from. With one request every $a$
rounds, a change at a uniformly chosen calendar time waits $(a-1)/2$
rounds on average, whereas a change just after a request waits $a-1$
rounds. Lorden's criterion allows conditioning on the history just
after a request, which recovers the full gap. For arbitrary adaptive
rules, the proof replaces the periodic picture with an exact path
identity and a split by source request mass, without renewal
assumptions.

\Needspace{8\baselineskip}
\begin{theorem}[Request-epoch converse and local critical-budget separation]
\label{thm:critical-budget}
Retain nominal safety, either acquisition class of
Theorem~\ref{thm:additive}, fixed models, and $\tau<r<1$.

\textbf{(a) General joint range.}
Under \eqref{eq:M3} with a common $c_0\le C_0$ and \eqref{eq:FW},
as $b\downarrow0$, uniformly for $A\ge b^{-5}$,
\begin{equation}
 D_P(T_A)\ge\frac{\log A}{h(r)}+\frac1b
 -O\!\left(\mathcal E(b,A)\right).
 \label{eq:epoch-general}
\end{equation}

\textbf{(b) Local critical budget.}
Fix $c>0$, $C_0\ge0$, and $\Gamma>(C_0+1)/c^2$. Put
$q=\log A$, $b=c/q$, and define
\[
\begin{aligned}
 d_A&=\left\lceil(q+b^{-1}+1)^2\right\rceil,\\
 H&=(2+c^{-1})^2+1,\quad
 W_A=\left\lceil(\Gamma+H)q^2\right\rceil.
\end{aligned}
\]
Suppose there exists a deterministic integer $K_A$, possibly depending
on the rule, satisfying the local prefix budget:
\begin{equation}
 \left\lceil(C_0+1)b^{-2}\right\rceil\le K_A\le
 \lfloor\Gamma q^2\rfloor,\quad E_0N_{K_A}\le bK_A+C_0.
 \tag{LP}\label{eq:local-prefix}
\end{equation}
Assume only local finite-window false-alarm control:
\begin{equation}
 Q(T_A\le s)\le s/A
 \quad(Q\in\mathcal H_0,\ 1\le s\le W_A).
 \tag{LFW}\label{eq:local-fw}
\end{equation}
For all sufficiently large $A$,
\begin{equation}
 D_P(T_A)\ge q\left\{\frac1{h(r)}+\frac1c\right\}
 -\frac3{h(r)}\log q-C.
 \label{eq:critical-finite}
\end{equation}
The constant and sufficient threshold depend only on the fixed model,
$\tau,c,C_0,\Gamma$, uniformly over the rules and their choices of $K_A$.
Thus, for the class $\Pi_A^{\mathrm{loc}}(C_0,\Gamma)$ just specified,
\[
 \liminf_{A\to\infty}\inf_{\pi\in\Pi_A^{\mathrm{loc}}(C_0,\Gamma)}
 \frac{D_P(T_\pi)}{\log A}\ge\frac1{h(r)}+\frac1c.
\]
\end{theorem}

Part (a) raises the waiting coefficient of Theorem~\ref{thm:additive}
from $1/2$ to $1$ over a joint range of $(b,A)$. Part (b) obtains the
same coefficient at the critical budget from conditions that involve
only one source prefix and an initial window.

\paragraph{Why request epochs give the unit coefficient.}
All expectations below are taken in the same joint experiment. Write $K=K_A$ and $m=K+d_A\le W_A$.
Add a virtual request at time 0 and define
\[
\begin{aligned}\iota_1&=1,\quad \iota_\nu=R_{\nu-1}\ (2\le\nu\le K),\\
n&=1+E_0N_{K-1}.\end{aligned}
\]
For every unstopped request path, exactly
\begin{equation}
\begin{split}
\sum_{\nu=1}^K\iota_\nu
 (\min\{\alpha_\nu,K\}-\nu)^+
 &=K-(1+N_{K-1}).
\end{split}
\label{eq:main-epoch-identity}
\end{equation}
Each interval is measured from a request rather than from a uniformly
chosen calendar time, so its whole unaudited length is counted once,
including the final clipped interval. No renewal or stationarity
assumption is involved.

If $x=E_0N_K>bK/2$, use
\[
\begin{aligned}
\bar P(J=\nu,d\omega)&=\frac{\iota_\nu(\omega)}nP_\nu(d\omega),\\
\bar Q(J=\nu,d\omega)&=\frac{\iota_\nu(\omega)}nQ_\nu^\dagger(d\omega).
\end{aligned}
\]
Because $\iota_\nu$ is determined by the past and both laws agree
before $\nu$, both measures have mass one, and conditioning on the
common selector leaves the likelihood ratio unchanged. With
$y=E_{\bar P}Y_J$, $z=E_{\bar P}Z_J$, Lorden's criterion yields
$y+z\le D_P(T)$, while the stopped KL remains at most
$\kappa_\tau+h(r)z$.

Put $\delta=W_A/A$. Local validity is applied to each unconditioned
reference $Q_\nu^\dagger$, never to a conditioned one. Since
$n\ge x>bK/2$,
\[
\bar Q(T\le m)\le\frac{K\delta}{n}\le\frac{2\delta}{b}.
\]
The path identity and the correction for premature alarms give
\[
y\ge\frac Kn-1-O(K\delta/b)\ge\frac1b-O(1).
\]
Indeed, $n\le1+bK+C_0$ and
$K\ge(C_0+1)b^{-2}$ bound the deterministic loss, while
$K\delta/b=O(q^5/A)$. No individual request probability needs a
positive lower bound.

If instead $x\le bK/2$, use the uniform-calendar experiment with
$M=\lceil2/b\rceil$. The gap bound now uses only the mean at this one
$K$, not a new all-prefix budget at rate $b/2$, and gives
$y\ge1/b-O(1)$. In both cases we have $y+z\le D$, a waiting term of
$1/b$ and the same stopped-KL bound. The binary test uses
\[
\log\frac b{2\delta}=q-3\log q-O(1),
\]
and its denominator correction costs only $O(1)$ because
$d_A=\Theta(q^2)$. This yields the finite remainder in
Theorem~\ref{thm:critical-budget}(b), uniformly over policies and their
prefix lengths $K_A$. Appendices~\ref{app:local-finite}--\ref{app:local-limits}
give the full details, including measurability of the events involved.

\paragraph{Checking the local conditions.}
The prefix length $K_A$ is fixed in advance, not chosen from the data,
and the budget is required only at that prefix. Its range keeps all
proof horizons inside $W_A$, so the constants are uniform over rules.
For the factors in \eqref{eq:M2}, Theorem~\ref{thm:validity} gives
(FW), and hence (LFW), for the SR/CUSUM mixtures. An all-prefix source
budget with allowance at most $C_0$ implies (LP) for large $A$. Both
conditions are therefore verified analytically from the source law; an
audit fraction observed on a finite sample does not verify them.

\paragraph{Budget and error calibration.}
The window can also use a separately specified error level. Write
$\alpha_A^*=\sup_{Q\in\mathcal H_0}Q(T_A\le W_A)$.
\begin{corollary}[Budget and local calibration]
\label{cor:local-calibration}
Retain the models and \eqref{eq:local-prefix} of
Theorem~\ref{thm:critical-budget}(b), without assuming
\eqref{eq:local-fw}. Let a deterministic calibration bound
$\delta_A\ge\alpha_A^*$ satisfy
\[
 \begin{aligned}
 0<\delta_A&<b/2,\quad
 \ell_A:=\log(1/\delta_A)\le L_0q,\\
 q^3\delta_A&\le E_{\mathrm{loc}},
 \end{aligned}
\]
where $L_0,E_{\mathrm{loc}}>0$ are common fixed constants. Then, uniformly,
\begin{equation}
 D_P(T_A)\ge
 \frac{\log(1/\delta_A)}{h(r)}+\frac1b
 -\frac{\log(1/b)}{h(r)}-C.
 \label{eq:decoupled-local}
\end{equation}
The common constant may also depend on $L_0,E_{\mathrm{loc}}$.
\end{corollary}

For example, if $\alpha_A^*\le q^{-3}$, take $\delta_A=q^{-3}$
and $E_{\mathrm{loc}}=1$. The other conditions hold eventually, and
\eqref{eq:decoupled-local} becomes
\[
 D_P(T_A)\ge \frac1b+\frac{2\log q}{h(r)}-C.
\]
The wait of order $1/b$ therefore survives calibration far weaker than
$W_A/A$. The two leading costs have different sources: $1/b$ comes from
acquisition and $\log(1/\delta_A)/h(r)$ from error control. They still
interact through the logarithmic conditioning loss and the finite-window
conditions, which limit how large the evidence term can be relative to
the horizon (Appendix~\ref{app:arl-boundary}).

\paragraph{Why ARL alone fails.}
Draw $I\sim\operatorname{Bernoulli}(b)$
independently in $\mathcal F_0$, with the same law under every null.
Keep $R_t=I$ for all $t$, including after alarm, and set
\[
 T_A=\begin{cases}
 1,&I=0,\\
 \text{full-audit SR crossing at } A/b,&I=1.
 \end{cases}
\]
On the active branch, use the likelihood factor for Bernoulli risk $r$
versus $\tau$. Then, for every source prefix and every full-risk null,
respectively,
\[
 E_0N_K=bK,\qquad E_QT_A\ge 1-b+b(A/b)\ge A.
\]
Yet the overshoot bound for the product walk and SR domination give
\[
 D_P(T_A)\le\frac{\log(A/b)+\log(r/\tau)}{h(r)}
 =\frac q{h(r)}+O(\log q).
\]
The inactive branch alarms immediately, so a path that survives round 1
is on the full-sampling branch, and the unconditional budget says
nothing about that branch. Here $\alpha_A^*\ge1-b\to1$, so the
calibration conditions of Corollary~\ref{cor:local-calibration} fail.
Appendix~\ref{app:arl-boundary} proves full-null validity and the
delay bound over all histories. Local error control is thus what
preserves the waiting term in this converse.

\section{Upper bounds and covariate value}
\label{sec:upper}

For an audited Bernoulli loss, let
\[
\psi(\lambda;r)=r\log(1+\lambda(1-\tau))
 +(1-r)\log(1-\lambda\tau).
\]
Writing $\rho=\tau+\lambda\tau(1-\tau)$ gives
\begin{equation}
\psi(\lambda;r)=h(r)-\operatorname{kl}(r,\rho).
\label{eq:M8}
\end{equation}
This is the Bernoulli form of the standard log-growth identity
\citep{Kelly56,GdHK23}: the growth lost relative to the Kelly bet is
the KL divergence to the risk being bet on. A geometric grid of mesh
$1/\log A$ has $J_A=O(\log A)$ points and approximates Kelly growth
uniformly on a fixed $\mathcal R=[\tau+\Delta,\bar r]$, with
$\Delta>0$, $\bar r<1$ (Lemma~\ref{lem:grid}).

\begin{theorem}[C and Hybrid with unknown post-change risk]\label{thm:upper}

Use the growing grid of Lemma~\ref{lem:grid}, fixed before monitoring at each \(A\ge e^2\). Use deterministic nonnegative weights summing to 1, with every weight satisfying \(w_{j,A}\ge c_w/J_A\) for a fixed \(c_w>0\); uniform weights give \(c_w=1\). Keep \(\mathcal R\) fixed. In the covariate model also keep \(d,\delta>0\) and the other design parameters fixed.

\textbf{(a) Covariate harmful shift.} With the high-power block construction,
\begin{equation}
\begin{aligned}D_C(T_A)&\le\frac{\log A}{h(r)}+O(\mathcal E(b,A)),\\
\quad
D_H(T_A)&\le\frac{\log A}{h(r)}+O(\mathcal E(b_C,A)).\end{aligned}\label{eq:M9}
\end{equation}

\textbf{(b) Pure label change.} For Hybrid,
\begin{equation}
D_H(T_A)\le\frac{\log A}{h(r)}
+\frac{e_0}{p_{\rm esc}(r)b_L}
+O(1+\log\log A).\label{eq:M10}
\end{equation}

The remainders can be chosen uniformly for \(r\in\mathcal R\). The ARL guarantee and nominal cumulative budget are unchanged.

\end{theorem}

Lemmas~\ref{lem:controller}--\ref{lem:grid} supply the renewal,
audit-clock and grid bounds used here
(Appendices~\ref{app:controller}--\ref{app:grid}).
The label channel adds at most $e_0/(p_{\rm esc}b_L)$ rounds, while a
high-power covariate trigger reduces the acquisition overhead to
logarithmic order. If $\log(1/b)=o(\log A)$, C and fixed-fraction
Hybrid attain $D/\log A\to1/h(r)$ on the fixed Gaussian harmful-input
model.

\paragraph{First-order value.}
For the Gaussian harmful-shift law $P$ of
Theorem~\ref{thm:upper}(a), let $\Pi_A^{\rm lab}$ be the label-history
class of Theorem~\ref{thm:critical-budget}(b), with common $C_0,\Gamma$.
C meets stronger validity and budget conditions than those required of
this class, and the two theorems give
\begin{equation}
\begin{split}
\inf_{\pi\in\Pi_A^{\rm lab}}D_P(T_\pi)-D_P(T_A^C)
 &\ge \frac1b-O(\log\log A),\\
 b&=c/\log A.
\end{split}
\label{eq:covariate-value}
\end{equation}
Covariates thus remove a first-order cost of at least $1/b$ rounds.
When inputs do not change and are IID, the waiting cost applies to
every admissible acquisition kernel instead.

Table~\ref{tab:1} collects the resulting delay scales.

\begin{table}[ht]
\centering
\small
\caption{Delay scales as $b\downarrow0$ with $A\ge b^{-3}$, $\log\log A=o(1/b)$ and fixed positive channel fractions (Corollary~\ref{cor:comparison}). The unit waiting coefficient in Theorem~\ref{thm:critical-budget}(a) uses $A\ge b^{-5}$. The covariate column uses the fixed Gaussian harmful shift.}
\label{tab:1}
\setlength{\tabcolsep}{4pt}
\renewcommand{\arraystretch}{1.15}
\begin{tabular}{@{}lll@{}}
\toprule
\textbf{Policy} & \textbf{Covariate shift} & \textbf{Pure label change} \\ \midrule
L & $\log A/h+\Theta(1/b)$ & $\log A/h+\Theta(1/b)$ \\
C & $\log A/h+O(\mathcal E)$ & No label fallback \\
Hybrid & $\log A/h+O(\mathcal E)$ & $\log A/h+O(1/b_L)$ \\
\bottomrule
\end{tabular}
\end{table}

\FloatBarrier

\section{Experiments}
\label{sec:experiments}

We test two predictions. Under harmful input shifts, covariate-triggered
acquisition should stay fast as labels become scarcer, while
label-history acquisition slows down. When the inputs do not change, a
label-history fallback should be needed. ACS cells use 500 paired paths
at $A=10^6$ and a 200,000-round post-change horizon. We report
restricted mean delays; censoring, paired descriptive intervals and full
protocols are in Appendices~\ref{app:implementation-limits}--\ref{app:traveltime}.

\begin{figure}[H]
\centering
\includegraphics[width=0.56\linewidth]{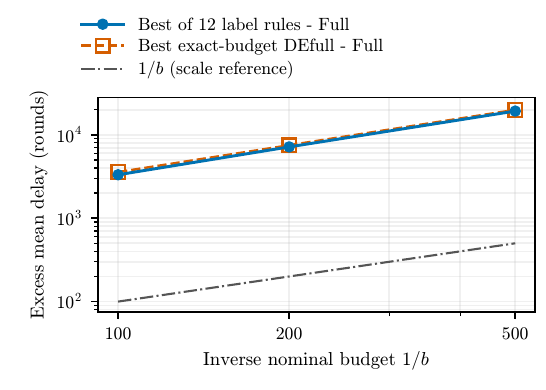}
\caption{Excess mean delay over full audit in the synthetic SR study at $\nu=1$. The $1/b$ line from Theorem~\ref{thm:critical-budget} is shown for scale; $b$ times the excess is 33--39.}
\label{fig:1}
\end{figure}

\paragraph{Synthetic budget dependence.}
With source risk 0.07, $\tau=0.10$ and changed risk 0.14, four fixed
directions supply 512 paths per budget. At $A=10^7$, C stays near
1,300 rounds as $b$ falls, while the fastest of 12 evaluated label
controllers rises from 4,692 to 20,601 rounds. An exact-budget DEfull
subfamily shows the same pattern (Figure~\ref{fig:1}).

\paragraph{Two ACS tasks.}
We resample uniformly from fixed, unweighted 2018 folktables pools
\citep{DHMS21}. Each task has one fixed LightGBM model trained on
California data; separate calibration sets fix $\tau$ and the source
input geometry, and targets are selected before any detection run.
Income has $\tau=0.2012$ and source evaluation risk 0.1715; TravelTime
has $\tau=0.3444$ and source risk 0.3095. The TravelTime model has a
7.82 percentage-point gap between training and holdout risk; the
interstate evaluation applies to this fixed CA-trained model. The nominal
budgets are 1\%, 0.5\% and 0.2\%.

Across both tasks, even the \emph{post hoc} fastest of three
DE-inspired sampling--sleep controls has restricted mean delay
3.4--19.5 times C's, and the gap grows as the budget shrinks
(Table~\ref{tab:cross-task}), as the inverse-budget term in
Theorem~\ref{thm:critical-budget} predicts. Two choices favor these
baselines: they are calibrated with the exact empirical source risk,
and we report the best of three designs after seeing the results.
Uniform acquisition uses the same eight-component direct-loss CUSUM as
C. Appendices~\ref{app:shared-baselines} and~\ref{app:traveltime} give
fixed-design intervals, Uniform survival curves, censoring, target
selection, calibration and controller details.

\begin{table}[H]
\centering\small
\caption{Cross-task results at $A=10^6$. Natural-shift ranges cover the three harmful targets of each task. Best DE/C uses the post hoc best of three prespecified DE designs, so no interval is given for the selected design. Flip cells give restricted mean rounds with censored paths out of 500 in parentheses. H$_{\rm task}$ is the capped H$_D$ supplement for Income and the uncapped Hybrid for TravelTime; L is each task\textquotesingle s primary label controller. At Income 1\%, H$_D$ can gather label evidence before C activates, giving ratios slightly below one; only SD's pointwise paired 95\% interval excludes 1 (Appendix~\ref{app:hd-small-gain}).}
\label{tab:cross-task}
\setlength{\tabcolsep}{3pt}
\renewcommand{\arraystretch}{1.12}
\begin{tabularx}{\textwidth}{@{}llYYYYYY@{}}
\toprule
 & & \multicolumn{3}{c}{Natural harmful input shifts} & \multicolumn{3}{c}{Input-preserving loss flips} \\
\cmidrule(lr){3-5}\cmidrule(l){6-8}
\textbf{Task} & \textbf{$b$} & \textbf{C delay} & \textbf{Best DE/C} & \textbf{H$_{\rm task}$/C} & \textbf{L} & \textbf{C} & \textbf{H$_{\rm task}$} \\ \midrule
Income & 1\% & 1,063--2,030 & 3.44--3.57 & 0.995--0.999 & 4,906 (0) & 136,060 (153) & 9,614 (0) \\
Income & 0.5\% & 1,131--2,118 & 5.65--5.88 & 1.013--1.024 & 8,417 (0) & 171,413 (316) & 17,597 (0) \\
Income & 0.2\% & 1,245--2,230 & 12.15--12.92 & 1.006--1.012 & 18,957 (0) & 193,622 (458) & 41,934 (4) \\
TravelTime & 1\% & 838--1,029 & 4.43--4.61 & 1.002--1.003 & 11,052 (0) & 191,466 (411) & 18,634 (0) \\
TravelTime & 0.5\% & 847--1,042 & 7.49--7.61 & 1.001--1.002 & 18,211 (0) & 199,508 (493) & 36,372 (0) \\
TravelTime & 0.2\% & 858--1,051 & 18.35--19.45 & 1.001--1.002 & 42,502 (2) & 200,000 (500) & 70,236 (49) \\
\bottomrule
\end{tabularx}
\end{table}

Input-preserving loss flips expose C's weakness. At TravelTime budget
0.2\%, C has 500/500 censored paths, versus 2/500 for L and 49/500 for
Hybrid. Here $h\approx0.00542$, so the information scale at the nominal
rate, $\log A/(b h)\approx1.27\times10^6$ rounds, already exceeds the
200,000-round horizon. The label-history fallback restores detection on
most paths.

Benign shifts are expensive for the covariate channel. In DC, C
requests 99.79\%--99.87\% of labels; in AK, C and Hybrid request about
99.98\%. Because the budget is calibrated on the source, nothing limits
labeling after a benign change. On Income, a confirmation-reset variant
reduces DC labeling once it has adapted (Appendix~\ref{app:benign-reset});
alarm counts and the TravelTime acquisition profiles are in
Appendix~\ref{app:traveltime}.

\section{Related work}
\label{sec:related}

Risk monitoring \citep{PR22,TVNN25} separates performance changes from
input shifts. \citet{TVNN25} observe every loss, whereas we budget
adaptive label acquisition. Label-free methods based on proxies
\citep{AB24,schirmer2025tta} and monitoring with synthetic labels
\citep{ZCYS26} address complementary observation models.
MLDemon \citep{GZZ22} combines feature signals with on-demand labels,
request-and-reverify \citep{YWP18} screens unlabeled data before
verifying with labels, and Drift2Act \citep{LYEOM26} lets drift
evidence set the size of label audits. These works establish the
acquisition idea. MLDemon analyzes an amortized monitoring loss under
gradual drift; Drift2Act uses window-risk upper certificates to gate
interventions. Our target is valid stopping and Lorden delay under a source-prefix budget, together with a
lower bound over all label-history rules.
Active prediction-powered inference \citep{XKM24,SW26} motivates our
augmented factors; SR/CUSUM e-detectors \citep{SRR23} and
growth-optimal testing \citep{GdHK23,LRR25,LRR26} provide the validity
tools.

Bayesian sleep/wake scheduling \citep{PK08} optimizes observation and
detection costs. DE-CuSum \citep{BV12,BV14}, observation scheduling
\citep{RJS17} and cost-constrained experiment design \citep{LB25,LB26}
already control acquisition by evidence and establish efficiency under
a fixed budget. Sampling-constrained identification
\citep{tsopelakos2023anomaly} has a different terminal objective and
resource constraint. Appendix~\ref{app:related-details} compares their observation and
resource models with ours.

Stopped-trajectory KL decompositions are established in bandit change
detection \citep{GSL21}. Multi-stream sampling control already has
resource-dependent overheads \citep{XMM21,XM23}, and information-clock
converses \citep{LTV22} require growth and tail conditions that an
expected budget does not supply uniformly. Our technical step combines a
budget-driven waiting comparison with a safe random-start reference in
one experiment. Transient sparse detection \citep{CT17} has related
sampling obstructions under different timing and resource criteria.

\section{Discussion and limitations}\label{sec:discussion}

Free covariates can remove a first-order cost of $1/b$ rounds even
though they cannot certify risk on their own. This rests on a converse
uniform over label-history rules and a construction on the same fixed
Gaussian model; when inputs do not change, the waiting cost remains.

\paragraph{The label-history coefficient.}
The unit waiting coefficient concerns the first request. With fresh
Bernoulli($b$) audit coins, the mean wait for that request is $1/b$
including the request round, or $(1-b)/b$ excluding it, so the waiting
coefficient itself is tight. Attaining the total lower bound would also
require collecting evidence efficiently under the source budget, and
whether a label-history rule can do so is open. Figure~\ref{fig:1}
measures the mean overhead of specific controllers at a finite
threshold, not this optimum.

\paragraph{Using the two channels.}
C suits settings where harmful changes are expected to move a
source-calibrated input representation. Hybrid adds a fallback for
risk changes that leave inputs unchanged. With $b_L=\theta b$ and
$b_C=(1-\theta)b$, fixed $0<\theta<1$, its fallback term scales as
$e_0/(p_{\rm esc}\theta b)$, so a larger share for L improves that
bound at C's expense; we evaluate equal splitting. Post-shift label use
should be reported alongside delay, since benign shifts can exhaust a
practical labeling budget even when the source calibration holds.
Validity and budget should be verified analytically rather than
inferred from observed alarm counts.

\paragraph{Limitations.}
(1) The optimal label-history coefficient is not determined.
(2) We do not characterize weak or delayed input changes; our upper
bound holds for a fixed Gaussian shift.
(3) Benign shifts can cause a high labeling cost, and the reset variant
that reduces it was evaluated only on Income.
(4) Small natural risk exceedances are not evaluated.
(5) Both real-data tasks come from ACS.
(6) The weakest sufficient error condition is unknown.
Appendix~\ref{app:implementation-limits} lists the controller, grid,
calibration and budget details behind these points.

\FloatBarrier
\label{end:main}
\clearpage
\phantomsection\label{start:references}
\bibliography{references}
\clearpage
\appendix
\raggedbottom
\numberwithin{theorem}{section}
\phantomsection\label{start:supplement}
\section{Validity and full-information bounds}\label{app:validity}

\FloatBarrier
\subsection{Risk-valid mixtures and proof}
\begin{theorem}[Risk-valid SR and CUSUM mixtures]\label{thm:validity}

Suppose each nonnegative factor \(G_t^{(j)}\) satisfies \(E_Q[G_t^{(j)}\mid\mathcal F_{t-1}]\le1\) for every \(Q\in\mathcal H_0\). Start each component at zero and use either
\[
V_t^{(j)}=G_t^{(j)}(1+V_{t-1}^{(j)}),
\quad\text{or}\quad
U_t^{(j)}=G_t^{(j)}\max\{1,U_{t-1}^{(j)}\}.
\]

For a finite deterministic nonnegative weighted average \(S_t\), with weights summing to 1, let \(T_A=\inf\{t:S_t\ge A\}\). Then
\[
E_QT_A\ge A,\qquad Q(T_A\le m)\le m/A
\]

for every null \(Q\) and integer \(m\ge1\). A finite number of components may depend on \(A\) if the mixture is fixed before monitoring.

\end{theorem}

For SR, conditional factor validity gives \(E_Q[V_t^{(j)}\mid\mathcal F_{t-1}]\le1+V_{t-1}^{(j)}\). For CUSUM the corresponding bound is
\[
E_Q[U_t^{(j)}\mid\mathcal F_{t-1}]
\le\max\{1,U_{t-1}^{(j)}\}\le1+U_{t-1}^{(j)}.
\]

Fixed convex averaging preserves nonnegativity, the zero start, and drift at most 1. For \(T=T_A\), summing stopped increments yields
\[
E_QS_{T\wedge n}\le E_Q(T\wedge n),
\quad E_Q(T\wedge n)\ge A\,Q(T\le n).
\]

If \(Q(T=\infty)>0\), the mean run length is infinite. Otherwise monotone convergence gives \(E_QT\ge A\).

For a fixed integer \(m\), \(M_t=S_t+m-t\), \(0\le t\le m\), is a nonnegative supermartingale with \(M_0=m\). If \(T\le m\), then \(\max_{t\le m}M_t\ge A\). The maximal inequality gives \(Q(T\le m)\le m/A\).

For either recursion, induction shows that for every \(s\le t\),
\[
U_t^{(j)}\ge\prod_{i=s}^tG_i^{(j)}.
\]

For SR, use \(1+V_{t-1}\ge1\) when starting a product; for CUSUM use \(\max(1,U_{t-1})\ge1\). The induction step follows by multiplication by the nonnegative factor. If both recursions use the same factors, induction also gives \(U_t^{(j)}\le V_t^{(j)}\). These are component-level statements; mixtures are formed after running the components.

\FloatBarrier
\subsection{Full-information lower bound and proof}
\begin{theorem}[Full-information first-order lower bound]\label{thm:information}

For any admissible, possibly threshold-dependent acquisition kernel and stopping rule satisfying \eqref{eq:FW},
\[
\liminf_{A\to\infty}\frac{E_{1,P}T_A}{\log A}\ge\frac1{h(r)},
\qquad
\liminf_{A\to\infty}\frac{D_P(T_A)}{\log A}\ge\frac1{h(r)}.
\]

\end{theorem}

Define the full-data reference by
\begin{equation}
Q^*(dx,1)=\frac{\tau}{r}P(dx,1),\qquad
Q^*(dx,0)=\frac{1-\tau}{1-r}P(dx,0).\tag{A.1}\label{eq:A1}
\end{equation}

Since \(r<1\), both reweighting factors are finite. The reference has unit mass, risk \(\tau\), and \(D(P\Vert Q^*)=h(r)\). At any observed past, the chosen acquisition mechanism is a common data-processing kernel under \(P\) and \(Q^*\); hence the one-round observed conditional KL \(I_t\) is at most \(h(r)\).

The full-data likelihood ratio takes only the two values \(r/\tau\) and \((1-r)/(1-\tau)\). Its observed counterpart is a \(Q^*\)-conditional average of those values. Consequently, the observed log likelihood ratio increment \(\xi_t\) lies in
\[
[\ell_0,\ell_1]=[\log((1-r)/(1-\tau)),\log(r/\tau)].
\]

Writing \(Z_m=\sum_{t=1}^m\xi_t\), its predictable drift is at most \(mh(r)\), and its martingale differences are bounded by \(\ell_1-\ell_0\). For any \(c>h(r)\),
\[
P(Z_m>cm)\le
\exp\left(-\frac{(c-h(r))^2m}{2(\ell_1-\ell_0)^2}\right).
\]

For $\{T_A\le m\}\in\mathcal F_m$, change of measure at the
deterministic horizon $m$ and the assumed \eqref{eq:FW} bound give
\[
P(T_A\le m)\le e^{cm}m/A+P(Z_m>cm).
\]

Take \(m=\lfloor(1-\eta)\log A/h(r)\rfloor\) and \(h(r)<c<h(r)/(1-\eta)\). Both terms vanish uniformly over admissible threshold-dependent acquisition rules. Therefore \(E_1T_A/\log A\) has lower limit at least \((1-\eta)/h(r)\). Let \(\eta\downarrow0\), and use \(D_P(T_A)\ge E_1T_A\).

On a fixed compact risk interval, the same argument is uniform in \(r\): \(h(r)\) is bounded above and away from zero, and \(\ell_1-\ell_0\) is uniformly bounded. This is the lower-bound uniformity used with Theorem~\ref{thm:upper}.

\FloatBarrier
\subsection{A finite full-audit reference scale}

For finite \(d=E_1T_A\), a bounded stopped KL chain under \eqref{eq:A1} gives KL at \(T_A\wedge m\) at most \(h(r)d\). With \(p=P(T_A\le m)\) and \(q_0=Q^*(T_A\le m)\le m/A\), binary data processing and Markov's inequality imply
\[
h(r)d\ge p\log(A/m)-\log2
\ge(1-d/m)\log(A/m)-\log2.
\]

For sufficiently large \(A\), choose \(m=\lceil(\log A+1)^2\rceil<A\). Then
\begin{equation}
d\ge
\frac{\log(A/m)-\log2}{h(r)+\log(A/m)/m}
=\frac{\log A}{h(r)}-O(1+\log\log A).\tag{A.2}\label{eq:A2}
\end{equation}

An infinite \(d\) satisfies the bound trivially. A full-audit Kelly CUSUM, by bounded-stopping Wald and product domination, has WADD at most \((\log A+\log(r/\tau))/h(r)\). Thus the best full-audit benchmark differs from \(\log A/h(r)\) by at most logarithmic terms. In the waiting-cost comparison where \(\log\log A=o(1/b)\), this distinction is negligible at the \(1/b\) scale.

\FloatBarrier
\FloatBarrier
\subsection{Hard-region identification}
\begin{proposition}[Exact hard-region identification criterion]\label{prop:identification}

For a fixed audit rule \(p=1_B\), let \(P_X=\mu\), \(a(x)=P(L=1\mid X=x)\), and
\[
\mathcal J(p)=\inf_{Q:E_QL\le\tau}D(P^p\Vert Q^p),\qquad
\rho_B=E_P[a(X)1_B(X)].
\]

Here \(P^p\) is the law of \((X,R,RL)\) under the common acquisition mechanism. Then
\[
\mathcal J(1_B)=0\quad\Longleftrightarrow\quad\rho_B\le\tau.
\]

If \(\rho_B>\tau\), \(\mathcal J(1_B)\ge\operatorname{kl}(\rho_B,\tau)>0\).

\end{proposition}
If \(\rho_B\le\tau\), retain the covariate law and conditional labels on \(B\), and set all losses outside \(B\) to zero. This produces an observationally identical valid null. If \(\rho_B>\tau\), the observed variable \(RL\) has mean \(\rho_B\), whereas every null gives it mean at most \(\tau\); binary data processing proves the bound.

\section{A finite additive converse for both acquisition classes}\label{app:additive}

\FloatBarrier
\subsection{Waiting before the first post-change label}

For each \(\nu\), let \(\alpha_\nu=\inf\{t\ge\nu:R_t=1\}\), and define the delay components
\[
\mu_\nu=E_\nu[(T-\nu)^+]=E_\nu Y_\nu+E_\nu Z_\nu,
\]
\[
Y_\nu=(\min\{T,\alpha_\nu\}-\nu)^+,\qquad
Z_\nu=(T-\alpha_\nu)^+.
\]

Set \(Z_\nu=0\) if \(\alpha_\nu=\infty\). It suffices to consider finite WADD.

For label-only acquisition, the wait for the first post-change request has the same law as under \(P_\infty\): the controller does not use online inputs, and no new label has arrived yet. For arbitrary acquisition with unchanged IID inputs, the same coupling works because the covariate marginals agree before and after the change and the full-data rounds are IID. In both cases, reuse the initialization, audit coins, pre-change data and common input sequence until the first new requested loss.

For alarm control, construct \(Q_\nu\) using \(P_0\) before the change and \(P_X\otimes\operatorname{Bernoulli}(\tau)\) afterward. By Assumption~\ref{ass:nominal}, it is a full-null process and has the same observed prefix as \(P_\nu\) before the first post-change label. For \(j\ge1\),
\begin{equation}
P_\nu(Y_\nu\ge j)\ge
P_\infty(\alpha_\nu-\nu\ge j)-(\nu+j-1)/A.\tag{B.1}\label{eq:B1}
\end{equation}

An alarm on the prefix is bounded under \(Q_\nu\) by the finite-window inequality.

For integers \(K,M\ge1\), define
\[
\varphi_M(x)=
\begin{cases}x^2/2,&0\le x\le M,\\Mx-M^2/2,&x>M,\end{cases}
\quad
\Phi_{K,M}(n)=(n+1)\varphi_M((K-n)/(n+1)).
\]

A fixed audit path with \(N_K\) requests divides its \(K-N_K\) unaudited rounds into \(N_K+1\) gaps \(g_i\). The sum of truncated forward waiting times is at least \(\sum_i\varphi_M(g_i)\), including for the final gap, whose next audit may lie after \(K\). Convexity gives
\[
\sum_{\nu=1}^K((\alpha_\nu-\nu)\wedge M)
\ge\Phi_{K,M}(N_K).
\]

The perspective function \(\Phi\) is convex and nonincreasing. Indeed, with \(g=(K-n)/(n+1)\), its derivative is \(\varphi_M(g)-(g+1)\varphi_M'(g)\le0\). Jensen and \eqref{eq:M3} therefore give
\[
E_\infty\Phi_{K,M}(N_K)\ge\Phi_{K,M}(n_*),
\quad n_*=\min\{K,bK+c_0\}.
\]

Summing \eqref{eq:B1}, and using \(\sum_{\nu,j}(\nu+j-1)/(KA)=M(K+M)/(2A)\), yields
\begin{equation}
\frac1K\sum_{\nu=1}^KE_\nu Y_\nu\ge B_{K,M},\quad
B_{K,M}=
\left[\frac{\Phi_{K,M}(n_*)}{K}-\frac{M(K+M)}{2A}\right]_+.\tag{B.2}\label{eq:B2}
\end{equation}

For the waiting-only asymptotic, set \(\kappa_b=\sqrt{(c_0+1)bA}\), \(M=\lceil1/b\rceil\), \(K=\lceil\kappa_b/b\rceil\). If \((c_0+1)/(bA)\to0\), the main term is \((1-b)^2/(2b)(1-o(1))\) and the window penalty is relatively negligible. Some restriction on the threshold is necessary: a valid timer that never audits, with \(S_t=t\), has delay of order \(A\), so a \(1/b\) waiting lower bound cannot hold for all rules when \(A=o(1/b)\).

\FloatBarrier
\subsection{Adding the evidence cost}

Let \(\kappa_\tau=\max\{\log(1/\tau),\log(1/(1-\tau))\}\). Construct the predictable hybrid reference \(Q_\nu^\dagger\): use \(P_0\) before \(\nu\), \(P_X\otimes\operatorname{Bernoulli}(\tau)\) from \(\nu\) through the first post-change request, and \(Q^*\) of \eqref{eq:A1} from the following round onward. By Assumption~\ref{ass:nominal}, its pre-change segment is safe. Both post-change reference laws have risk \(\tau\), and the switch uses only past request indicators, so the reference remains in \(\mathcal H_0\).

Before the first post-change label, \((X,R)\) has the same conditional law under the alternative and reference. Even if acquisition uses current \(X\), its one-round conditional observed KL is
\[
\int p_t(x)\operatorname{kl}(a(x),\tau)\,P_X(dx)
\le\kappa_\tau\int p_t(x)\,P_X(dx).
\]

The probability \(p_t(x)\) must stay inside the integral, because the kernel may depend on the current input. The expected compensated count of the first request, stopped at a bounded time, is at most 1. After that request, the conditional observed KL is at most \(h(r)\) by data processing.

For \(U=T\wedge m\), replace all observations after \(U\) by a cemetery symbol and use the ordinary finite-horizon KL chain rule. This gives
\begin{equation}
D(P_\nu|_{\mathcal F_U}\Vert Q_\nu^\dagger|_{\mathcal F_U})
\le\kappa_\tau+h(r)E_\nu[(U-\alpha_\nu)^+]
\le\kappa_\tau+h(r)E_\nu Z_\nu.\tag{B.3}\label{eq:B3}
\end{equation}

Only bounded stopping times are used, so no optional-stopping identity for unbounded times is needed.

For \(1\le K<m<A\), the event \(\{T\le m\}\) has reference probability at most \(m/A\). Binary data processing and Markov's inequality give, for \(\nu\le K\),
\[
\kappa_\tau+h(r)E_\nu Z_\nu
\ge\left(1-\frac{\mu_\nu}{m-K}\right)\log(A/m)-\log2.
\]

Substitute \(\mu_\nu=E_\nu Y_\nu+E_\nu Z_\nu\), average over \(\nu\le K\), and use \eqref{eq:B2}. Since WADD dominates the average \(\mu_\nu\), we obtain the finite inequality
\begin{equation}
\boxed{
D_P(T_A)\ge
\left[
\frac{B_{K,M}+(\log(A/m)-\kappa_\tau-\log2)/h(r)}
{1+\log(A/m)/(h(r)(m-K))}
\right]_+.
}\tag{B.4}\label{eq:B4}
\end{equation}

Both acquisition classes of Theorem~\ref{thm:additive} satisfy exactly the coupling needed for \eqref{eq:B2}, so this also proves the finite additive extension for unchanged IID inputs.

\FloatBarrier
\subsection{Joint-limit consequence}

For a common \(c_0\le C_0\), choose
\[
M=\lceil b^{-1}\rceil,\quad K=\lceil(C_0+1)b^{-2}\rceil,\quad
m=K+\lceil(\log A+b^{-1}+1)^2\rceil.
\]

Uniformly for \(A\ge b^{-3}\), \(m/A\to0\). The quadratic branch applies and
\[
\Phi_{K,M}(bK+c_0)/K=(1-b)^2/(2b)-O(1).
\]

The window penalty is \(O(1)\). Also \(\log m=O(\mathcal E(b,A))\), and multiplying the denominator correction in \eqref{eq:B4} by its numerator contributes \(O(1)\). Substitution proves \eqref{eq:M5}. The constants can be uniform on the fixed compact risk interval because \(h(r)\) is bounded away from zero.

The role of IID inputs in the unchanged-input condition is substantive. If source inputs are IID fair bits but all post-change inputs equal one common fair bit, every single-time marginal stays the same while the free temporal pattern changes. A trigger based on \(k\) consecutive identical bits has source probability \(2^{1-k}\) but fires persistently after \(k\) changed rounds. This is outside the assumed IID model.

\FloatBarrier
\subsection{Request-epoch strengthening for Lorden delay}
\label{app:epoch}

We prove the all-prefix, (FW) instance of
Theorem~\ref{thm:critical-budget} without a stationarity or renewal
assumption on acquisition. The weaker local assumptions are handled
in Appendices~\ref{app:local-finite}--\ref{app:local-limits}. All requests below refer to the specified
unstopped extension. Put $h=h(r)$, $q=\log A$, and write $C_0\ge0$ for
the common prefix allowance. It suffices to consider finite Lorden delay.
Use the variables $Y_\nu,Z_\nu$ and safe reference $Q_\nu^\dagger$ defined
above. If $E\in\mathcal F_{\nu-1}$ has positive source probability, the
same bounded-horizon chain rule gives
\begin{equation}
 \operatorname{KL}\!\left(
 P_\nu|_{\mathcal F_U,E}\,\Vert\,
 Q_\nu^\dagger|_{\mathcal F_U,E}\right)
 \le\kappa_\tau+hE_\nu[Z_\nu\mid E],
 \quad U=T\wedge m.
 \tag{B.5}\label{eq:epoch-conditional-kl}
\end{equation}
Indeed, the conditional pre-change distributions agree, and the
post-change transition comparisons used in \eqref{eq:B3} remain valid.
The conditioned reference need not be safe before $\nu$, so we apply
(FW) only to the unconditioned $Q_\nu^\dagger$.

\paragraph{Finite path identity and normalization.}
Use $\iota_\nu$ for the request-epoch selector, distinct from
the block states $H_j^C$ and the constant $H$. Add a virtual
request at time zero, and set
\[
 \iota_1=1,\quad \iota_\nu=R_{\nu-1}\ (2\le\nu\le K),\qquad
 p_\nu=E_0\iota_\nu,\quad n=\sum_{\nu=1}^Kp_\nu=1+E_0N_{K-1}.
\]
If the request times in $\{0,\ldots,K-1\}$ are
$0=s_0<\cdots<s_{\ell-1}$, append $s_\ell=K$. Then exactly
\begin{equation}
 \sum_{\nu=1}^K \iota_\nu(\min\{\alpha_\nu,K\}-\nu)^+
 =\sum_{i=0}^{\ell-1}(s_{i+1}-s_i-1)
 =K-(1+N_{K-1}).
 \tag{B.6}\label{eq:epoch-path}
\end{equation}
The last interval is clipped at the common endpoint $K$; the identity
also covers no actual requests or an infinite next request.
For $p_\nu>0$, weight the experiment conditioned on $\iota_\nu=1$ by
$p_\nu/n$. Equivalently, on an augmented space with analyst's index $J$,
\[
 \bar P(J=\nu,d\omega)=\frac{\iota_\nu(\omega)}nP_\nu(d\omega),
 \qquad
 \bar Q(J=\nu,d\omega)=\frac{\iota_\nu(\omega)}n
 Q_\nu^\dagger(d\omega).
\]
Both total masses equal one because $\iota_\nu\in\mathcal F_{\nu-1}$ and
$E_\nu \iota_\nu=E_{Q_\nu^\dagger}\iota_\nu=p_\nu$. Zero-mass terms are omitted;
$J$ is not revealed to the procedure. With
$y=E_{\bar P}Y_J$, $z=E_{\bar P}Z_J$, Lorden's criterion gives
\[
 \mu=y+z=\frac1n\sum_{\nu=1}^K
 E_\nu[\iota_\nu(T-\nu)^+]\le D_P(T).
\]
We never need a particular history that attains the essential supremum.

\paragraph{Waiting and the cost of conditioning.}
Put $w_\nu=(\min\{\alpha_\nu,K\}-\nu)^+$. The first-request coupling
for label-only acquisition or unchanged IID inputs holds conditionally on pre-change history,
so $E_\nu \iota_\nu w_\nu=E_0\iota_\nu w_\nu$. Pathwise,
\[
 Y_\nu\ge w_\nu-(K-\nu)
 1\{T<\min(\alpha_\nu,K)\}.
\]
To make the event comparison explicit, set
$\mathcal A_\nu=\{T<\min(\alpha_\nu,K)\}$. For deterministic $t\ge\nu$,
on $\mathcal C_{\nu,t}=\{R_\nu=\cdots=R_t=0\}$ every post-change
queried-loss coordinate is zero. Since $T$ is an observed stopping
time, $\{T=t\}\cap\mathcal C_{\nu,t}$ is measurable from
$\mathcal F_{\nu-1}$ and $(X_s,R_s)_{\nu\le s\le t}$ on this
no-request record. The common observation kernels give the same
subprobability law for these records under $P_\nu$ and $Q_\nu^\dagger$.
For $t<\nu$, equality follows from their common pre-change law.
Consequently,
\[
 P_\nu(T=t,\alpha_\nu>t)
 =Q_\nu^\dagger(T=t,\alpha_\nu>t)\quad(t<K).
\]
The disjoint union over $t=1,\ldots,K-1$ is $\mathcal A_\nu$, so
$P_\nu(\mathcal A_\nu)=Q_\nu^\dagger(\mathcal A_\nu)
\le Q_\nu^\dagger(T\le K-1)\le(K-1)/A$.
Summing $(K-\nu)(K-1)/A$ and using \eqref{eq:epoch-path} yields
\begin{equation}
 y\ge\frac Kn-1-\frac{K(K-1)^2}{2An}.
 \tag{B.7}\label{eq:epoch-wait}
\end{equation}
Meanwhile, the common distribution of $J$ and
\eqref{eq:epoch-conditional-kl} imply
\begin{equation}
 \begin{split}
 \operatorname{KL}(\bar P|_{J,\mathcal F_U}\Vert
                   \bar Q|_{J,\mathcal F_U})
 &\le\frac1n\sum_\nu
   \{\kappa_\tau p_\nu+hE_\nu(\iota_\nu Z_\nu)\}
   =\kappa_\tau+hz,\\
 \bar Q(T\le m)
 &=\frac1n\sum_\nu Q_\nu^\dagger(\iota_\nu=1,T\le m)
   \le\frac{Km}{An}.
 \end{split}
 \tag{B.8}\label{eq:epoch-kl-fa}
\end{equation}
Neither a lower bound on an individual $p_\nu$ nor (FW) for a
conditioned reference has been used.

\paragraph{Low and high source request mass.}
For $0<b\le1/2$, take
\[
 K=\left\lceil(C_0+1)b^{-2}\right\rceil,\qquad
 M=\lceil2/b\rceil,\qquad
 d=\left\lceil(q+b^{-1}+1)^2\right\rceil,\qquad m=K+d.
\]
Let $x=E_0N_K$, and define
\[
 \epsilon_L=\frac{M(K+M)}{2A},\qquad
 \epsilon_H=\frac{(K-1)^2}{Ab},\qquad
 B_*=[b^{-1}-3-\max(\epsilon_L,\epsilon_H)]_+.
\]
If $x\le bK/2$, use the unconditioned uniform index on
$\{1,\ldots,K\}$. The finite gap argument giving \eqref{eq:B2} needs
only this mean at this $K$, not a new all-prefix budget. Put $\beta=b/2$.
Since $M\ge1/\beta$, the quadratic branch gives
\[
 \frac{\Phi_{K,M}(\beta K)}K
 =\frac{(1-\beta)^2}{2(\beta+1/K)},\qquad
 0\le\frac{(1-\beta)^2}{2\beta}
       -\frac{\Phi_{K,M}(\beta K)}K
 \le\frac{2}{b^2K}\le2.
\]
As $(1-\beta)^2/(2\beta)=b^{-1}-1+b/4$, the resulting waiting
mean is at least $[b^{-1}-3-\epsilon_L]_+\ge B_*$. The same experiment
has stopped KL at most $\kappa_\tau+hz$ and reference alarm probability
at most $m/A\le2m/(Ab)$.

If $x>bK/2$, use the request-weighted experiment above. Because
$E_0R_K\le1$, $n=1+E_0N_{K-1}\ge x>bK/2$. Also
\[
 n\le u=1+b(K-1)+C_0,\qquad
 0\le\frac1b-\frac Ku=\frac{C_0+1-b}{bu}
 \le\frac{C_0+1}{b^2K}\le1.
\]
Equation~\eqref{eq:epoch-wait} therefore gives
$y\ge b^{-1}-2-\epsilon_H$, hence $y\ge B_*$.
Equation~\eqref{eq:epoch-kl-fa} bounds the reference alarm probability
by $2m/(Ab)$.

Thus, whichever deterministic case applies to the source expectation,
there is a proof experiment with $\mu=y+z\le D$, $y\ge B_*$,
stopped KL at most $\kappa_\tau+hz$, and reference alarm probability
at most $2m/(Ab)$. If $Ab>2m$, put $L_*=\log(Ab/(2m))>0$.
Markov and the binary-entropy bound give
$\kappa_\tau+hz\ge(1-\mu/d)L_*-\log2$. Solving yields
\begin{equation}
 \boxed{\displaystyle
 D_P(T_A)\ge
 \left[\frac{B_*+(L_*-\kappa_\tau-\log2)/h}
 {1+L_*/(hd)}\right]_+.}
 \tag{B.9}\label{eq:epoch-finite}
\end{equation}
The two costs are combined on the same selected histories, and the
detector is never restarted at the first label.

\paragraph{Uniform critical and joint limits.}
For fixed $c>0$ and $b=c/q$, the common windows have $K,m=O(q^2)$
and $M=O(q)$, with constants depending only on $c,C_0$. Hence
\[
 \epsilon_L=O(q^3e^{-q}),\quad
 \epsilon_H=O(q^5e^{-q}),\quad
 B_*\ge q/c-3-O(q^5e^{-q}),\quad
 L_*=q-3\log q-O(1).
\]
The condition $Ab>2m$ eventually holds uniformly over the procedure
class. Since $0\le B_*\le q/c$, $0<L_*\le q$ and $d\ge q^2$,
the denominator in \eqref{eq:epoch-finite} costs at most
$1/(ch)+1/h^2$. Substitution proves \eqref{eq:critical-finite}; a fixed
multiple of $q^5e^{-q}$ is absorbed in $C$, using
$q^5e^{-q}\le5^5e^{-5}$. The bound is therefore uniform over threshold-dependent rules with the
same $C_0$, not merely asymptotic for each rule separately.

For the general range $A\ge b^{-5}$, $q\ge5\log(1/b)$,
$\epsilon_H=O((Ab^5)^{-1})=O(1)$ and $\epsilon_L=O(1)$.
Thus $B_*\ge b^{-1}-O(1)$,
$\log(2m/b)=O(\log(1/b)+\log(q+1))$, and $Ab/(2m)\to\infty$
uniformly in this range. The denominator loss is bounded by a fixed
multiple of
\[
 \frac{q^2+q/b}{(q+b^{-1}+1)^2}=O(1).
\]
This proves \eqref{eq:epoch-general}. Over the wider range
$A\ge b^{-3}$ the argument cannot absorb the high-mass error, which is
why Theorem~\ref{thm:additive} keeps the half coefficient there.

\FloatBarrier
\subsection{A budget does not imply a concentrating information clock}
\label{app:clock-example}

The following example shows that a source budget does not by itself
make the observed information concentrate, so information-clock
converses cannot be applied automatically. Take constant inputs,
$r_0=0.1$, $\tau=0.2$, $r=0.3$ and $b=c/\log A$.
Use the unbounded label controller with increments
$Y=L\log(\tau/r_0)+(1-L)\log((1-\tau)/(1-r_0))$ and sleep
$W=\lceil4(1-b)/b\rceil$. Its source excursion mean satisfies
\[
 e\le\frac{\log(9/8)}{\operatorname{kl}(0.1,0.2)}<4.
\]
For an exact certificate, $4\operatorname{kl}(0.1,0.2)-\log(9/8)
=[13\log(9/8)-2\log2]/5>0$, since $9^{13}>4\,8^{13}$.
Independently choose, with equal probabilities, to start immediately or
to wait $\lceil2\log A\rceil$ rounds before starting this controller.
The source renewal argument gives $E_0N_K\le bK+4$ on both branches.
With a Kelly CUSUM, the Lorden delay is $O(\log A)$: the extra initial
sleep is $O(\log A)$, and the controller's post-change sleep bound is
$W/p_{\rm esc}$ with $p_{\rm esc}\ge0.3(1-0.1/0.2)=0.15$.
Here $E_r e^{-Y}=0.9375<1$ supplies the exponential-supermartingale bound.

Set $n_A=\lfloor\log A\rfloor$. Against the safe IID Bernoulli$(\tau)$
reference, the unstopped observed log likelihood is
$\Lambda_n=\sum_{t\le n}R_t\log\{f_r(L_t)/f_\tau(L_t)\}$.
Its mean is $g_A(n_A)=hE_1N_{n_A}\le hn_A/2$, because the delayed
branch has not sampled. On the immediate branch, permanent sampling
has probability at least $0.15$; the IID strong law then gives
$\Lambda_{n_A}/n_A\to h$ on that event, with the constructions coupled
through one label sequence. Consequently, for every $0<\eta<1$,
\[
 \liminf_{A\to\infty}
 P_1\{\Lambda_{n_A}\ge(1+\eta)g_A(n_A)\}\ge0.075.
\]
So even among rules with $O(\log A)$ delay, the unconditional mean
information need not concentrate on this scale. Conditioning on the
initialization removes this particular mixture, but a conditional proof
must then justify its own budget and error bounds. The stopped-mean
argument above needs no concentration.

\FloatBarrier
\subsection{A finite local-calibration inequality}
\label{app:local-finite}

Retain the fixed IID models, nominal safety, and either acquisition
class of Theorem~\ref{thm:additive}. Here $T$ is a general observed
stopping time; no drift-statistic representation is required.
Put $h=h(r)$, $\kappa=\kappa_\tau$, $0<b\le1/2$, and choose integers
$K\ge\lceil(C_0+1)b^{-2}\rceil$, $M=\lceil2/b\rceil$ and $d\ge M$.
Let $m=K+d$. Assume only
\[
 E_0N_K\le bK+C_0,\qquad
 \sup_{Q\in\mathcal H_0}Q(T\le m)\le\delta,\qquad 0<\delta<b/2.
\]
Neither another prefix budget nor a probability bound at another
time is needed. Set
\[
 \epsilon_\delta=\max\{M\delta,(K-1)\delta/b\},\quad
 B_\delta=[b^{-1}-3-\epsilon_\delta]_+,\quad
 \Lambda_\delta=\log\frac b{2\delta}.
\]
We prove
\begin{equation}
 \boxed{\displaystyle
 D_P(T)\ge
 \left[\frac{B_\delta+(\Lambda_\delta-\kappa-\log2)/h}
 {1+\Lambda_\delta/(hd)}\right]_+.}
 \tag{B.10}\label{eq:local-finite}
\end{equation}

The safe reference $Q_\nu^\dagger$, delay components, conditional
stopped KL bound, and request-epoch identity are those of
Appendix~\ref{app:epoch}. They do not themselves use (FW).
Write $x=E_0N_K$. Throughout this argument, $\bar P,\bar Q$
denote the selected joint laws of the analyst's index and observed
record, as in Appendix~\ref{app:epoch}. If $x\le bK/2$, use
\[
 \bar P(J=\nu,d\omega)=K^{-1}P_\nu(d\omega),\qquad
 \bar Q(J=\nu,d\omega)=K^{-1}Q_\nu^\dagger(d\omega).
\]
The deterministic-time event decomposition in
Appendix~\ref{app:epoch} also applies with the local probability bound. For $\nu\le K$ and $j\le M$, premature alarms occur before
$K+M\le m$, so their probability is at most $\delta$.
The truncated-gap argument and $\beta=b/2$ yield
\[
 \begin{split}
 y&\ge[\Phi_{K,M}(\beta K)/K-M\delta]_+,\\
 \Phi_{K,M}(\beta K)/K
 &=\frac{(1-\beta)^2}{2(\beta+1/K)}
 \ge b^{-1}-1+b/4-\frac2{b^2K}.
 \end{split}
\]
Since $K\ge(C_0+1)/b^2$, $y\ge B_\delta$. In this same experiment
the stopped KL is at most $\kappa+hz$ and the reference alarm
probability is at most $\delta\le2\delta/b$.

If $x>bK/2$, use the request-weighted joint laws $\bar P,\bar Q$
with selector $\iota_\nu$. Only the single
prefix assumption is needed, because
\[
 x\le n=1+E_0N_{K-1}\le1+x\le u:=1+bK+C_0,\qquad
 0\le b^{-1}-K/u=\frac{C_0+1}{bu}\le1.
\]
The request-epoch identity and premature-alarm subtraction give
\[
 y\ge K/n-1-\frac{K(K-1)\delta}{2n}
 \ge b^{-1}-2-(K-1)\delta/b\ge B_\delta.
\]
The normalized joint reference satisfies
\[
 \bar Q(T\le m)
 =\frac1n\sum_{\nu=1}^K
 Q_\nu^\dagger(\iota_\nu=1,T\le m)
 \le K\delta/n\le2\delta/b.
\]
The conditional stopped KL bound still gives $\kappa+hz$.
In both experiments $\mu=y+z\le D_P(T)$, $J\le K$, and
$\bar P(T>m)\le\mu/d$. Binary data processing gives
\[
 \kappa+hz\ge(1-\mu/d)\Lambda_\delta-\log2.
\]
Substituting $z=\mu-y$ and solving proves
\eqref{eq:local-finite}. The waiting penalty is therefore
$\max\{M\delta,(K-1)\delta/b\}$, smaller than a bound of order
$K^2\delta/b$.

\FloatBarrier
\subsection{Flexible prefix windows and separated leading costs}
\label{app:local-limits}

Fix $c,C_0,\Gamma$ as in Theorem~\ref{thm:critical-budget}(b).
For $q\ge1$, the prescribed $d_A$ obeys
$M\le d_A\le Hq^2$. Every prefix length $K_A$ allowed by
\eqref{eq:local-prefix} has $m=K_A+d_A\le W_A$.
The budget only has to hold at that prefix, which may depend on the
rule; all constants below use its common range.

Under a calibration bound $\delta_A$ at $W_A$, monotonicity supplies
the bound required at $m$. If $q^3\delta_A\le E_{\mathrm{loc}}$, then, for
$a_c=2/c+1$,
\[
 \epsilon_{\delta_A}
 \le E_{\mathrm{loc}}\max\{a_c,\Gamma/c\}=:E_*,
 \qquad B_{\delta_A}\ge q/c-3-E_*.
\]
Put $\ell_A=\log(1/\delta_A)$ and
$\Lambda=\ell_A-\log(2/b)>0$. If $\ell_A\le L_0q$, the denominator
in \eqref{eq:local-finite} loses at most
$L_0/(ch)+L_0^2/h^2$. To see this for either sign of the numerator,
use
\[
 [X/(1+\eta)]_+\ge X-\eta X_+,\quad
 X_+\le q/c+L_0q/h,\quad \eta=\Lambda/(hd_A)\le L_0/(hq).
\]
Consequently,
\[
 D_P(T_A)\ge\frac{\ell_A}{h}+\frac1b
 -\frac{\log(1/b)}h-C,
\]
where one may take
\[
 C=3+E_*+\frac{\kappa+2\log2}{h}
       +\frac{L_0}{ch}+\frac{L_0^2}{h^2}.
\]
This proves Corollary~\ref{cor:local-calibration}, including its
uniformity over prefix lengths and threshold-dependent rules.

For \eqref{eq:local-fw}, take $\delta_A=W_A/A$.
Now $\ell_A=q-\log W_A\le q$, $q^3\delta_A=O(q^5e^{-q})$,
and $\log W_A\le2\log q+\log(\Gamma+H+1)$.
For sufficiently large $A$, $\delta_A<b/2$; all thresholds are common.
Substitution proves \eqref{eq:critical-finite} under the local
assumptions. Exactly the same argument uses only the endpoint
$\alpha_A^*\le W_A/A$.
For $\delta_A=q^\rho/A$ with fixed $\rho\ge0$, use
$\ell_A=q-\rho\log q$ to obtain the stated
$-(\rho+1)\log q/h-O(1)$ remainder.

Only the critical-budget result is localized; part (a) keeps its
all-prefix and (FW) assumptions.

\FloatBarrier
\subsection{Two boundaries of local calibration}
\label{app:arl-boundary}

\paragraph{ARL alone does not preserve the waiting term.}
Fix $b=c/\log A$. Draw $I\sim\mathrm{Bernoulli}(b)$ independently
at initialization and include it in $\mathcal F_0$, with the same
marginal under every data law. Set $R_t=I$ for all $t$, including
after alarm. If $I=0$, alarm at $T_A=1$; if $I=1$, use full-audit
SR at threshold $a_A^{\mathrm{SR}}=A/b$ with factor
\[
 g(L)=\left(\frac r\tau\right)^L
       \left(\frac{1-r}{1-\tau}\right)^{1-L}
 =1+\frac{r-\tau}{\tau(1-\tau)}(L-\tau).
\]
This belongs to the label-history class, and to the general class
with unchanged IID inputs. It has $E_0N_K=bK$ for every $K$.
For every $Q\in\mathcal H_0$, conditional on $I=1$ the factor mean
is at most one because initialization is already in the filtration.
For its SR crossing $\sigma_A^{\mathrm{SR}}$, bounded stopping gives
\[
 a_A^{\mathrm{SR}} Q(\sigma_A^{\mathrm{SR}}\le n\mid I=1)
 \le E_Q[\sigma_A^{\mathrm{SR}}\wedge n\mid I=1].
\]
If the conditional mean is infinite there is nothing to prove;
otherwise letting $n\to\infty$ shows it is at least $a_A^{\mathrm{SR}}$. Therefore
$E_QT_A\ge1-b+b\,a_A^{\mathrm{SR}}\ge A$ for every safe process, including
history-dependent ones.

On the changed IID law, log factors have mean $h(r)$ and upper
increment $u=\log(r/\tau)$. The product walk crossing $\log a_A^{\mathrm{SR}}$
has mean duration at most $(\log a_A^{\mathrm{SR}}+u)/h(r)$, by bounded-stopping
Wald and the overshoot bound, followed by monotone convergence.
SR dominates this product from every changepoint and every observed
history. The inactive branch has Lorden delay at most one, so
\[
 D_P(T_A)\le
 \frac{\log(A/b)+\log(r/\tau)}{h(r)}
 =\frac{\log A}{h(r)}+O(\log\log A).
\]
So for rules controlled only through ARL, the additional $1/c$
coefficient fails even with an exact all-prefix budget. The
construction ignores inputs, so constant inputs with Bernoulli source
and target losses give a concrete IID instance with unchanged inputs.

Here $\alpha_A^*\ge1-b\to1$. Both the small-error waiting
correction and $\Lambda_\delta>0$ fail: this is not an instance
of \eqref{eq:decoupled-local} with only its evidence term removed.
For $\nu\ge2$, any positive-probability survival event $T_A\ge\nu$
forces $I=1$, where acquisition has rate one. Survival conditioning therefore destroys the unconditional source
budget.

\paragraph{A finite window limits the unclipped information term.}
Let $W_A=O(q^2)$ be the common endpoint and set
$\varepsilon_A=\exp(-q^3)$. Draw an independent bit with success
probability $\varepsilon_A$. Never request labels. Alarm at 1
on success and at $W_A+1$ otherwise.
For every safe law, and with initialization included in the
observed filtration,
\[
 \alpha_A^*=\varepsilon_A,\qquad E_0N_K=0,\qquad
 D_P(T_A)=W_A+1=O(q^2).
\]
Every polynomial waiting-error expression, including
$K^2\alpha_A^*/b$ for $K=O(q^2)$, tends to zero.
Nevertheless $\log(1/\alpha_A^*)=q^3$, so the proposed
unqualified lower bound
$D\ge\log(1/\alpha_A^*)/h+1/b-O(\log q)$ is false.
The denominator in \eqref{eq:local-finite} reflects this limit of a
finite window. Some control of the ratio of information to horizon is
needed; the bounded scale or clipped calibration above is one
sufficient choice.

\FloatBarrier
\FloatBarrier
\subsection{Calibration, endpoint and delay-criterion details}
The same comparison exposes the local false-alarm level separately.
For each rule, write
$\alpha_A^*=\sup_{Q\in\mathcal H_0}Q(T_A\le W_A)$.

Appendices~\ref{app:local-finite}--\ref{app:local-limits} prove
Corollary~\ref{cor:local-calibration}. We add remarks on the endpoint
and the delay criterion.

The leading costs separate local calibration from budget waiting, but
their finite-horizon error conditions still interact. One may take
$\delta_A=\alpha_A^*$ when it meets the displayed conditions.
For smaller actual error, a valid clipped choice is
$\delta_A=\max\{\alpha_A^*,A^{-L_0}\}$, provided the common error bound
also holds. A fixed finite window cannot support an arbitrarily large
unclipped $\log(1/\alpha_A^*)$ term.

\paragraph{Polynomial local calibration retains the waiting scale.}
For \(\delta_A=q^{-3}\), the calibration assumptions hold eventually
with \(q^3\delta_A=1\) and \(\ell_A=3\log q\).
Since \(\log(1/b)=\log q-\log c\), \eqref{eq:decoupled-local} gives
\[
 D_P(T_A)\ge\frac1b+\frac{2\log q}{h(r)}-C.
\]
Thus the \(1/b\) waiting term persists even when the net evidence
contribution in this bound is only logarithmic in \(q\).
The budget also enters the logarithmic conditioning loss, and the local
error conditions are essential: the ARL-only counterexample has
\(\alpha_A^*\to1\).

\paragraph{One common endpoint is enough.}
Although \eqref{eq:local-fw} is convenient to state, its conclusion
only needs $\alpha_A^*\le W_A/A$. Monotonicity then bounds every
earlier premature-alarm probability by the same $W_A/A$.
The corresponding errors are $O(q^3/A)$ and $O(q^5/A)$.
For example, calibration $q^\rho/A$, for fixed $\rho\ge0$, instead gives
the same leading terms with remainder
$-(\rho+1)\log q/h(r)-O(1)$.

The half-coefficient bound of Theorem~\ref{thm:additive} also
lower-bounds Pollak delay: whenever \(P_\nu(T\ge\nu)>0\),
\[
 E_\nu[(T-\nu)^+]
 =P_\nu(T\ge\nu)E_\nu[T-\nu\mid T\ge\nu]
 \le E_\nu[T-\nu\mid T\ge\nu].
\]
Its optimal correction under that criterion is not determined here.

\section{Label-controller renewal and escape arguments}\label{app:controller}
\begin{lemma}[Label-controller budget and total sleep]\label{lem:controller}

For the controller in Section~\ref{sec:controllers}, with \(d_0=\log((1-r_0)/(1-\tau))\),
\[
1\le e_0\le d_0/\operatorname{kl}(r_0,\tau)<\infty,
\quad E_0N_K^L\le b_LK+e_0.
\]

For every post-change \(r\ge\tau\),
\[
p_{\rm esc}(r)\ge r(1-r_0/\tau)\ge\tau-r_0>0.
\]

If \(\Sigma\) is the total number of future sleep rounds from the change, including the current residual sleep, then uniformly over prior histories
\[
E_{\nu,P}[\Sigma\mid\mathcal F_{\nu-1}]\le W/p_{\rm esc}(r).
\]

\end{lemma}

Write \(a_0=\log(\tau/r_0)>0\), \(d_0=\log((1-r_0)/(1-\tau))>0\). Under source rate \(r_0\), the score increment has mean \(-\operatorname{kl}(r_0,\tau)\). Let \(E\) be its first passage below zero from 0. The stopped score at \(E\wedge n\) is always at least \(-d_0\). Bounded-stopping Wald gives
\[
\operatorname{kl}(r_0,\tau)E(E\wedge n)\le d_0.
\]

Monotone convergence proves \(1\le e_0=EE<\infty\) and the displayed upper bound.

If the actual source rate is at most \(r_0\), couple labels with common uniforms. The lower-rate excursion has no larger score at each sampling step and crosses below zero no later. Its mean length \(e\) is at most \(e_0\). This is monotonicity of individual excursions, not a claim that entire calendar audit indicators are ordered.

Let \(M_K\) be the number of excursions started by round \(K\). Whether excursion \(i\) starts is determined before its fresh labels, so the expected sum of its full length over started excursions equals \(eEM_K\). Covering the incomplete final excursion gives \(EN_K^L\le eEM_K\). Every later excursion starts only after its preceding \(W\) sleep rounds have finished, hence pathwise
\[
K\ge N_K^L+W(M_K-1).
\]

Combining the two inequalities yields
\[
EN_K^L\le\frac e{e+W}K+\frac{eW}{e+W}\le b_LK+e_0.
\]

For \(C\)'s low first block and nominal trigger probability at most \(b_C\), \(EN_K^C\le b_CK\), so the Hybrid union has allowance \(e_0\).

For post-change \(r\ge\tau\),
\[
E_r e^{-Y}=r\,r_0/\tau+(1-r)(1-r_0)/(1-\tau)\le1.
\]

For the unreset random walk beginning at score \(x>0\), \(e^{-Z_n}\) is a nonnegative supermartingale, and the probability of ever entering the negative half-line is at most \(e^{-x}\). With probability \(r\), a first positive label reaches \(a_0\). Therefore
\[
p_{\rm esc}(r)\ge r(1-e^{-a_0})=r(1-r_0/\tau)\ge\tau-r_0.
\]

Starting at any nonnegative score only increases survival probability under the same increments. At the change, at most \(W\) rounds of an existing sleep remain. Thereafter each failed excursion incurs \(W\) more sleep rounds and has conditional escape probability at least \(p_{\rm esc}\). If \(F\) is the number of failures before permanent sampling,
\[
P(F\ge k\mid\mathcal F_{\nu-1})\le(1-p_{\rm esc})^k,
\quad EF\le(1-p_{\rm esc})/p_{\rm esc}.
\]

Thus \(E[\Sigma\mid\mathcal F_{\nu-1}]\le W/p_{\rm esc}\), and infinitely many future label-controller requests occur almost surely. Because \(e_0\ge1\),
\[
W\le e_0(1-b_L)/b_L+1\le e_0/b_L.
\]

A fixed positive cap breaks this argument: with both outcomes possible, the finite-state active chain eventually returns to sleep, so permanent escape fails. The capped implementation therefore does not inherit this fallback bound.

\FloatBarrier
\subsection{An actual-rate prefix certificate for the original capped L}
\label{app:acs-prefix}
This statement concerns the $+25/-7$, cap-$1000$ implementation and its
held-out source law, not the uncapped sensitivity variant. Let
$p=6710/39133$ and $d=7-32p>0$. At active score $i$, the excursion
stopped at a bounded time has score at least $-7$, while capping can
only decrease its drift. Its expected remaining active length obeys
\[
 e_i\le(i+7)/d\le1007/d<666.
\]
For a sleep of $W$ rounds let $e=e_0$ and $b'=e/(e+W)$.
On the finite controller state space, put
\[
 u(i)=(1-b')e_i-b'W\quad\text{in an active state},\qquad
 u(\text{sleep }k)=-kb'\quad(1\le k\le W).
\]
The excursion recursion gives $u-Pu=a-b'$, where $a$ indicates an
active round. Its span is $(1-b')\max_i e_i\le1007/d<666$.
Telescoping from any initial state distribution therefore gives
\[
 E_0N_K-b'K=E_0u(Z_0)-E_0u(Z_K)\le666.
\]
This prefix begins after the experimental warmup. Charging warmup labels
requires including their initialization cost. Source-chain rates are
approximately $1.004275\%$, $0.502281\%$, $0.201042\%$; for the last
one an upward certified allowance is $0.00201041855559$.
The certificate uses the actual source error fraction; the saved
initialization need not be exactly stationary. A residual enclosure for
the finite linear system is retained in the supplementary audit records.
The certificate covers L at its actual rate, not the nominal target;
C and Hybrid use bootstrap-calibrated budgets. A large allowance such
as 666 also makes any finite numerical converse weak.

\FloatBarrier
\section{Audit-clock and block interfaces}\label{app:audit}
\begin{lemma}[Audit-clock and block interfaces]\label{lem:audit}

For a fixed finite mixture of \(J\) components, suppose a component has post-change conditional mean log-growth \(g>0\) on every detection audit, and positive log increment at most \(\ell_+\). Then Hybrid satisfies
\begin{equation}
D_H\le\frac{\log(JA)+\ell_+}{g}
+\frac{e_0}{p_{\rm esc}(r)b_L}.\tag{D.1}\label{eq:M6}
\end{equation}

If the covariate block trigger has probability \(\pi>0\), both \(C\) and Hybrid satisfy
\begin{equation}
D\le2w+\frac{\log(JA)+\ell_+}{\pi g}+\frac{2w}{\pi}.\tag{D.2}\label{eq:M7}
\end{equation}

\end{lemma}

\FloatBarrier
\subsection{Audit-clock bound}

Choose a component whose post-change mean log-growth is \(g>0\). Requests are decided before the current input, and full-data rounds are IID, so every requested loss has conditional law Bernoulli\((r)\). The selected log factors \(\xi_k\), ordered by actual Hybrid audit number, have conditional mean \(g\) and bounded increments. Appendix~\ref{app:controller} guarantees that all audit times are finite.

Let \(z=\log(JA)\) and
\[
K_*=\inf\{k:\xi_1+\cdots+\xi_k\ge z\}.
\]

For \(K_*\wedge n\), bounded-stopping Wald and the bounded positive overshoot imply
\[
gE[K_*\wedge n\mid\mathcal F_{\nu-1}]
=E\left[\sum_{i=1}^{K_*\wedge n}\xi_i\mid\mathcal F_{\nu-1}\right]
\le z+\ell_+.
\]

Monotone convergence proves finite \(EK_*\) and its bound. Product domination ensures alarm by that audit. Every preceding unaudited Hybrid round belongs to a sleep of \(L\). Therefore
\[
(T_A-\nu+1)^+\le K_*+\Sigma
\]

pathwise, including the case of an earlier alarm. Adding conditional expectations, without assuming independence, proves \eqref{eq:M6}.

\FloatBarrier
\subsection{A boundary compensator for blocks}

Discard at most \(2w\) rounds to align a complete post-change block origin. In the new coordinates, \(H_1^C\) is arbitrary; \(H_{j+1}^C\) is determined by the fresh full-data block \(j\). With full-data block-end proof filtration \(\mathcal B_j\),
\[
E[H_{j+1}^C\mid\mathcal B_{j-1}]=\pi,\qquad
M_n=\sum_{j=1}^n(H_{j+1}^C-\pi)
\]

is a martingale. Enlarging the proof filtration to include unrequested losses is legitimate under IID full-data blocks; the actual states remain predictable.

Let \(Z_t\) be the selected Hybrid product's log-growth from this origin, \(\tau_z\) its first crossing of \(z\), \(\tau_n=\tau_z\wedge nw\), and \(\sigma_n=\lceil\tau_n/w\rceil\). The last is a bounded block-end stopping time. Optional stopping and a telescoping boundary term give
\[
E\sum_{j=1}^{\sigma_n}H_j^C
=\pi E\sigma_n+EH_1^C-EH_{\sigma_n+1}^C
\ge\pi E\sigma_n-1.
\]

The unfinished final block loses at most \(w\) high-state rounds, so \(EN_C(\tau_n)\ge\pi E\tau_n-2w\).

Meanwhile \(Z_t-gN_H(t)\) is a round-level martingale. Crucially, both counts are evaluated at this same Hybrid stopping time:
\[
g(\pi E\tau_n-2w)\le gEN_C(\tau_n)
\le gEN_H(\tau_n)=EZ_{\tau_n}\le z+\ell_+.
\]

For C alone the two counts coincide. Hence
\[
E\tau_n\le(z+\ell_+)/(\pi g)+2w/\pi.
\]

Monotone convergence proves finite expected crossing time, and product domination plus the initial alignment proves \eqref{eq:M7}, uniformly over histories. The argument uses neither independence of adjacent block rewards nor a pathwise ordering of two alarm times.

For example, additional audits need not accelerate a fixed path: with \(\tau=0.2,\lambda=1,A=3\) and losses \(1,0,1\), auditing only rounds 1 and 3 gives CUSUM \(1.8,1.8,3.24\), whereas auditing all three gives \(1.8,1.44,2.592\). The proof therefore uses drifts and counts rather than pathwise monotonicity.

\FloatBarrier
\section{Growing grids and upper bounds}\label{app:grid}
\begin{lemma}[A risk-adaptive betting grid]\label{lem:grid}

Fix a known design interval \(\mathcal R=[\tau+\Delta,\bar r]\) with \(\Delta>0\) and \(\bar r<1\). The detector does not know the actual \(r\). Let
\[
\psi(\lambda;r)=r\log(1+\lambda(1-\tau))+(1-r)\log(1-\lambda\tau),
\quad
\lambda^\star(r)=\frac{r-\tau}{\tau(1-\tau)}.
\]

For \(\lambda_{\min}=\lambda^\star(\tau+\Delta)\), \(\lambda_{\max}=\lambda^\star(\bar r)\), define
\[
\begin{aligned}
\lambda_{j,A}&=\lambda_{\max}(1+\varepsilon_A)^{-j},
\quad j=0,\ldots,J_A-1,\\
J_A&=1+\left\lceil\frac{\log(\lambda_{\max}/\lambda_{\min})}
{\log(1+\varepsilon_A)}\right\rceil.
\end{aligned}
\]

For every \(r\in\mathcal R\), the largest grid point not exceeding \(\lambda^\star(r)\) satisfies
\[
\psi(\lambda_{j,A};r)\ge h(r)/(1+\varepsilon_A).
\]

With \(\varepsilon_A=1/\log A\), \(J_A=O(\log A)\), \(\log J_A=O(\log\log A)\), and the positive log increment is uniformly at most \(B=\log(\bar r/\tau)\).

\end{lemma}
For uniform weights, a finite version makes the costs explicit. Write \(q=\log A\), \(B=\log(\bar r/\tau)\). Then
\[
D_{\mathrm{cov}}\le\frac q{h(r)}
+\frac{4+3(\log J_A+B)}{h(r)}+6w,
\]
\[
D_{H,\mathrm{label}}\le\frac q{h(r)}
+\frac{e_0}{p_{\rm esc}(r)b_L}
+\frac{1+\tfrac32(\log J_A+B)}{h(r)}.
\]

Weights bounded below as stated add only a fixed \(\log(1/c_w)\) inside the mixing term. Appendix~\ref{app:grid} derives these bounds from Lemma~\ref{lem:audit}, without absorbing a growing \(J_A\) into a constant.

\begin{corollary}[Comparison with the lower bounds]\label{cor:comparison}

Let \(b_L=\theta b\), \(b_C=(1-\theta)b\), where \(0<\theta<1\) is fixed.

\begin{itemize}
\item Under covariate harmful shift and \(\log(1/b)=o(\log A)\), \(C\) and Hybrid attain \(D/\log A\to1/h(r)\), without knowing \(r\).
\item Under pure label change, \(A\ge b^{-3}\), and
\(\log\log A=o(1/b)\), Theorems~\ref{thm:additive} and~\ref{thm:upper} share the same evidence term
and match the inverse-budget waiting order. Relative to this broader
range's half-coefficient lower bound, the upper/lower waiting-constant
ratio is bounded by \(2e_0/(\theta p_{\rm esc}(r))\), or
\(4e_0/p_{\rm esc}(r)\) for equal allocation.
\item At the critical budget \(b=c/\log A\), Theorem~\ref{thm:critical-budget}
raises the lower waiting coefficient to one. The corresponding
upper/lower waiting-constant ratio is bounded by
\(e_0/(\theta p_{\rm esc}(r))\), or \(2e_0/p_{\rm esc}(r)\) for equal
allocation. The label-only version has \(\theta=1\).
These compare theoretical waiting terms, not total delays or
fixed-changepoint empirical mean differences.
\item To additionally obtain full-information first-order optimality under pure label change requires \(1/b_L=o(\log A)\). A condition only on \(\log(1/b)\) cannot remove the waiting term.
\end{itemize}

\end{corollary}

The derivative of \(\psi\) is
\[
\partial_\lambda\psi=
\frac{r-\tau-\lambda\tau(1-\tau)}
{(1+\lambda(1-\tau))(1-\lambda\tau)}.
\]

Its derivative is strictly decreasing, giving the stated unique Kelly maximizer. For the lower adjacent geometric point, \(\theta=\lambda_j/\lambda^\star\in[(1+\varepsilon)^{-1},1]\). Concavity and \(\psi(0)=0\) give \(\psi(\lambda_j;r)\ge\theta h(r)\ge h(r)/(1+\varepsilon)\).

Put \(c=\log(\lambda_{\max}/\lambda_{\min})\). For \(q=\log A\ge2\) and \(\varepsilon=1/q\), \(x/(1+x)\le\log(1+x)\le x\) implies
\[
1+cq\le J_A\le2+c(q+1).
\]

The degenerate single-risk interval uses \(J_A=1\). Also every grid factor lies between \((1-\bar r)/(1-\tau)\) and \(\bar r/\tau\). Thus drift, positive overshoot, and negative increment bounds are uniform in \(A,r\).

At each fixed \(A\), apply Lemma~\ref{lem:audit} to the grid component chosen above; different thresholds may use different components and martingales. For uniform weights and \(\pi_A\ge1-1/q\),
\[
D_{\rm cov}\le2w+
\frac{(1+1/q)(q+\log J_A+B)}{(1-1/q)h(r)}
+\frac{2w}{1-1/q}.
\]

Since \((1+1/q)/(1-1/q)\le3\), \(2q/(q-1)\le4\), and \(2+2/(1-1/q)\le6\), this gives the finite covariate inequality in Theorem~\ref{thm:upper}. The pure-label inequality follows directly from \eqref{eq:M6}, using \(1+1/q\le3/2\). For weights at least \(c_w/J_A\), the product threshold is at most \(J_AA/c_w\), adding a fixed logarithmic penalty.

For completeness, the high-power block construction itself has no dependence on \(J_A\). With a central chi-square tail \(\alpha\le b_*\), its threshold obeys
\[
c_{b_*}\le4\log(2^{d/2}/\alpha).
\]

For \(Z\sim N(0,I_d)\),
\[
P\{\|Z\|>\sqrt{4\log(2^{d/2}\log A)}\}\le1/\log A.
\]

Taking
\[
w=\left\lceil
\frac{(\sqrt{c_{b_*}}+\sqrt{4\log(2^{d/2}\log A)})^2}{\delta^2}
\right\rceil
\]

gives the required power by the triangle inequality and the stated order for \(w\). Combined with \(h(r)\ge h_{\min}>0\) and \(\log J_A=O(\log\log A)\), the displayed remainders follow uniformly.

For the optional quadratic refinement, let \(\rho=\tau+\lambda_j\tau(1-\tau)\). Then \(\tau\le\rho\le r\), \(r-\rho\le\varepsilon(r-\tau)\), and \eqref{eq:M8} holds. Integrating in the second KL argument,
\[
\operatorname{kl}(r,\rho)
=\int_\rho^r\frac{r-s}{s(1-s)}\,ds
\le\frac{(r-\rho)^2}{2\tau(1-\bar r)}.
\]

Since \(h(\tau)=h'(\tau)=0\) and \(h''(r)=1/[r(1-r)]\ge4\), \(h(r)\ge2(r-\tau)^2\). Therefore
\[
\psi(\lambda_j;r)\ge h(r)(1-C\varepsilon^2),
\qquad C=1/[4\tau(1-\bar r)].
\]

If \(\varepsilon_A=1/\sqrt q\), then \(J_A\le2+c(\sqrt q+1)\). For \(q\ge\max(2,2C)\), \(1/g_A(r)\le(1+2C/q)/h(r)\), preserving the same remainder order. The experiments do not use this alternative.

For reference, deterministic evaluation with \(\tau=0.2012,\bar r=0.5,\Delta=0.005\) gives:

\begin{table}[ht]
\centering
\small
\caption{Growing-grid guarantees and numerical minima for the fixed design interval. Numerical minima are floating-point evaluations, not certified lower bounds.}
\label{tab:10}
\setlength{\tabcolsep}{4pt}
\renewcommand{\arraystretch}{1.15}
\begin{tabular}{@{}lrrr@{}}
\toprule
\textbf{\(A\)} & \textbf{\(J_A\)} & \textbf{G1 bound} & \textbf{Minimum \(g/h\)} \\ \midrule
1e+04 & 41 & 0.902060 & 0.997381154 \\
1e+06 & 60 & 0.932503 & 0.998794830 \\
1e+08 & 79 & 0.948508 & 0.999310467 \\
\bottomrule
\end{tabular}
\end{table}

The reported minima are floating-point evaluations at the endpoints and affine-envelope intersections. On a positive affine envelope segment \(L(r)\), the derivative numerator of \(L(r)/h(r)\) is \(L'h-Lh'\), whose derivative is \(-Lh''<0\); an interior stationary point is therefore a maximum. This reduces minimization to the finite candidate set; the guarantee itself comes from the inequalities above.

If \(\Delta_A\to0\), the finite formulas still apply, but their \(1/h(r)\) terms can no longer be absorbed into a fixed constant. Continuous-prior mixtures and dynamically added components would need separate arguments and are not used here.

\FloatBarrier
\section{Fixed-audit information and oracle benchmarks}\label{app:fixed-audit}

These supplementary results give the fixed-audit information comparisons and the oracle factors used in the synthetic study.

\FloatBarrier
\subsection{Observed KL and concavity}

For a fixed measurable audit function \(p\), write \(P_X=\mu\), \(a(x)=P(L=1\mid X=x)\), \(Q_X=\nu\), and \(q_0(x)=Q(L=1\mid X=x)\). The common audit kernel gives
\begin{equation}
D(P^p\Vert Q^p)
=D(\mu\Vert\nu)+\int p(x)\operatorname{kl}(a(x),q_0(x))\,\mu(dx).\tag{F.1}\label{eq:F1}
\end{equation}

Observed inputs contribute the first term, whereas missing labels contribute no label KL. This identity is affine in \(p\) for every fixed \(Q\). Because the full-data class \(\{Q:E_QL\le\tau\}\) does not depend on \(p\), its pointwise infimum \(\mathcal J(p)\) is concave:
\[
\mathcal J(\theta p_1+(1-\theta)p_2)
\ge\theta\mathcal J(p_1)+(1-\theta)\mathcal J(p_2).
\]

Maximizing it under \(E_0p\le b\) and \(0\le p\le1\) is concave maximization over a convex set. This does not prove existence in an unrestricted function space or identify a universally optimal fixed ranking. A shared risk constraint couples the conditional regions after taking the infimum.

\FloatBarrier
\subsection{Explicit full-null growth-optimal factor under exploration}\label{app:oracle-factor}

Assume \(p_*\le p(x)\le1\), \(0<a(x)<1\), \(E_PL>\tau\), and fix measurable versions on the whole covariate space. Extend \(a\) inside \((0,1)\) on any \(\mu\)-null sets, since nulls may place mass there. All quantities below are pointwise functions of \(x\); the scalar \(\lambda\in(0,1/\tau)\) is chosen before the current input.

Use local notation \(c_{\lambda,0}=1-\lambda\tau\), \(c_{\lambda,1}=1+\lambda(1-\tau)\), \(d_\lambda=\lambda/p\), \(B_\lambda=(1-pa)d_\lambda-c_{\lambda,0}\), and
\begin{equation}
f_0=\frac{-B_\lambda+\sqrt{B_\lambda^2+4c_{\lambda,0}(1-a)d_\lambda}}2,\quad
f_1=f_0+d_\lambda,\quad
f_M=\frac{f_0f_1}{f_0+(1-a)d_\lambda}.\tag{F.2}\label{eq:F2}
\end{equation}

Let \(G_\lambda\) equal \(f_M\) on a missing label, \(f_0\) on an audited negative label, and \(f_1\) on an audited positive label. This local \(f_0\) is unrelated to the controller excursion constant \(e_0\).

\textbf{Proposition F.1.} These factors are positive and satisfy
\begin{equation}
(1-p)f_M+pf_0=c_{\lambda,0},\quad
(1-p)f_M+pf_1=c_{\lambda,1},\quad
\frac{1-a}{f_0}+\frac a{f_1}=\frac1{f_M}.\tag{F.3}\label{eq:F3}
\end{equation}

Consequently \(E_QG_\lambda=1+\lambda(E_QL-\tau)\le1\) for every full-risk null under this acquisition kernel. With \(m_\lambda=(f_M-c_{\lambda,0})/\lambda\),
\[
G_\lambda=
1+\lambda\left[m_\lambda(X)+\frac R{p(X)}(L-m_\lambda(X))-\tau\right].
\]

\textbf{Proof.} For fixed \(a,p,\lambda\), maximize
\[
(1-p)\log u+p(1-a)\log v_0+pa\log v_1
\]

subject to \((1-p)u+pv_0=c_{\lambda,0}\), \((1-p)u+pv_1=c_{\lambda,1}\), with positive factors.

\emph{Case \(0<p<1\).} Eliminating \(v_0,v_1\) leaves a function of \(u\in(0,c_{\lambda,0}/(1-p))\) that is strictly concave: the term \((1-p)\log u\) is strictly concave and the remaining terms are concave. The objective tends to \(-\infty\) at both endpoints. Its first-order equation \(1/u=(1-a)/v_0+a/v_1\) therefore has a unique solution, yielding the factors in \eqref{eq:F2}.

\emph{Case \(p=1\).} The objective does not depend on \(u\). The realized factors are \(c_{\lambda,0},c_{\lambda,1}\), which \eqref{eq:F2} also gives at \(p=1\). The \(f_M\) in \eqref{eq:F2} satisfies the harmonic identity for every admissible \(p\), since \((1-a)f_1+af_0=f_0+(1-a)d_\lambda\); this also preserves \eqref{eq:F3} when the missing-label branch is unused.

In both cases, the first two identities imply \(E[G_\lambda\mid X,L]=1+\lambda(L-\tau)\); integration proves validity.

\textbf{Proposition F.2.} The function \(g(\lambda)=E_P\log G_\lambda\) has an interior maximizer \(\widehat\lambda_p\). There exists \(\widetilde Q_p\) with risk \(\tau\) such that
\begin{equation}
G_{\widehat\lambda_p}=\frac{dP^p}{d\widetilde Q_p^p},\qquad
g(\widehat\lambda_p)=\mathcal J(p)
=\sup_{G\text{ valid for the observed null}}E_P\log G.\tag{F.4}\label{eq:F4}
\end{equation}

\textbf{Proof.} Set \(u_0=(1-a)/f_0\), \(u_1=a/f_1\). Then \(u_0+u_1=1/f_M\) and \(c_{\lambda,0}u_0+c_{\lambda,1}u_1=1\) pointwise. The identities yield \(c_{\lambda,0}\le f_M\le c_{\lambda,1}\), \(f_0\ge c_{\lambda,0}(1-a)\), \(f_1\ge c_{\lambda,0}a\), \(f_0\le c_{\lambda,0}/p_*\), and \(f_1\le c_{\lambda,1}/p_*\). They ensure expected-log integrability and envelope differentiation on compact interior intervals:
\[
g'(\lambda)=\int[(1-\tau)u_1-\tau u_0]\,d\mu
=M_1(\lambda)-\tau M(\lambda),
\]

where \(M=\int(u_0+u_1)d\mu\) and \(M_1=\int u_1d\mu\).

A feasible augmented bet has positive derivative \(r-\tau\) at zero, so \(g\) is positive somewhere. As \(\lambda\uparrow1/\tau\), audited negative outcomes have positive probability and \(f_0\le c_{\lambda,0}/p_*\to0\), whereas other positive log contributions remain bounded above. Thus \(g\to-\infty\); continuity gives an interior maximum. At the maximum, \(M_1=\tau M\), and integrating \(c_{\lambda,0}u_0+c_{\lambda,1}u_1=1\) gives \(M=1\), \(M_1=\tau\).

Define \(\widetilde Q_p(dx,0)=u_0(x)\mu(dx)\), \(\widetilde Q_p(dx,1)=u_1(x)\mu(dx)\). It has unit mass and risk \(\tau\); its input marginal is \(\mu/f_M\). Applying the common audit kernel gives \eqref{eq:F4} branchwise. For any other valid \(G\), Jensen yields
\[
E_P\log(G/G_{\widehat\lambda_p})\le\log E_{\widetilde Q_p^p}G\le0.
\]

The same entropy inequality bounds \(E_P\log G_{\widehat\lambda_p}\) by KL to every observed null, with equality at \(\widetilde Q_p^p\).

This construction specializes growth-optimal e-variable theory to the observed audit mechanism.

\textbf{Corollary F.3.} For the fixed acquisition and fixed oracle factor, SR or CUSUM has \(E_1T_A/\log A\to1/\mathcal J(p)\) and \(D_P(T_A)/\log A\to1/\mathcal J(p)\).

Write $I_p=\mathcal J(p)$,
$B_p=\log(c_{\widehat\lambda_p,1}/p_*)$ and
$v_p=\operatorname{Var}_P(\log G_{\widehat\lambda_p})<\infty$.
The second moment is finite by Proposition F.2's bounds and
$\sup_{0<x\le1}x\log^2x<\infty$.
Bounded-stopping Wald at the first product crossing, followed by
monotone convergence of its mean, gives the uniform-history upper bound
$(\log A+B_p)/I_p$; SR/CUSUM dominates this product from every start.
For the lower bound, use the maximal change-of-measure argument for
\eqref{eq:F6}, with $(\bar I,\bar v,\kappa_\tau)$ replaced by
$(I_p,v_p,B_p)$. For fixed $0<\epsilon<1$, take
$m=\lfloor(1-\epsilon)\log A/I_p\rfloor$ and
$u=\epsilon\log A/2$. Its two error terms vanish as $A\to\infty$.
Thus $\liminf E_1T_A/\log A\ge(1-\epsilon)/I_p$;
letting $\epsilon\downarrow0$ and using $D_P(T_A)\ge E_1T_A$
gives both matching lower constants. Here the observation rule and
oracle factor are fixed throughout the threshold limit.

\subsection{Fixed-input auditing in the Gaussian shift model}

\textbf{Theorem F.4.} For a fixed acquisition function \(p\), independent of \(A\), with \(E_0p\le b\), the valid SR/CUSUM alarm family in the Gaussian input model obeys
\begin{equation}
\liminf_{A\to\infty}\frac{D_u(T_A)}{\log A}
\ge\frac1{\kappa_\tau U_\delta(b)},\quad
U_\delta(b)=\Phi(\Phi^{-1}(b)+\delta).\tag{F.5}\label{eq:F5}
\end{equation}

The rule may know the direction \(u\). This is not an exchange of the threshold limit with an infimum over all \(A\)-dependent rules.

\textbf{Proof.} The Neyman--Pearson input region \(p^*=1\{dP_u/dP_0>c\}\), with source probability \(b\), maximizes \(E_up\) under the source budget. Indeed, \((p-p^*)(dP_u/dP_0-c)\le0\) pointwise and \(E_0(p-p^*)\le0\). For a Gaussian mean shift the likelihood ratio is increasing in \(u^\top X\), giving \(E_up\le U_\delta(b)\).

Use the reference with the alternative input marginal but loss probability identically \(\tau\). Its observed KL is
\[
I=E_u[p(X)\operatorname{kl}(a(X),\tau)]
\le\kappa_\tau U_\delta(b).
\]

For fixed $0<\epsilon<1$, apply \eqref{eq:F6} with
$m=\lfloor(1-\epsilon)\log A/\bar I\rfloor$ and
$u=\epsilon\log A/2$, where $\bar I=\kappa_\tau U_\delta(b)$.
The change-of-measure term is at most $mA^{-\epsilon/2}$, and the
maximal-deviation term also tends to zero. Since $D_u(T_A)\ge E_1T_A$,
letting $\epsilon\downarrow0$ proves \eqref{eq:F5}.
If $I=0$, the observed laws coincide and the SR/CUSUM ARL guarantee
in Theorem~\ref{thm:validity} gives $E_1T_A\ge A$.

For a finite set of directions \(u_1,\ldots,u_K\), put \(L_k=dP_k/dP_0\). Cauchy--Schwarz gives the additional input-allocation bound
\[
\min_kE_kp\le\min\{1,b+\sqrt{bV}\},\qquad
V=K^{-2}\sum_{j,k}\exp(\delta^2u_j^\top u_k)-1.
\]

For orthogonal directions \(V=(e^{\delta^2}-1)/K\). A worst-direction delay lower bound follows by using this allocation bound in the reference argument. Keeping \(K,\delta\), and the label model fixed makes these input allocation bounds tend to zero with \(b\). It does not prove an exact inverse-budget rate for all fixed strategies.

\FloatBarrier
\subsection{Finite-threshold bounds}

For a fixed observation rule and safe reference, let $\xi_t$ be
IID log-likelihood increments with mean $I\ge0$, variance at most $v$,
and $\xi_t-I\le B$. Put $\Lambda_k=\sum_{t=1}^k\xi_t$.
The scalar Freedman maximal inequality \citep[Theorem~1.1]{Tropp11Freedman}
gives
\[
P\!\left(\max_{k\le m}(\Lambda_k-kI)>u\right)
\le \exp\!\left\{-\frac{u^2}{2(mv+Bu/3)}\right\}.
\]
At the bounded stopping time $T\wedge m$, change of measure and
$Q(T\le m)\le m/A$ imply
\[
P(T\le m)\le \frac mA e^{mI+u}
 +P\!\left(\max_{k\le m}(\Lambda_k-kI)>u\right).
\]
Indeed, on $\{T\le m,\Lambda_T\le mI+u\}$ the stopped likelihood ratio
is at most $e^{mI+u}$, and its complement is covered by the maximal event.
Together with $E_1T\ge mP(T>m)$ this proves the generic bound used above.

For Theorem F.4's reference, $\xi_t=0$ when unaudited and
$\xi_t\le\kappa_\tau$. With $I=E\xi_t\ge0$, choose
$\bar I=\kappa_\tau U_\delta(b)$ and
$\bar v=(\kappa_\tau^2+8/e^2)U_\delta(b)$.
The conditional second-moment bound follows from
$x\log^2x\le4/e^2$ and the nonpositive cross terms in the squared
Bernoulli log ratios. Thus $I\le\bar I$, $\xi_t-I\le\kappa_\tau$ and
$\operatorname{Var}(\xi_t)\le\bar v$, giving
\begin{equation}
E_1T_A\ge m\left[
1-\frac mA e^{m\bar I+u}
-\exp\left\{-\frac{u^2}{2(m\bar v+\kappa_\tau u/3)}\right\}
\right]_+\tag{F.6}\label{eq:F6}
\end{equation}

for integer \(m\ge1\), \(u>0\).

For the single-component known-\(r\) oracle block policy, first align the blocks. Let block 0 be the first complete block starting at or after \(\nu\). At most \(w-1\) rounds precede it, and block 0 itself is spent before its summary sets \(H^C_1\). At most \(2w\) post-change rounds are discarded in total. Number the subsequent blocks \(j=1,2,\ldots\), and write \(\mathcal B_j\) for their rounds; every block from 0 onward consists of post-change rounds. (This relabels the aligned blocks; \(H^C_1\) below is the state of the first aligned block, not the initial state \(H^C_1=0\) of Section~\ref{sec:controllers}.)

After this alignment, define
\[
Y_j=H^C_j\sum_{t\in\mathcal B_j}\log G_t,\qquad j\ge1,
\]

where \(G_t\) is the known-\(r\) Kelly full-audit factor. Write \(\pi=P(H^C_j=1)\), which is constant for \(j\ge1\), and \(h=h(r)\). The state \(H^C_j\) depends only on block \(j-1\), independently of the losses in block \(j\), so \(EY_j=w\pi h\), including for \(j=1\). Each \(Y_j\) depends only on blocks \(j-1\) and \(j\). Thus \((Y_j)\) is one-dependent, and the odd and even subfamilies are each independent. Every \(Y_j\) lies in an interval of length \(wD_\ell\), where
\[
D_\ell=\log(r/\tau)-\log((1-r)/(1-\tau)).
\]

Split the deficit \(nw\pi h-z\) equally between the parity sums. Each has at most \((n+1)/2\) terms, so Hoeffding's inequality and a union bound give, when \(nw\pi h>z\),
\[
P\left\{\sum_{j=1}^nY_j<z\right\}
\le2\exp\left[-\frac{(nw\pi h-z)^2}{(n+1)w^2D_\ell^2}\right].
\]

Define \(\psi_n\) as this bound clipped at 1, or 1 when its condition fails; in particular \(\psi_0=1\). With \(z=\log A\), let
\[
N=\inf\left\{n\ge1:\sum_{j=1}^nY_j\ge z\right\}.
\]

Product domination gives an alarm by the end of block \(N\), so \((T_A-\nu+1)^+\le2w+wN\). Since \(\{N>n\}\) implies \(\sum_{j=1}^nY_j<z\), the tail sum gives \(EN\le\sum_{n\ge0}\psi_n\). The aligned post-change blocks are independent of the history at \(\nu-1\), so the same bound holds conditionally on that history. Therefore
\[
D_{\rm block}\le2w+w\sum_{n\ge0}\psi_n.
\]

\FloatBarrier
\FloatBarrier
\subsection{General input-dependent acquisition}\label{app:input-kernels}
For a general acquisition rule, a kernel \(p_t(\cdot)\) is selected from past information. After observing \(X_t\), an audit is drawn with probability \(p_t(X_t)\), independently of \(L_t\) conditional on \((\mathcal F_{t-1},X_t)\). With positive probabilities and a past-measurable predictor function \(m_t\), the augmented loss
\[
\widetilde L_t=m_t(X_t)+\frac{R_t}{p_t(X_t)}(L_t-m_t(X_t))
\]

has conditional mean \(E[L_t\mid\mathcal F_{t-1}]\). Thus \(1+\lambda_t(\widetilde L_t-\tau)\) is a valid factor whenever it is nonnegative and \(\lambda_t\ge0\) is chosen before the current input.

\section{Additional experimental results}
\setcounter{subsection}{-1}\subsection{Implementation limitations}\label{app:implementation-limits}

This subsection records implementation choices that affect how the
experimental results should be read. The main limitations of the paper
are listed in Section~\ref{sec:discussion}.

\paragraph{Controllers and betting grid.}
The original Income label controller rounds its increments to
\(+25,-7\) and caps the score at 1,000 (Appendix~\ref{app:income-design}). Before
capping, rounding moves its zero-drift risk from about 0.1954 to
\(7/32=0.21875>\tau\). All original Income targets have risks above 0.25; risks
between \(\tau\) and \(7/32\) are examined with an uncapped variant in
Appendix~\ref{sec:rounded-uncapped}. Both tasks use a fixed
eight-component grid instead of the growing grid of Lemma~\ref{lem:grid}.
At the three harmful pool risks its growth ratios \(g/h\) are
approximately 0.9986, 0.9999 and 0.9452, and at the label-flip risk
approximately 0.9777. Because the audit-evidence scale is $\log A/g$,
these ratios enter the first-order delay. F0 and F1 use
propensity-corrected factors with their own safe scales, so their
results combine acquisition and detector effects. The continuous
synthetic controllers are approximately calibrated; DEfull is the
exact-budget family.

\paragraph{Budgets.}
For the original capped L on the exact held-out source law, the
regenerative rates are approximately $1.004275\%$, $0.502281\%$ and
$0.201042\%$, slightly above the nominal targets, and
$E_0N_K\le b'K+666$ holds for every monitored prefix at that chain's
actual rate $b'$ (Appendix~\ref{app:acs-prefix}). C and Hybrid use
bootstrap-calibrated thresholds without such a certificate. Measured
source-prefix costs vary: L reaches 0.224\% at a nominal 0.2\% budget
in SD, and Hybrid's rates span 0.420\%--0.531\% at nominal 0.5\%
(Appendices~\ref{app:hybrid-second} and~\ref{app:source-prefix}).

\paragraph{Evaluation.}
Uniform resampling from the fixed, unweighted pools makes each side IID
under its pool's empirical law, and the label flips keep the complete
input law, so the unchanged-input condition holds exactly in that
experiment. Intervals are pointwise, condition on the fitted models,
pool selection and pre-change survival, and are not adjusted for
multiple comparisons. Interstate shifts may change both the input law
and the conditional label law. Each task has one benign target. The
original Income Hybrid is evaluated at one allocation and budget; the
capped H$_D$ and the TravelTime Hybrid cover three budgets with
different implementations.

\paragraph{Unfavorable outcomes.}
Several results count against our methods and are reported with the
corresponding experiments: heavy censoring of C under input-preserving
loss flips (Tables~\ref{tab:6} and~\ref{tab:tt-all}), near-full
labeling by C and Hybrid after benign shifts
(Appendices~\ref{app:benign-reset} and~\ref{app:traveltime}), heavy
censoring of the original controllers at small exceedances
(Appendix~\ref{sec:rounded-uncapped}), the failed attempt to align L
and C by empirical ARL (Appendix~\ref{app:income-benign}), and the
PublicCoverage study, which stopped before detection
(Appendix~\ref{app:coverage-feasibility}).

\FloatBarrier
\subsection{A matched-risk sensitivity check}\label{app:matched-risk}

At shift magnitude \(\delta=3\), the conditional-risk model
\[
a(x)\approx\sigma(-4.296+0.07973\|x\|^2)
\]

keeps source and target risks at 0.07 and 0.14. With 256 paths and no censoring, the known-direction static mean is 4,694 (SE 123), versus C at 1,374 (SE 35), giving a ratio of approximately 3.42. This checkpoint changes the conditional-risk shape while holding its two marginal risks fixed. The static policy knows the shift direction \(u\) and uses a fixed input-dependent audit probability
\[
p(x)=\begin{cases}1,&u^\top x\ge c,\\p_{\min},&u^\top x<c,\end{cases}
\qquad p_{\min}=0.0005,\quad b=0.01,\quad
c=\Phi^{-1}\!\left(1-\frac{b-p_{\min}}{1-p_{\min}}\right)\approx2.34534.
\]
Thus its nominal audit rate is \(p_{\min}+(1-p_{\min})P_0(u^\top X\ge c)=b\). This is the known-direction Neyman--Pearson region with an exploration floor, and its acquisition rule does not adapt to the observed history. The comparison uses SR at \(A=10^7\); the static policy uses the oracle factors of Appendix~\ref{app:oracle-factor} with a fitted, fixed betting parameter. The C comparator is the known-model norm-block controller in this checkpoint.

\FloatBarrier
\subsection{No-change SR/CUSUM counts}\label{app:nochange}

At \(A=10^4\) and 205,000 total rounds:

\begin{table}[ht]
\centering
\small
\caption{No-change alarms among 500 streams by round 205,000 at \(A=10^4\). Censoring prevents these counts from establishing actual ARL.}
\label{tab:11}
\setlength{\tabcolsep}{4pt}
\renewcommand{\arraystretch}{1.15}
\begin{tabular}{@{}lrrrr@{}}
\toprule
\textbf{Budget} & \textbf{SR L} & \textbf{SR C} & \textbf{CUSUM L} & \textbf{CUSUM C} \\ \midrule
1.0\% & 500/500 & 500/500 & 3/500 & 3/500 \\
0.5\% & 500/500 & 500/500 & 1/500 & 0/500 \\
0.2\% & 500/500 & 500/500 & 0/500 & 0/500 \\
\bottomrule
\end{tabular}
\end{table}

With factor 1, SR adds one unit each round, whereas CUSUM returns to or stays above its unit baseline without repeated addition. The skipped-round update therefore affects finite-horizon false-alarm behavior.

\FloatBarrier
\subsection{Benign-shift cost recovery}\label{app:benign-reset}

For DC, CUSUM at \(A=10^6\), the full post-change and late-window acquisition costs are:

\begin{table}[ht]
\centering
\small
\caption{Benign-shift acquisition in DC at \(A=10^6\). The final window and whole post-change window show the initial cost of resetting.}
\label{tab:12}
\setlength{\tabcolsep}{4pt}
\renewcommand{\arraystretch}{1.15}
\begin{tabular}{@{}lrrrr@{}}
\toprule
\textbf{Budget} & \textbf{C whole} & \textbf{Reset whole} & \textbf{Reset final} & \textbf{Reset alarms} \\ \midrule
1.0\% & 99.871\% & 2.174\% & 0.702\% & 0/500 \\
0.5\% & 99.850\% & 1.887\% & 0.400\% & 0/500 \\
0.2\% & 99.786\% & 1.765\% & 0.151\% & 0/500 \\
\bottomrule
\end{tabular}
\end{table}

For reverse-null confirmation, high-state factors are \(1+\eta_j(\tau-L_t)\), \(0\le\eta_j\le1/(1-\tau)\); low state uses factor 1. An attempt has error allowance \(\delta_k=0.01/[k(k+1)]\), and the implementation allows at most three resets. Reset requires a completed block and at least \(\max(2000,8w)\) high-state labels. The new mean/covariance reference is estimated from the high-state observations with declared ridge and threshold inflation; main alarm capital is not reset.

The confirmation guarantee applies to a reverse null that holds from each attempt's declared start: it requires risk at least \(\tau\) throughout the attempt, so wealth accumulated before a later harmful change gives no automatic post-change error bound. The reset also departs from the fixed-reference model of Theorem~\ref{thm:upper}.

\begin{figure}[H]
\centering
\includegraphics[width=\linewidth]{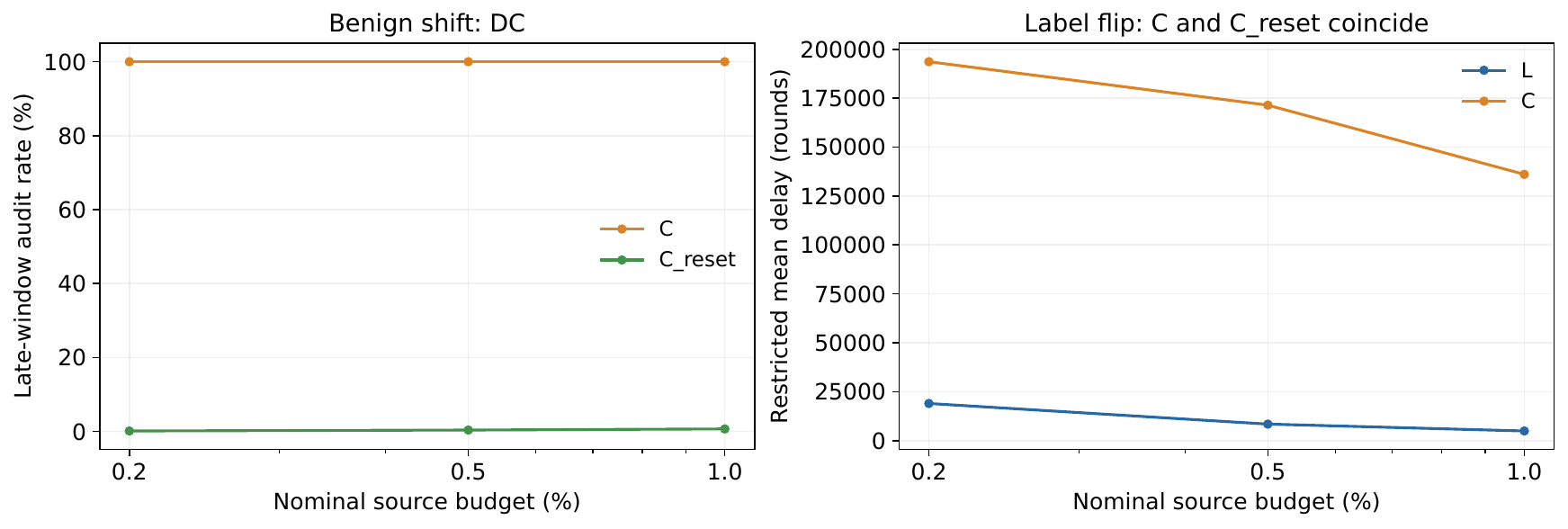}
\caption{CUSUM benign-shift cost recovery and the unchanged-input label-flip stress test at \(A=10^6\). C and C\_reset have identical plotted delay values in the right panel, so that panel shows C once. Tables report whole-horizon acquisition costs and censoring.}
\label{fig:4}
\end{figure}

\FloatBarrier
\subsection{Hybrid at the second threshold and measured budget}\label{app:hybrid-second}

The lower-threshold paired ratios are:

\begin{table}[ht]
\centering
\small
\caption{Hybrid supplement at the second threshold \(A=10^4\), budget 0.5\%. All 500 pairs survive the pre-change period.}
\label{tab:13}
\setlength{\tabcolsep}{4pt}
\renewcommand{\arraystretch}{1.15}
\begin{tabular}{@{}llrr@{}}
\toprule
\textbf{Scenario} & \textbf{Comparison} & \textbf{Ratio [95\% CI], \(A=10^4\)} & \textbf{Pairs} \\ \midrule
SD & Hybrid/C & 1.026 [1.013, 1.041] & 500 \\
WV & Hybrid/C & 1.020 [1.006, 1.036] & 500 \\
MT & Hybrid/C & 1.023 [1.014, 1.034] & 500 \\
Label flip & Hybrid/L & 1.919 [1.720, 2.134] & 500 \\
\bottomrule
\end{tabular}
\end{table}

Across the two thresholds, the six harmful-shift Hybrid/C point ratios range from 1.017 to 1.026, with 95\% interval endpoints spanning 1.006--1.041. In the two label-flip cells, the Hybrid/L ratio is 1.919; its upper 95\% endpoints are 2.134 and 2.125, and Hybrid has no censored streams. These results come from the targeted Hybrid supplement, which set an upper ratio endpoint of at most 2.5 and a Hybrid censoring rate of at most 1\% under label flips as benchmarks for the anticipated twofold cost.

\begin{table}[ht]
\centering
\small
\caption{Measured Hybrid source-prefix acquisition rates and Monte Carlo standard errors. The diagnostic flag threshold is 0.55\%.}
\label{tab:14}
\setlength{\tabcolsep}{4pt}
\renewcommand{\arraystretch}{1.15}
\begin{tabular}{@{}lrrc@{}}
\toprule
\textbf{Scenario} & \textbf{Pre-change rate} & \textbf{MC SE (pp)} & \textbf{Above 0.55\%} \\ \midrule
SD & 0.420\% & 0.056 & False \\
WV & 0.520\% & 0.064 & False \\
MT & 0.434\% & 0.056 & False \\
Label flip & 0.531\% & 0.071 & False \\
\bottomrule
\end{tabular}
\end{table}

The flag threshold is 110\% of the nominal 0.5\% budget. The measured rates summarize finite evaluation prefixes; the source-bootstrap and source-chain calculations are separate calibration diagnostics.

\FloatBarrier
\subsection{Source-prefix cost diagnostics}\label{app:source-prefix}

For L in SD at \(b=0.002\), the measured source-prefix rate is 0.224\%, with Monte Carlo SE about 0.023 percentage points. The analytic long-run rate on the nominal source is approximately 0.201\%; the two quantities refer to different time scales and source distributions.

\FloatBarrier
\subsection{All five implemented ACS policies at one budget}\label{app:five-policies}

For completeness, the CUSUM results at \(b=0.005,A=10^6\) include both fixed-audit controls and the full-audit reference:

\begin{table}[ht]
\centering
\small
\caption{Five ACS policies at budget 0.5\% and \(A=10^6\). The censoring horizon is 200,000 post-change rounds.}
\label{tab:15}
\setlength{\tabcolsep}{4pt}
\renewcommand{\arraystretch}{1.15}
\begin{tabular}{@{}llrrrr@{}}
\toprule
\textbf{Target} & \textbf{Policy} & \textbf{Restricted delay} & \textbf{Early alarms} & \textbf{Censored} & \textbf{Pre-change rate} \\ \midrule
SD & F0 & 147,603.1 & 0 & 120 & 0.4906\% \\
SD & F1 & 200,000.0 & 0 & 500 & 0.4920\% \\
SD & L & 6,800.4 & 0 & 0 & 0.5414\% \\
SD & C & 1,225.1 & 0 & 0 & 0.4086\% \\
SD & Full & 878.2 & 0 & 0 & 100.0000\% \\
WV & F0 & 141,809.6 & 0 & 102 & 0.5062\% \\
WV & F1 & 200,000.0 & 0 & 500 & 0.4895\% \\
WV & L & 7,263.6 & 0 & 0 & 0.4757\% \\
WV & C & 1,130.9 & 0 & 0 & 0.4432\% \\
WV & Full & 794.1 & 1 & 0 & 100.0000\% \\
MT & F0 & 192,760.1 & 0 & 424 & 0.5038\% \\
MT & F1 & 200,000.0 & 0 & 500 & 0.4870\% \\
MT & L & 13,099.1 & 0 & 0 & 0.4736\% \\
MT & C & 2,118.4 & 0 & 0 & 0.4666\% \\
MT & Full & 1,777.4 & 0 & 0 & 100.0000\% \\
\bottomrule
\end{tabular}
\end{table}

A censored stream contributes the full 200,000-round post-change horizon. F0 and F1 use the augmented-loss factors and globally safe scales of Appendix~\ref{app:income-design}, whereas L, C and Full use direct full-audit factors, so the F0/F1 results also reflect their betting scales.

\FloatBarrier
\subsection{SR harmful-shift comparison}

The SR detector at \(A=10^6\) gives the following paired L/C ratios on the same harmful targets. The budget dependence matches CUSUM; the values differ because the recursion differs.

\begin{table}[ht]
\centering
\small
\caption{ACS harmful-shift delay ratios for the SR detector at \(A=10^6\). Brackets give descriptive paired 95\% bootstrap intervals.}
\label{tab:16}
\setlength{\tabcolsep}{4pt}
\renewcommand{\arraystretch}{1.15}
\begin{tabular}{@{}lrrr@{}}
\toprule
\textbf{Budget} & \textbf{SD: L/C} & \textbf{WV: L/C} & \textbf{MT: L/C} \\ \midrule
1.0\% & 5.653 [5.175, 6.136] & 5.127 [4.692, 5.572] & 5.564 [5.111, 6.044] \\
0.5\% & 8.262 [7.580, 8.979] & 9.610 [8.674, 10.595] & 10.412 [9.485, 11.355] \\
0.2\% & 19.350 [17.610, 21.198] & 19.484 [17.704, 21.440] & 22.169 [20.200, 24.174] \\
\bottomrule
\end{tabular}
\end{table}

\FloatBarrier
\subsection{Nearest-rounded, uncapped controller sensitivity}
\label{sec:rounded-uncapped}

This subsequent implementation-sensitivity study retains the ACS pools,
fitted models, risk threshold, eight-component CUSUM detector, source
warmup and 5,000 monitored pre-change rounds. It compares the original
controller L with a nearest-rounded, uncapped variant, denoted L$_N$,
and the corresponding half-budget Hybrid H$_N$. We added this study after the initial protocol.

L$_N$ replaces increments $+25,-7$ by $+26,-6$ in units of 0.01 and
removes the positive score cap. Its mean integer increment at risk $r$
is $32r-6$, with zero-drift point $3/16=0.1875<\tau$.
The design-baseline excursion mean is approximately 10.77043,
compared with 4.21745 for the original implemented controller.
Sleep lengths are calibrated from this excursion mean. A finite-state
occupation approximation initializes the score (states through 3,000),
followed by the existing warmup; the runtime score is uncapped.
The two added controllers use separate hash-based initialization draws
that do not consume the archived shared random-number stream.

\paragraph{Adjustment coefficient and fallback.}
The rounded score is not the exact log-likelihood score of Section~\ref{sec:controllers},
but an adjustment coefficient supplies its escape bound. For
$Y_N=0.26L-0.06(1-L)$, put
\[
M_r(\theta)=r e^{-0.26\theta}+(1-r)e^{0.06\theta}.
\]
For $3/16<r<1$, $M_r(0)=1$, $M'_r(0)=0.06-0.32r<0$,
$M_r$ is strictly convex, and $M_r(\theta)\to\infty$.
Hence there is a unique positive root $\theta_*(r)$ of
$M_r(\theta)=1$. On the active-label clock, let $Z_n$ be the unreset score walk
within an excursion. Then $e^{-\theta_*(r) Z_n}$ is a nonnegative martingale.
The maximal inequality bounds the probability of ever entering the
negative half-line from score $x\ge0$ by $e^{-\theta_*(r)x}$.
A first positive label reaches score 0.26, giving
\[
p_N(r):=r\{1-e^{-0.26\theta_*(r)}\}\le p_{\mathrm{esc},N}(r),
\qquad E[\Sigma_N\mid\mathcal F_{\nu-1}]\le\frac{W_N}{p_N(r)}.
\]
The second inequality follows by geometric domination of failed
excursions and an initial sleep residual of at most $W_N$, uniformly
over nonnegative starting scores. At $r=1$, escape is certain and
remaining sleep is at most $W_N$; no finite positive equality root is
needed. Since $M_r(\theta)$ decreases in $r$, the fixed coefficient
$\theta_*(\tau)$ gives a uniform positive escape bound
$\tau\{1-e^{-0.26\theta_*(\tau)}\}$ for $r\ge\tau$.
Numerically, $\theta_*(\tau)\approx0.552206$ and
$p_N(\tau)\approx0.026909$; at $r=0.21$,
$\theta_*(r)\approx0.897234$ and $p_N(r)\approx0.043694$.
The algorithm itself does not use these coefficients.

Controller escape holds for $r>3/16$. A risk-alarm delay bound further
requires $r>\tau$ and a betting component with positive audit log-growth
$g$. Under the IID change model, Lemma~\ref{lem:audit}'s audit-clock argument gives
for an equal mixture of $J$ components
\[
D_{H_N}\le\frac{\log(JA)+\ell_+}{g}+\frac{W_N}{p_N(r)}.
\]
The growing grid on the fixed alternative class of Lemma~\ref{lem:grid} can be
used as before, with this new waiting constant. The implemented fixed
eight-component grid need not have positive growth arbitrarily close
to $\tau$, and for $3/16<r\le\tau$ there is no alarm guarantee.

\paragraph{Nominal prefix budget and initialization.}
For this rounded variant assume $r_{\mathrm{src}}\le r_0<3/16$, and
write $d_N=0.06-0.32r_0>0$. At bounded stops of the zero-start source
excursion, the score is at least $-0.06$, so Wald's identity gives
$e_N\le0.06/d_N<\infty$. Monotone coupling bounds excursions at
lower source risks by $e_N$. With
$W_N=\lceil e_N(1-b_L)/b_L\rceil$, the renewal argument of Appendix~\ref{app:controller}
therefore gives $E_0N_K\le b_LK+e_N$ from a fresh active zero start.
For an initial active score $X_0\ge0$ with finite first moment, let
$I_{\mathrm{active}}$ indicate an active initial state. Separating
the initial excursion gives the sufficient allowance
\[
c_N=e_N+\frac{E[I_{\mathrm{active}}(X_0+0.06)]}{d_N}.
\]
A common allowance across a policy family requires this initial-moment
bound uniformly bounded. Initial sleeping consumes no labels. Alarm-factor validity is
unaffected by predictable acquisition.

The design source risk is $r_0=0.17120225$, whereas the empirical source
pool risk is approximately 0.17146654. The latter yields excursion mean
10.94685 and slightly larger long-run audit rates
(Table~\ref{tab:fix-budget}); the sensitivity study keeps the design
calibration.

\begin{table}[ht]
\centering
\small
\caption{Sleep lengths and long-run label-channel audit rates for L$_N$, computed from the renewal argument rather than measured.}
\label{tab:fix-budget}
\setlength{\tabcolsep}{3pt}
\renewcommand{\arraystretch}{1.15}
\begin{tabular}{@{}lrrr@{}}
\toprule
\textbf{Nominal channel budget} & \textbf{Sleep} & \textbf{Design-risk rate} & \textbf{Source-pool rate} \\ \midrule
1.0\% & 1,067 & 0.9993\% & 1.0155\% \\
0.5\% & 2,144 & 0.4998\% & 0.5080\% \\
0.2\% & 5,375 & 0.2000\% & 0.2032\% \\
0.25\% (Hybrid channel) & 4,298 & 0.2500\% & 0.2540\% \\
\bottomrule
\end{tabular}
\end{table}

\paragraph{Archive checks and analysis.}
The supplied C++ port matches archived baseline outputs in 24
policy--scenario--budget comparisons, each with 500 streams.
All 14 cells were then regenerated independently by extending the
original Python/numba Hybrid kernel with L$_N$ and H$_N$, without
running the C++ simulation, using the existing pools, seeds, parameters
and separately reconstructed initialization CDFs (software versions in
Appendix~\ref{app:experimental-record}). All 35,000 policy--stream records match the supplied
C++ results in both alarm times, pre-change label counts and label
counts at the $A=10^6$ alarm (the stored zero sentinel is retained for
censoring). A pilot also checked the extension against the original
three-policy numba kernel. Settings were frozen before regeneration,
although the C++ results were already known.
Means, censoring and paired intervals were independently recomputed.
Each evaluated cell has 500 paired streams and no pre-change alarms.
Intervals use 2,000 paired bootstrap resamples, conditional on the
fixed pipeline and evaluated cells, without multiplicity adjustment.

\paragraph{Small exceedances.}
At nominal budget 0.5\%, post-change inputs are resampled from the
source pool and each loss is flipped with probability $q$. Thus
$r=q+(1-2q)r_{\mathrm{src,eval}}$. The target risks 0.210, 0.215,
0.230 and 0.250 use $q$ approximately 0.05864, 0.06625, 0.08908 and
0.11952; the last cell retains the original $q=0.15$ and has risk
approximately 0.27003. The same indexed input draws and hash-based flip
uniforms are shared across the scan. Flip indicators are nested in
$q$, but the resulting binary losses need not be monotone in $q$.

\begin{table}[ht]
\centering
\small
\caption{Input-preserving loss-flip scan at $A=10^6$, nominal budget 0.5\%, and post-change horizon $H=10^6$. Entries are restricted mean delays in rounds, with censoring counts out of 500 in parentheses.}
\label{tab:fix-scan}
\setlength{\tabcolsep}{3pt}
\renewcommand{\arraystretch}{1.15}
\begin{tabular}{@{}lrrrrr@{}}
\toprule
\textbf{Risk} & \textbf{L} & \textbf{L$_N$} & \textbf{Hybrid} & \textbf{H$_N$} & \textbf{C} \\ \midrule
0.210 & 977,023 (472) & 103,511 (0) & 989,544 (483) & 146,702 (0) & 999,181 (499) \\
0.215 & 585,064 (107) & 61,774 (0) & 765,428 (251) & 95,387 (0) & 998,892 (497) \\
0.230 & 41,367 (0) & 28,456 (0) & 74,331 (0) & 50,348 (0) & 914,112 (353) \\
0.250 & 14,451 (0) & 16,536 (0) & 25,553 (0) & 34,360 (0) & 522,283 (31) \\
0.27003 & 8,417 (0) & 13,068 (0) & 16,151 (0) & 23,359 (0) & 288,842 (4) \\
\bottomrule
\end{tabular}
\end{table}

\begin{table}[ht]
\centering
\small
\caption{The two small-exceedance cells restricted to the original post-change horizon $H=200{,}000$, using the same stored runs. Entries are restricted means (censored streams out of 500).}
\label{tab:fix-short}
\setlength{\tabcolsep}{3pt}
\renewcommand{\arraystretch}{1.15}
\begin{tabular}{@{}lrrrrr@{}}
\toprule
\textbf{Risk} & \textbf{L} & \textbf{L$_N$} & \textbf{Hybrid} & \textbf{H$_N$} & \textbf{C} \\ \midrule
0.210 & 199,377 (496) & 101,852 (18) & 199,853 (498) & 128,025 (102) & 200,000 (500) \\
0.215 & 188,880 (432) & 61,503 (3) & 194,726 (466) & 91,068 (33) & 200,000 (500) \\
\bottomrule
\end{tabular}
\end{table}

\begin{table}[H]
\centering
\small
\caption{Paired ratios of restricted mean delays for the scan at $H=10^6$, with pointwise 95\% bootstrap intervals. Heavy censoring in the original controllers limits interpretation as uncensored mean-delay ratios.}
\label{tab:fix-scan-ratios}
\setlength{\tabcolsep}{3pt}
\renewcommand{\arraystretch}{1.15}
\begin{tabular}{@{}lrr@{}}
\toprule
\textbf{Risk} & \textbf{L/L$_N$} & \textbf{H$_N$/L$_N$} \\ \midrule
0.210 & 9.44 [9.04, 9.84] & 1.42 [1.33, 1.51] \\
0.215 & 9.47 [8.80, 10.15] & 1.54 [1.42, 1.67] \\
0.230 & 1.45 [1.31, 1.61] & 1.77 [1.59, 1.97] \\
0.250 & 0.87 [0.79, 0.97] & 2.08 [1.83, 2.35] \\
0.27003 & 0.64 [0.57, 0.72] & 1.79 [1.58, 2.03] \\
\bottomrule
\end{tabular}
\end{table}

At the two evaluated risks below $7/32$, L$_N$ and H$_N$ have no
censored streams by $10^6$ rounds, unlike the original controllers.
At the original $200{,}000$-round horizon, censoring remains:
18 and 3 streams for L$_N$, and 102 and 33 for H$_N$.
The scan shows better finite-horizon detection at these two points; it
does not cover the whole interval $(\tau,7/32)$. The original L can
still alarm when its controller drift is negative, because the risk
detector keeps evidence across excursions.

\FloatBarrier
\paragraph{Interstate shifts.}
L$_N$ has larger restricted mean delay than L in all nine evaluated
interstate cells, with ratios approximately 1.22--1.80.

\begin{table}[H]
\centering
\small
\caption{Interstate sensitivity at $A=10^6$ and $H=200{,}000$. Delays are restricted means over 500 paired streams. Brackets give pointwise 95\% paired bootstrap intervals. L and C are uncensored; L$_N$ has 2/500 censored streams in WV and MT at budget 0.2\% and none elsewhere.}
\label{tab:fix-states}
\setlength{\tabcolsep}{3pt}
\renewcommand{\arraystretch}{1.15}
\begin{tabularx}{\linewidth}{@{}llrrrYY@{}}
\toprule
\textbf{State} & \textbf{Budget} & \textbf{L} & \textbf{L$_N$} & \textbf{C} & \textbf{L$_N$/L} & \textbf{L$_N$/C} \\ \midrule
SD & 1.0\% & 4,260 & 6,676 & 1,156 & 1.57 [1.41, 1.74] & 5.78 [5.34, 6.22] \\
WV & 1.0\% & 3,728 & 6,450 & 1,063 & 1.73 [1.56, 1.92] & 6.07 [5.59, 6.54] \\
MT & 1.0\% & 6,937 & 8,831 & 2,030 & 1.27 [1.16, 1.40] & 4.35 [4.02, 4.71] \\
SD & 0.5\% & 6,800 & 11,312 & 1,225 & 1.66 [1.49, 1.87] & 9.23 [8.42, 10.07] \\
WV & 0.5\% & 7,264 & 11,082 & 1,131 & 1.53 [1.36, 1.71] & 9.80 [8.99, 10.69] \\
MT & 0.5\% & 13,099 & 16,042 & 2,118 & 1.22 [1.09, 1.37] & 7.57 [6.88, 8.26] \\
SD & 0.2\% & 16,706 & 29,996 & 1,293 & 1.80 [1.60, 2.00] & 23.20 [21.32, 25.14] \\
WV & 0.2\% & 16,481 & 28,567 & 1,245 & 1.73 [1.54, 1.96] & 22.94 [20.85, 25.22] \\
MT & 0.2\% & 29,039 & 37,970 & 2,230 & 1.31 [1.17, 1.46] & 17.03 [15.56, 18.54] \\
\bottomrule
\end{tabularx}
\end{table}

\begin{table}[ht]
\centering
\small
\caption{Hybrid sensitivity at nominal budget 0.5\%, $A=10^6$ and $H=200{,}000$. All 500 streams per cell are uncensored. Brackets give pointwise 95\% paired bootstrap intervals.}
\label{tab:fix-hybrid}
\setlength{\tabcolsep}{3pt}
\renewcommand{\arraystretch}{1.15}
\begin{tabular}{@{}lrrrr@{}}
\toprule
\textbf{State} & \textbf{C} & \textbf{Hybrid} & \textbf{H$_N$} & \textbf{H$_N$/C} \\ \midrule
SD & 1,225.1 & 1,253.0 & 1,257.2 & 1.026 [1.015, 1.039] \\
WV & 1,130.9 & 1,154.1 & 1,160.8 & 1.027 [1.016, 1.039] \\
MT & 2,118.4 & 2,153.6 & 2,148.3 & 1.014 [1.006, 1.023] \\
\bottomrule
\end{tabular}
\end{table}

We keep the original L in the main tables; using L$_N$ would only
enlarge the reported L/C ratios in these cells. Because L$_N$ changes
rounding, capping, initialization and sleep at once, we do not
attribute the difference to any single change. H$_N$/C point ratios
remain 1.014--1.027 on the three interstate cells. Under the $q=0.15$
label perturbation, H$_N$ is slower than the original Hybrid
(restricted means 23,359 versus 16,151 rounds).

\FloatBarrier
\subsection{Shared-detector acquisition controls and administrative censoring}
\label{app:shared-baselines}

This later supplement fixed its own protocol before its new detection
runs; the earlier ACS results were already known. It retains
the source and target pools, model, threshold, eight betting components,
source warmup, 5,000 monitored pre-change rounds and 200,000-round
post-change horizon. It runs six scenarios at three budgets, with 500
paired streams per cell and thresholds $10^4,10^6$. The primary threshold
is $10^6$. All original L/C/Full alarms and counts reproduce the archived
streams; uniform requests reproduce the original F0 requests. The latter
now update the common direct-loss CUSUM, without F0's augmented-loss scale.

The three label-history controls use design risks
$r_d=\tau,\tau+0.02,\tau+0.06$. Their integer increments are
$\lfloor\log(r_d/r_0)/0.01\rfloor$ and
$-\lceil\log((1-r_0)/(1-r_d))/0.01\rceil$, with positive cap 1000.
After a negative score, they sleep a fixed number of rounds and restart
at zero. These DE-inspired acquisition rules feed the common risk
alarm; they do not reproduce the original DE-CuSum alarm.

For these comparators only, the exact finite source pool's error
fraction calibrates the finite-state excursion mean and stationary
initialization. Sleep is rounded upward to meet the channel's nominal
rate. Access to the source law favors the comparators. Warmup label
costs are recorded separately. C keeps its original full-channel
bootstrap calibration.

\begin{table}[H]
\centering\small
\caption{Strengthened label-history controls at $A=10^6$. Entries are restricted mean delays. ``Best/C'' is the post hoc minimum mean across the three fixed designs divided by C's mean, an oracle envelope without a selected-baseline confidence interval.}
\label{tab:de-shared}
\setlength{\tabcolsep}{3pt}
\begin{tabularx}{\linewidth}{@{}llrrrrY@{}}
\toprule
Budget & Target & C & DE$_\tau$ & DE$_{\tau+.02}$ & DE$_{\tau+.06}$ & Best/C \\ \midrule
1\% & SD & 1,155.5 & 4,655.4 & 4,093.2 & 3,971.9 & 3.437 \\
1\% & WV & 1,062.9 & 4,315.8 & 3,673.4 & 3,792.4 & 3.456 \\
1\% & MT & 2,029.7 & 7,437.6 & 7,255.2 & 7,379.4 & 3.574 \\
0.5\% & SD & 1,225.1 & 9,122.0 & 7,009.5 & 7,027.8 & 5.722 \\
0.5\% & WV & 1,130.9 & 8,288.4 & 6,719.6 & 6,644.4 & 5.876 \\
0.5\% & MT & 2,118.4 & 13,438.3 & 11,978.5 & 12,407.5 & 5.655 \\
0.2\% & SD & 1,293.1 & 19,789.6 & 16,255.3 & 19,021.0 & 12.571 \\
0.2\% & WV & 1,245.1 & 18,887.6 & 16,494.3 & 16,081.5 & 12.916 \\
0.2\% & MT & 2,230.0 & 30,380.9 & 27,091.8 & 31,040.5 & 12.149 \\
\bottomrule\end{tabularx}
\end{table}

\begin{table}[ht]
\centering\small
\caption{Uniform acquisition with the common CUSUM at $A=10^6$. Delays are restricted means through 200,000 post-change rounds, conditional on pre-change survival. C has no censored streams in these cells. NR means that the Kaplan--Meier median is not reached within the horizon.}
\label{tab:uniform-common}
\setlength{\tabcolsep}{3pt}
\begin{tabular}{@{}llrrrr@{}}
\toprule
Budget & Target & Uniform delay & C delay & Uniform censored & Uniform median \\ \midrule
1\% & SD & 82,920.0 & 1,155.5 & 1/500 (0.2\%) & 80,110 \\
1\% & WV & 80,486.1 & 1,062.9 & 0/500 (0.0\%) & 77,860 \\
1\% & MT & 161,889.8 & 2,029.7 & 190/500 (38.0\%) & 178,865 \\
0.5\% & SD & 155,557.0 & 1,225.1 & 152/500 (30.4\%) & 164,202 \\
0.5\% & WV & 150,605.6 & 1,130.9 & 130/500 (26.0\%) & 153,936 \\
0.5\% & MT & 194,496.5 & 2,118.4 & 432/500 (86.4\%) & $\mathrm{NR}$ \\
0.2\% & SD & 195,950.2 & 1,293.1 & 460/500 (92.0\%) & $\mathrm{NR}$ \\
0.2\% & WV & 196,861.5 & 1,245.1 & 465/500 (93.0\%) & $\mathrm{NR}$ \\
0.2\% & MT & 199,702.4 & 2,230.0 & 497/500 (99.4\%) & $\mathrm{NR}$ \\
\bottomrule\end{tabular}
\end{table}

Censoring is administrative at one common horizon within each cell.
The area under the Kaplan--Meier survival estimate through that horizon
therefore equals the empirical restricted mean used in the tables;
all 27 policy--scenario--budget calculations for Uniform, DE$_\tau$
and C agree numerically. The unobserved tail is not estimated, and
Uniform's median is not reached in four of the nine harmful cells.

\begin{figure}[H]
\centering
\includegraphics[width=\linewidth]{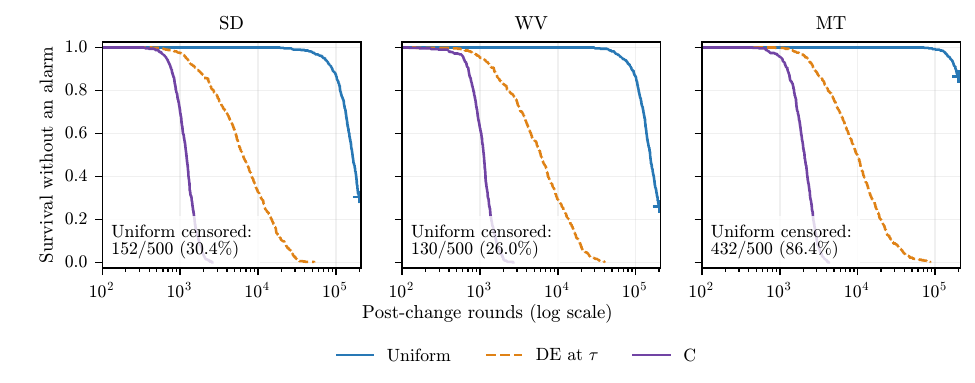}
\caption{Kaplan--Meier post-change survival curves at budget 0.5\% and $A=10^6$, conditional on each policy's pre-change survival. Curves compare the fixed DE$_\tau$ design, Uniform and C. A plus at the horizon marks remaining administrative censoring. The horizontal axis is logarithmic.}
\label{fig:uniform-km}
\end{figure}

H$_D$ combines half-budget DE$_\tau$ and half-budget C, updates its
label score only on its own requests, and sends the union to one common
CUSUM. Its source thresholds are calibrated before the extension's
detection outcomes. Its harmful-target H$_D$/C mean ratios are
0.995--1.024. Under input-preserving label perturbations:

\begin{table}[ht]
\centering\small
\caption{The new capped H$_D$ allocation check at $A=10^6$. Entries are ratios of restricted means; this is distinct from original Hybrid and uncapped H$_N$.}
\label{tab:hybrid-de-three}
\begin{tabular}{@{}lrrrr@{}}
\toprule
Budget & DE$_\tau$ delay & H$_D$ delay & H$_D$/DE$_\tau$ & H$_D$ censored \\ \midrule
1\% & 5,272.9 & 9,614.0 & 1.823 & 0/500 \\
0.5\% & 9,652.7 & 17,597.2 & 1.823 & 0/500 \\
0.2\% & 22,424.2 & 41,934.2 & 1.870 & 4/500 \\
\bottomrule\end{tabular}
\end{table}

These ratios are consistent with the larger waiting cost of a
half-budget label channel. All cells, both thresholds,
source rates, early alarms and censoring are in the supplementary
records.

\paragraph{H$_D$/C ratios slightly below one.}
\label{app:hd-small-gain}
At budget 1\% and $A=10^6$, all three targets have 500 paired
pre-change survivors and no administrative censoring. Table~\ref{tab:hd-small-gain}
uses the extension's existing 2,000-replicate paired bootstrap
intervals; none were reselected or multiplicity-adjusted.

\begin{table}[ht]
\centering\small
\caption{Small finite-path differences for Income H$_D$ versus C.
The interval estimates a ratio of paired mean delays, not a mean
of per-path ratios. Faster/equal/slower counts compare paired alarm
times. Extra labels are H$_D$-only changed labels before C's first
changed-label request, averaged over all 500 paths.}
\label{tab:hd-small-gain}
\setlength{\tabcolsep}{4pt}
\begin{tabular}{@{}lrrcr@{}}
\toprule
Target & H$_D$/C & Pointwise 95\% interval & Faster/equal/slower & Extra labels \\ \midrule
SD & 0.99553 & [0.99219,0.99836] & 16/475/9 & 5.146 \\
WV & 0.99504 & [0.98909,1.00058] & 20/460/20 & 7.716 \\
MT & 0.99904 & [0.99694,1.00102] & 8/479/13 & 4.728 \\
\bottomrule\end{tabular}
\end{table}

An instrumented replay on exactly these archived seeds reproduced
all seven original output fields for eight policies and both thresholds,
on all 1,500 paths. It generated no new independent streams and changed
no parameters. H$_D$ requested before C's first changed request on
104/500 SD, 95/500 WV and 103/500 MT paths. Its extra label factors
had positive mean accumulated log growth for the grid fractions 0.05
and 0.10, a plausible source of early evidence. Most paired alarm
times were nevertheless identical. Only SD's unadjusted interval lies
wholly below one; WV and MT are compatible with no mean difference.

H$_D$ need not dominate C path by path: its half-budget C threshold is
higher than C's full-budget threshold, its pre-change alarm capital
differs because its source requests differ, it can miss C-only
requests, and extra negative loss factors can reduce its capital. The
replay is a mechanism check and does not isolate the effect of the
extra labels. The replay, per-path diagnostics, original hashes and
audit code are included in the source package.

\FloatBarrier
\subsection{An attempted second-task extension stopped before detection}
\label{app:coverage-feasibility}

The same extension protocol attempted ACSPublicCoverage, using 2018
California as source. The source classifier was fitted once, with
138,554 filtered source records and disjoint training/calibration
splits. Each of the other 50 targets was divided into scenario-selection
and final-evaluation halves before selection. The protocol required
selection risk at least $\tau+0.02$ and a whitened input mean-shift score
at least the candidate median; its thresholds were not relaxed.

Source risk calibration gave $r_0=0.268856$ and $\tau=0.298856$.
No target passed the harmful-selection rule: the largest candidate
selection risk was 0.314353, below $\tau+0.02=0.318856$.
Separately, the fixed quantization produced increments $+10,-5$ for
the DE$_\tau$ design, giving zero drift at $1/3>\tau$ and mean integer
increment approximately $-0.51716$ at $\tau$. This failed the protocol's
pre-detection positive-drift check. All three full-channel C calibration
checks had passed, but no PublicCoverage detector was run.

The source model, splits, selection results and failed check are
retained. A source-only finer-precision repair is recorded as an
unexecuted draft; it would not create a qualifying harmful target.

\FloatBarrier
\subsection{Synthetic and Income experiments: full record}

\label{app:original-experimental-record}

We evaluate the waiting-cost mechanism on known-model synthetic data and ACS Income. The synthetic study varies the label budget while holding the change model fixed. The ACS study trains all learned components on the source domain and compares harmful interstate shifts with a constructed label perturbation on unchanged inputs. We use CUSUM for the main ACS comparisons; Appendix~\ref{app:implementation-limits} and the following subsections report SR and additional diagnostics; reproduction details are in Appendix~\ref{app:repro}.

\subsubsection{Synthetic budget dependence}
\label{app:original-synthetic}

The synthetic model has dimension \(d=20\), source risk 0.07, risk threshold \(\tau=0.10\), and post-change risk 0.14 at shift magnitude \(\delta\approx2.223\). We use four fixed directions, 128 paths per direction and budget, and change time 1. The budget comparison uses covariate-triggered blocks, label-history controllers, and a full-audit reference. Appendix~\ref{app:fixed-audit} gives the known-model betting factors and fixed-rule analysis.

Table~\ref{tab:2} compares the covariate controller with the fastest of 12 evaluated label-controller settings and with the exact-budget finite-state DEfull subfamily. At \(A=10^7\) and target trigger power 0.99, the covariate delay remains near 1,300 rounds as the budget falls, while the fastest label-only mean grows from 4,692 to 20,601 rounds.

\begin{table}[ht]
\centering
\small
\caption{Synthetic delays with target trigger power 0.99. All-12 selects the fastest evaluated label controller; DEfull restricts the comparison to the exact-budget subfamily.}
\label{tab:2}
\setlength{\tabcolsep}{4pt}
\renewcommand{\arraystretch}{1.15}
\begin{tabular}{@{}lrrrrr@{}}
\toprule
\textbf{Budget} & \textbf{Best label (12)} & \textbf{C} & \textbf{Label/C} & \textbf{Best DEfull} & \textbf{DEfull/C} \\ \midrule
1\% & 4,692 & 1,385 & 3.39 & 4,944 & 3.57 \\
0.5\% & 8,457 & 1,308 & 6.46 & 8,785 & 6.71 \\
0.2\% & 20,601 & 1,366 & 15.08 & 21,243 & 15.55 \\
\bottomrule
\end{tabular}
\end{table}

DEfull allocates its full budget to active excursions and uses a calibrated stationary finite-state initialization. Its delay ratio relative to C rises from 3.57 to 15.55 across the same budgets. The all-12 comparison also includes approximately calibrated continuous controllers; the exact-budget subfamily provides a separate resource-controlled comparison.

A matched-risk check at \(\delta=3\) keeps source and target risks at 0.07 and 0.14 and gives a known-direction static/C delay ratio of approximately 3.42 (Appendix~\ref{app:matched-risk}). Increasing the block trigger-power target from 0.9 to 0.99 to 0.999 brings its mean delay closer to full audit. At target 0.999, the three paired gaps are approximately 0.85,2.42, and 1.80 rounds, with intervals including zero.

Figure~\ref{fig:1} plots excess mean delay over full audit against
inverse budget in the known-model SR study. For the fastest of the 12
label controllers, \(b\) times the excess stays between 33.2 and 38.5,
with endpoint log--log slope approximately 1.09; DEfull behaves
similarly. The dashed \(1/b\) line marks the unit waiting coefficient
of Theorem~\ref{thm:critical-budget} for scale only: the experiment
fixes \(A\) and measures mean delay at \(\nu=1\), not Lorden delay
along the theorem's joint limit. The factor of about 35 is the measured
controller overhead; the theoretical upper bounds show separately how
constants such as \(e_0/p_{\rm esc}\) enter.

\subsubsection{ACS Income: design and evaluation}
\label{app:income-design}

We use 2018 ACS 1-Year person records through the ACSIncome task in folktables, introduced by \citet{DHMS21}. California is the source; the other 49 states and DC are target candidates. The evaluation distribution uniformly resamples filtered records from each pool, giving unweighted prediction error as the estimand.

Of 195,665 source records, 60\% train a fixed LightGBM classifier, 20\% provide calibration subsets, and 20\% form the source evaluation pool. Calibration data are divided among error-predictor fitting, covariate geometry, risk/allocation, and trigger calibration. A source-only logistic error predictor uses features and classifier confidence. Each target is split equally between scenario selection and final evaluation. The threshold is source risk-calibration error plus 0.03:
\[
\tau\approx0.2012,\qquad
\widehat r_{\mathrm{source,eval}}\approx0.1715.
\]

The covariate trigger uses five source principal components and the source error prediction, jointly centered and whitened with a fixed small ridge. It compares the previous block's squared whitened-mean norm with a source-bootstrap threshold. Block lengths are 342,368, and 401 at budgets 1\%, 0.5\%, and 0.2\%, respectively. Held-out error and covariate-score criteria select the evaluation pools in Table~\ref{tab:3}.

\begin{table}[ht]
\centering
\small
\caption{Evaluation pools selected before detection. Mean-shift scores summarize whitened means; proxy bias is error-prediction bias. All risks are unweighted finite-pool risks.}
\label{tab:3}
\setlength{\tabcolsep}{4pt}
\renewcommand{\arraystretch}{1.15}
\begin{tabular}{@{}lrrrr@{}}
\toprule
\textbf{Pool} & \textbf{Records} & \textbf{Error risk} & \textbf{Mean shift} & \textbf{Proxy bias} \\ \midrule
CA & 39133 & 17.147\% & 0.019 & -0.208\% \\
SD & 2450 & 27.796\% & 0.682 & -9.230\% \\
WV & 4052 & 28.036\% & 0.729 & -10.215\% \\
MT & 2732 & 25.403\% & 0.752 & -6.714\% \\
DC & 2094 & 14.852\% & 0.761 & 2.091\% \\
OK & 8959 & 23.005\% & 0.489 & -5.490\% \\
\bottomrule
\end{tabular}
\end{table}

DC and SD illustrate why input change alone is insufficient: their mean-shift scores are comparable, but their error rates lie on opposite sides of the risk threshold. WV and MT provide two further harmful shifts. OK has a smaller mean-shift score and still activates the trigger strongly. To test unchanged inputs directly, we also perturb source labels in Appendix~\ref{app:income-flip}.

Every scenario and budget uses 500 paired streams, 5,000 pre-change rounds, and a maximum of 200,000 post-change rounds, with change at \(\nu=5001\). An additional source warmup of \(\lceil50/b\rceil\) rounds precedes monitored time 1. Alarm statistics then start at zero, followed by the 5,000 monitored pre-change rounds. The five acquisition policies are:

\begin{table}[H]
\centering
\small
\caption{ACS acquisition policies and detector inputs. L, C, Full, and Hybrid share their full-audit betting scale; F0 and F1 use propensity-corrected factors.}
\label{tab:4}
\setlength{\tabcolsep}{4pt}
\renewcommand{\arraystretch}{1.15}
\begin{tabularx}{\linewidth}{@{}lXX@{}}
\toprule
\textbf{Policy} & \textbf{Acquisition} & \textbf{Detector input} \\ \midrule
F0 & Uniform audit probability \(b\) & Source prediction plus IPW residual, globally safe betting scale \\
F1 & Fixed source-trained input-dependent audit probability with floor \(0.1b\) & Same augmented form, scale limited by the global floor \\
L & Full-budget label-history controller & Full-audit factors when active; 1 when sleeping \\
C & Previous-block covariate trigger & Full-audit factors in high state; 1 otherwise \\
Full & Audit every label & Full-audit factors \\
\bottomrule
\end{tabularx}
\end{table}

F1 uses \(p(x)=\operatorname{clip}(c\max\{\widehat m(x),0.02\},0.1b,1)\), with its multiplier calibrated on the source allocation subset. F0 and F1 use the augmented loss in Section~\ref{sec:model}. The L/C comparison uses identical full-audit betting components, isolating the acquisition mechanism.

Every detector mixes the eight fractions
\[
(0.01,0.025,0.05,0.1,0.2,0.35,0.6,0.9)
\]

of its safe betting scale; the scale for L, C, Full, and Hybrid is \(1/\tau\). The implemented label controller uses a Bernoulli log-likelihood-ratio score with the source-calibrated baseline and design alternative \(\tau+0.02\), replacing the alternative \(\tau\) in Section~\ref{sec:controllers}. Rounding its increments down in units of 0.01 gives \(+25,-7\), with the positive score capped at 1,000. The cap removes permanent escape, so Lemma~\ref{lem:controller} does not apply directly (Appendix~\ref{app:controller}). The classifier, error predictor, and covariate geometry remain fixed throughout evaluation.

For horizon \(H=200000\), delay is the mean of \(\min(T-5000,H)\) among streams surviving the pre-change period. Censored streams contribute \(H\). Ratios and 95\% intervals use jointly surviving paired streams and 2,000 paired bootstrap resamples; early alarms and censoring are reported separately.

\subsubsection{Harmful interstate shifts}
\label{app:income-harmful}

The rounded Income L score has zero drift at $7/32>\tau$ before
capping. Appendix~\ref{sec:rounded-uncapped} evaluates a repaired
uncapped variant at smaller exceedances; the targets below are above
that zero-drift point.

Table~\ref{tab:5} shows the CUSUM results at \(A=10^6\). Across SD, WV, and MT, the L/C ratio rises from 3.418--3.687 at a 1\% budget to 12.919--13.237 at a 0.2\% budget. C's delay varies relatively little with budget, while L's delay increases substantially. All displayed harmful-shift cells are uncensored.

\begin{table}[H]
\centering
\small
\caption{ACS harmful-shift results for CUSUM at \(A=10^6\). Delays are restricted means; the last column reports censoring counts for L and C.}
\label{tab:5}
\setlength{\tabcolsep}{4pt}
\renewcommand{\arraystretch}{1.15}
\begin{tabularx}{\linewidth}{@{}llrrYc@{}}
\toprule
\textbf{Budget} & \textbf{Target} & \textbf{L delay} & \textbf{C delay} & \textbf{L/C [95\% CI]} & \textbf{Censored L/C} \\ \midrule
1.0\% & SD & 4,260.4 & 1,155.5 & 3.687 [3.392, 3.990] & 0/0 \\
1.0\% & WV & 3,727.6 & 1,062.9 & 3.507 [3.251, 3.789] & 0/0 \\
1.0\% & MT & 6,937.4 & 2,029.7 & 3.418 [3.180, 3.669] & 0/0 \\
0.5\% & SD & 6,800.4 & 1,225.1 & 5.551 [5.122, 6.019] & 0/0 \\
0.5\% & WV & 7,263.6 & 1,130.9 & 6.423 [5.896, 7.017] & 0/0 \\
0.5\% & MT & 13,099.1 & 2,118.4 & 6.183 [5.672, 6.674] & 0/0 \\
0.2\% & SD & 16,705.8 & 1,293.1 & 12.919 [11.796, 14.136] & 0/0 \\
0.2\% & WV & 16,480.9 & 1,245.1 & 13.237 [12.154, 14.458] & 0/0 \\
0.2\% & MT & 29,039.1 & 2,230.0 & 13.022 [11.972, 14.127] & 0/0 \\
\bottomrule
\end{tabularx}
\end{table}

\begin{figure}[H]
\centering
\includegraphics[width=\linewidth]{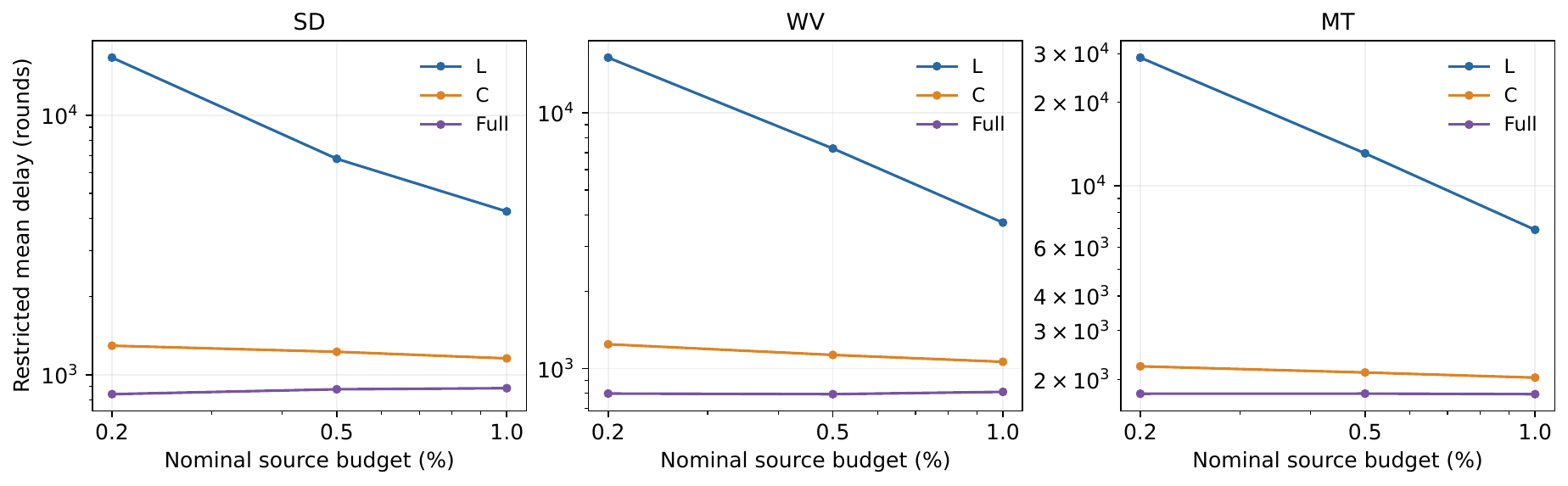}
\caption{CUSUM restricted mean delay at \(A=10^6\) on harmful ACS targets. Label-only delay increases as the nominal source budget shrinks, while covariate-triggered delay remains comparatively stable. C and its reset variant coincide in all displayed cells; only C is plotted.}
\label{fig:2}
\end{figure}

The same budget dependence appears across all three targets (Figure~\ref{fig:2}). At nominal budget 0.5\%, L's measured pre-change audit rate exceeds C's in each harmful scenario: 0.541\% versus 0.409\% for SD, 0.476\% versus 0.443\% for WV, and 0.474\% versus 0.467\% for MT (Table~\ref{tab:15}). L thus uses more labels in the measured source prefixes while detecting more slowly, making this budget comparison conservative for C. F1 performs poorly in this evaluation; the source error predictor underestimates target risk and its global propensity floor limits the betting scale. Appendix~\ref{app:five-policies} reports F0, F1, L, C, and Full together. Appendix~\ref{sec:rounded-uncapped} evaluates a nearest-rounded, uncapped label-controller variant, which has larger restricted mean delays than the original L in these nine cells under the retained calibration.

\paragraph{Shared-detector and strengthened label-history controls.}
A later supplement with its own fixed protocol adds independent uniform acquisition and
three DE-inspired sampling--sleep controls, all using the same direct-loss
CUSUM factors as C. The label controls receive the exact empirical
source risk for cost calibration. Even the post hoc minimum of their
three mean delays is 3.44--3.57,5.65--5.88 and 12.15--12.92 times C's
at budgets 1\%, 0.5\% and 0.2\%, respectively.
This is the pattern predicted by Theorem~\ref{thm:critical-budget},
under which every label-history rule pays an inverse-budget term.
Table~\ref{tab:uniform-common} reports uniform acquisition's restricted
means together with censoring percentages. Appendix~\ref{app:shared-baselines}
also gives survival curves and the full control definitions.

\subsubsection{Unchanged inputs and parallel acquisition}
\label{app:income-flip}

The input-preserving stress test resamples the source input pool and independently flips each post-change binary loss with probability 0.15. Its risk is
\[
r_{\mathrm{flip}}=0.15+0.7\,r_{\mathrm{source,eval}}\approx0.2700>\tau.
\]

C sees the unchanged input sequence and remains mostly inactive. Its restricted delay approaches the censoring horizon as the budget falls:

\begin{table}[ht]
\centering
\small
\caption{Unchanged-input label perturbations at \(A=10^6\): restricted delay and censoring for the label and covariate controllers.}
\label{tab:6}
\setlength{\tabcolsep}{4pt}
\renewcommand{\arraystretch}{1.15}
\begin{tabular}{@{}lrrr@{}}
\toprule
\textbf{Budget} & \textbf{L delay} & \textbf{C delay} & \textbf{C censored} \\ \midrule
1.0\% & 4,906.1 & 136,060.0 & 153/500 \\
0.5\% & 8,417.4 & 171,413.0 & 316/500 \\
0.2\% & 18,957.4 & 193,621.7 & 458/500 \\
\bottomrule
\end{tabular}
\end{table}

Hybrid allocates half of the 0.5\% budget to each channel. Its label-channel sleep is 1,683 rounds, and its covariate channel uses block length 368 and source-calibrated threshold approximately 21.09. The union feeds one eight-component CUSUM.

\begin{table}[ht]
\centering
\small
\caption{Hybrid at budget 0.5\% and \(A=10^6\). Brackets give paired 95\% bootstrap intervals.}
\label{tab:7}
\setlength{\tabcolsep}{4pt}
\renewcommand{\arraystretch}{1.15}
\begin{tabular}{@{}lrrrcr@{}}
\toprule
\textbf{Scenario} & \textbf{L delay} & \textbf{C delay} & \textbf{Hybrid delay} & \textbf{Censored L/C/H} & \textbf{Paired ratio [95\% CI]} \\ \midrule
SD & 6,800.4 & 1,225.1 & 1,253.0 & 0/0/0 & Hybrid/C 1.023 [1.011, 1.037] \\
WV & 7,263.6 & 1,130.9 & 1,154.1 & 0/0/0 & Hybrid/C 1.021 [1.010, 1.032] \\
MT & 13,099.1 & 2,118.4 & 2,153.6 & 0/0/0 & Hybrid/C 1.017 [1.010, 1.024] \\
Label flip & 8,417.4 & 171,413.0 & 16,150.8 & 0/316/0 & Hybrid/L 1.919 [1.723, 2.125] \\
\bottomrule
\end{tabular}
\end{table}

\begin{figure}[H]
\centering
\includegraphics[width=\linewidth]{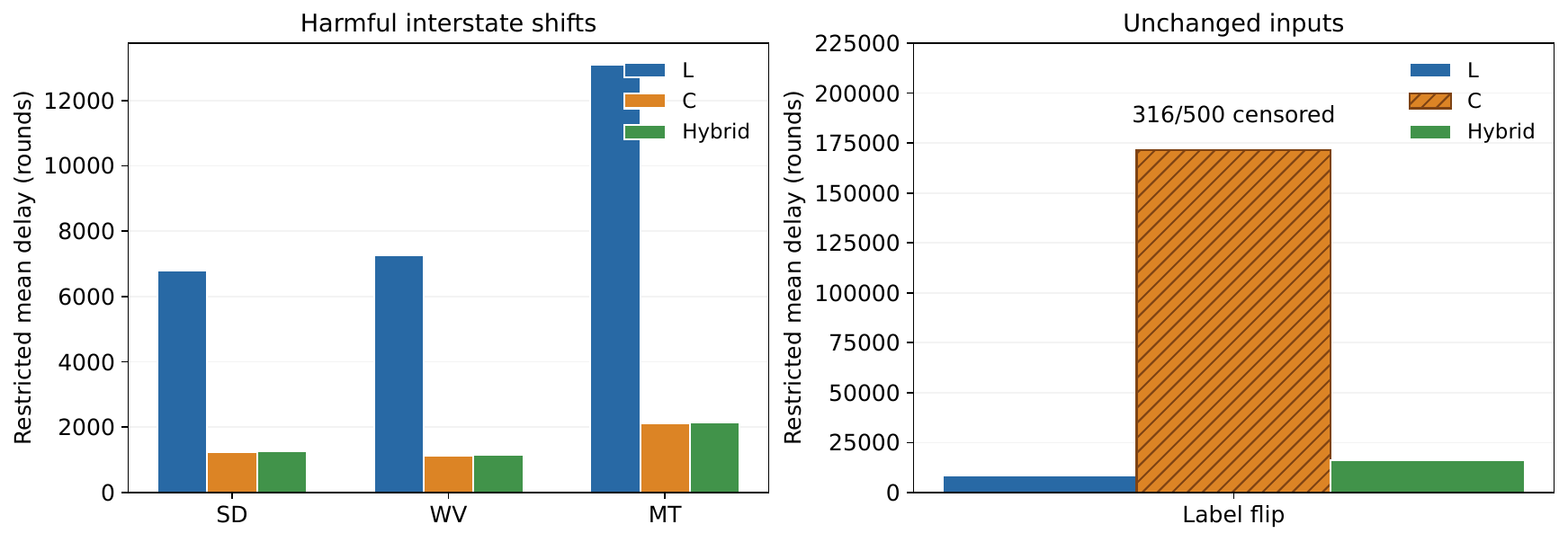}
\caption{Parallel acquisition at budget 0.5\% and \(A=10^6\). All bars are restricted mean delays with a 200,000-round post-change horizon. The hatched C bar in the right panel includes 316/500 censored streams. Hybrid stays close to C on harmful interstate shifts and restores detection under unchanged-input label perturbations.}
\label{fig:3}
\end{figure}

On the three harmful interstate shifts, the paired Hybrid/C delay ratios range from 1.017 to 1.023, with 95\% interval endpoints spanning 1.010--1.037. Under label perturbations, Hybrid's delay is 1.919 times L's, with interval [1.723,2.125]. Setting \(b_L=b/2\) doubles the waiting term in Theorem~\ref{thm:upper}(b) but not the evidence term, so a ratio somewhat below two is consistent with this allocation cost; for the capped implementation this reading is heuristic. Hybrid has no censored streams, compared with 316/500 for C (Figure~\ref{fig:3}). Appendix~\ref{app:hybrid-second} gives the second threshold and source-prefix acquisition rates.

A separate capped implementation, H$_D$, extends the allocation check
to all three budgets (Appendix~\ref{app:shared-baselines}). Under label
perturbations, its restricted delay is 1.82--1.87 times that of its
corresponding full-budget DE control. This is consistent with halving the label channel's budget, which
increases the waiting contribution. At budget 0.2\%, 4/500 H$_D$
streams remain censored at 200,000 rounds.

\subsubsection{Benign shifts and false alarms}
\label{app:income-benign}

DC reveals a cost of the covariate trigger: after a benign shift, C continues to audit almost every label. A confirmation-reset variant updates the covariate reference after accumulating evidence for risk below \(\tau\), while preserving alarm capital. Its final-100,000-round audit rates fall to 0.702\%, 0.400\%, and 0.151\%; whole-post-change rates are 2.174\%, 1.887\%, and 1.765\%, reflecting the initial adaptation cost. C and the reset variant have identical harmful-shift alarm paths in the evaluated cells. Appendix~\ref{app:benign-reset} describes the reset rule.

In no-change streams at \(A=10^4\), CUSUM produces 3/1/0 alarms for L and 3/0/0 for C among 500 runs at the three budgets. Appendix~\ref{app:nochange} contrasts this behavior with SR's repeated unit injection during skipped rounds.

To compare empirical false-alarm behavior, we also calibrate L and C toward a source mean run length of 200,000 at budget 0.5\%, using 400 calibration streams and 500 independent validation streams with horizon 2,000,000. The validation means differ:

\begin{table}[H]
\centering
\small
\caption{Source run-length calibration at budget 0.5\%: independent validation after targeting mean run length 200,000.}
\label{tab:8}
\setlength{\tabcolsep}{4pt}
\renewcommand{\arraystretch}{1.15}
\begin{tabular}{@{}lrrr@{}}
\toprule
\textbf{Policy} & \textbf{Threshold} & \textbf{Mean run length} & \textbf{Censored} \\ \midrule
L & 27.13 & 153,625.0 & 0/500 \\
C & 22.78 & 191,621.0 & 0/500 \\
\bottomrule
\end{tabular}
\end{table}

The L/C mean ratio is 0.802, with interval [0.711,0.902], so the calibration missed its target: at these thresholds C has the larger validated source mean run length. We therefore report no ARL-matched delay comparison.

\FloatBarrier
\subsection{TravelTime validation}

\label{app:traveltime}

\paragraph{Protocol, source diagnostic and target gates.}

The scientific protocol was locked before opening target evaluation halves. A source-only amendment on 2026-10-29 raises the loss-flip risk from 0.36662 (15\% flips) to $\tau+0.05=0.3944166$, using $p_0=10677/34502$ and $f=(\tau+0.05-p_0)/(1-2p_0)$. It preserves targets, risk threshold, budgets, seeds, score precision and the 200,000-round horizon. Calibration precedes target evaluation and the code/data freeze precedes the full 500-path cells. The protocol specified that an AK failure would stop the study rather than substitute RI; AK passed. The prior PublicCoverage stopping record remains unchanged.

The source classifier has training risk 0.23124 and held-out risk 0.30946, a 7.82 percentage-point gap. Screening found 46/50 targets above $\tau$ and 42/50 above $\tau+0.02$. The evaluation is conditional on this fixed CA-trained model and its source generalization gap. No model was refitted and no threshold retuned.

\begin{table}[ht]
\centering
\small
\caption{TravelTime evaluation pools, opened after the selection rule was locked. Risks are unweighted empirical losses. Exceedance is in percentage points. The shift is the norm of the saved whitened mean, not a risk statistic.}
\label{tab:tt-pools}
\setlength{\tabcolsep}{4pt}
\renewcommand{\arraystretch}{1.1}
\begin{tabular}{@{}lrrrr@{}}
\toprule
\textbf{Pool} & \textbf{Records} & \textbf{Risk} & \textbf{Risk $-\tau$} & \textbf{Input shift} \\ \midrule
CA & 34,502 & 30.946\% & $-3.496$ & Source \\
NE & 4,767 & 43.675\% & $+9.234$ & 2.121 \\
KS & 6,914 & 42.609\% & $+8.168$ & 2.026 \\
ND & 1,930 & 43.472\% & $+9.030$ & 2.256 \\
AK & 1,387 & 28.983\% & $-5.458$ & 2.246 \\
\bottomrule
\end{tabular}
\end{table}

\paragraph{Acquisition calibration and implementation.}

The five scalar chains are L, three capped DE designs and Hybrid\textquotesingle s own L. Rounded increment pairs are $(911,-448)$, $(1475,-758)$ and $(2517,-1408)$ in units $10^{-4}$. Primary L and the first DE have positive score drift at $\tau$; the other specialized designs have zero-drift risks 0.33945 and 0.35873, respectively, and do not both cover all $r>\tau$. Every chain starts active at score 0 before warmup. Sleeps use a certified upper bound $e_0^+$ on the mean excursion and $W=\lceil e_0^+(1-b_{\rm ch})/b_{\rm ch}\rceil$. The uncapped mean is about 15.36046 (certified enclosure in Appendix~\ref{app:experimental-record}); capped means are approximately 15.36044364, 8.45617553 and 5.81849752.

For capped source transition matrix $Q$, every Bellman equation is checked with high-precision decimal arithmetic at the exact empirical source risk. Residuals are below $10^{-10}$. Writing $\mu=p_0u-(1-p_0)v<0$, the comparison potential $V(s)=(s+v)/(-\mu)$ satisfies $(I-Q)V\ge 1$, converting a residual bound into an error bound. An auxiliary cap $H=200000$ is used only to enclose the uncapped mean. \par\begin{samepage}
For $\eta>0$ with $E_0e^{\eta Y}\le 1$, coupling until first reaching $H$ gives
\nopagebreak[4]
\[0\le e_{\infty}-e_H\le e^{-\eta H}\frac{H+u+v}{-\mu}.\]
\end{samepage}

The certified tail contribution is at most $1.131\times 10^{-8}$. This auxiliary numerical truncation does not cap the simulated primary L. The final relative enclosure width is $7.56\times 10^{-10}$, within the protocol's $10^{-6}$ gate. This calibration uses the exact source evaluation risk, which favors the label-history controls.

Block lengths 38, 40 and 44 are fixed before target evaluation. Each uses 400,000 source blocks for calibration and 400,000 independent blocks for validation. The six full/half-budget channel checks use one-sided Clopper--Pearson error 0.01/6; all upper limits are below channel budgets. The empirical threshold is the $(1-0.8b_{\rm ch})$ quantile, with no retuning. Hybrid\textquotesingle s union requests one label; its L score advances only on its own requests. Other C labels do not interrupt L sleep. Acquisition continues after alarms.

Survival summaries stop at the 200,000-round horizon; we do not extrapolate censored tails. SeedSequence keys separate warmup records, monitored records, Uniform coins and flips. Budgets share monitored records and Uniform coins despite different warmup lengths. No-change and flip share records. Short Python/numba checks agree exactly; Full alarm paths agree across budgets, and 144 KM areas equal restricted means.

\begin{table}[ht]
\centering
\small
\caption{TravelTime calibrated sleep rounds. L and Hybrid L are uncapped. Every DE keeps physical cap 10 (integer cap 100000); Hybrid L uses half the nominal budget.}
\label{tab:tt-sleeps}
\setlength{\tabcolsep}{4pt}
\renewcommand{\arraystretch}{1.1}
\begin{tabular}{@{}lrrr@{}}
\toprule
\textbf{Label channel} & \textbf{$b=1\%$} & \textbf{$b=0.5\%$} & \textbf{$b=0.2\%$} \\ \midrule
L & 1,521 & 3,057 & 7,665 \\
DE $_\tau$ & 1,521 & 3,057 & 7,665 \\
DE $_{\tau+.02}$ & 838 & 1,683 & 4,220 \\
DE $_{\tau+.06}$ & 577 & 1,158 & 2,904 \\
Hybrid L & 3,057 & 6,129 & 15,346 \\
\bottomrule
\end{tabular}
\end{table}

{
\small
\setlength{\tabcolsep}{6pt}
\renewcommand{\arraystretch}{1.1}
\begin{longtable}{@{}lrrr@{}}
\caption{All eight TravelTime policies on harmful and input-preserving scenarios at $A=10^6$. Entries are restricted means with censored paths out of 500 in parentheses. All cells have 500 pre-change survivors.}
\label{tab:tt-all}\\
\toprule
\textbf{Policy} & \textbf{$b=1\%$} & \textbf{$b=0.5\%$} & \textbf{$b=0.2\%$}\\ \midrule
\endfirsthead
\multicolumn{4}{c}{\tablename~\thetable{} (continued)}\\
\toprule
\textbf{Policy} & \textbf{$b=1\%$} & \textbf{$b=0.5\%$} & \textbf{$b=0.2\%$}\\ \midrule
\endhead
\midrule
\multicolumn{4}{r}{\emph{Continued on next page}}\\
\endfoot
\bottomrule
\endlastfoot
\multicolumn{4}{l}{\textbf{NE}} \\
L (uncapped) & 5,622.9 (0) & 9,883.4 (0) & 22,962.6 (0) \\
C & 837.9 (0) & 846.5 (0) & 857.6 (0) \\
DE $_\tau$ & 5,622.9 (0) & 9,883.4 (0) & 22,962.6 (0) \\
DE $_{\tau+.02}$ & 4,194.0 (0) & 8,606.2 (0) & 21,992.5 (0) \\
DE $_{\tau+.06}$ & 3,714.6 (0) & 6,419.7 (0) & 16,681.7 (0) \\
Uniform & 77,356.2 (1) & 142,961.4 (121) & 193,591.5 (446) \\
Hybrid & 840.7 (0) & 847.7 (0) & 858.8 (0) \\
Full & 802.7 (0) & 802.7 (0) & 802.7 (0) \\
\multicolumn{4}{l}{\textbf{KS}} \\
L (uncapped) & 6,618.7 (0) & 11,056.2 (0) & 25,858.6 (0) \\
C & 1,028.5 (0) & 1,041.8 (0) & 1,051.1 (0) \\
DE $_\tau$ & 6,618.7 (0) & 11,056.2 (0) & 25,858.6 (0) \\
DE $_{\tau+.02}$ & 5,126.1 (0) & 8,795.8 (0) & 25,997.1 (0) \\
DE $_{\tau+.06}$ & 4,739.4 (0) & 7,799.5 (0) & 19,289.1 (0) \\
Uniform & 101,390.4 (8) & 169,603.2 (243) & 198,461.0 (483) \\
Hybrid & 1,030.6 (0) & 1,043.4 (0) & 1,053.2 (0) \\
Full & 985.1 (0) & 985.1 (0) & 985.1 (0) \\
\multicolumn{4}{l}{\textbf{ND}} \\
L (uncapped) & 5,895.5 (0) & 9,362.1 (0) & 23,687.5 (1) \\
C & 904.3 (0) & 914.9 (0) & 930.1 (0) \\
DE $_\tau$ & 5,895.5 (0) & 9,362.1 (0) & 23,687.5 (1) \\
DE $_{\tau+.02}$ & 4,328.9 (0) & 8,233.8 (0) & 21,983.8 (0) \\
DE $_{\tau+.06}$ & 4,096.0 (0) & 6,959.4 (0) & 17,889.8 (0) \\
Uniform & 81,842.7 (0) & 153,784.4 (156) & 196,432.3 (467) \\
Hybrid & 906.9 (0) & 917.0 (0) & 930.7 (0) \\
Full & 867.0 (0) & 867.0 (0) & 867.0 (0) \\
\multicolumn{4}{l}{\textbf{Unchanged-input flip}} \\
L (uncapped) & 11,051.5 (0) & 18,211.0 (0) & 42,501.8 (2) \\
C & 191,466.1 (411) & 199,508.4 (493) & 200,000.0 (500) \\
DE $_\tau$ & 11,051.5 (0) & 18,211.0 (0) & 42,501.8 (2) \\
DE $_{\tau+.02}$ & 9,119.8 (0) & 15,812.7 (0) & 38,943.5 (0) \\
DE $_{\tau+.06}$ & 9,453.4 (0) & 15,403.3 (0) & 38,346.3 (0) \\
Uniform & 185,319.2 (364) & 198,528.2 (485) & 200,000.0 (500) \\
Hybrid & 18,633.9 (0) & 36,372.5 (0) & 70,236.3 (49) \\
Full & 2,699.6 (0) & 2,699.6 (0) & 2,699.6 (0) \\
\end{longtable}
}

\begin{table}[ht]
\centering
\small
\caption{TravelTime Uniform censoring and KM medians at the same 200,000-round horizon. NR denotes a median not reached within the horizon.}
\label{tab:tt-uniform}
\setlength{\tabcolsep}{4pt}
\renewcommand{\arraystretch}{1.1}
\begin{tabular}{@{}llrrr@{}}
\toprule
\textbf{Scenario} & \textbf{$b$} & \textbf{RMST} & \textbf{Censored} & \textbf{KM median} \\ \midrule
NE & 1\% & 77,356.2 & 1/500 (0.2\%) & 73,782 \\
NE & 0.5\% & 142,961.4 & 121/500 (24.2\%) & 145,247 \\
NE & 0.2\% & 193,591.5 & 446/500 (89.2\%) & NR \\
KS & 1\% & 101,390.4 & 8/500 (1.6\%) & 100,649 \\
KS & 0.5\% & 169,603.2 & 243/500 (48.6\%) & 198,754 \\
KS & 0.2\% & 198,461.0 & 483/500 (96.6\%) & NR \\
ND & 1\% & 81,842.7 & 0/500 (0.0\%) & 78,547 \\
ND & 0.5\% & 153,784.4 & 156/500 (31.2\%) & 164,312 \\
ND & 0.2\% & 196,432.3 & 467/500 (93.4\%) & NR \\
Flip & 1\% & 185,319.2 & 364/500 (72.8\%) & NR \\
Flip & 0.5\% & 198,528.2 & 485/500 (97.0\%) & NR \\
Flip & 0.2\% & 200,000.0 & 500/500 (100.0\%) & NR \\
\bottomrule
\end{tabular}
\end{table}

\paragraph{Safety and cost boundaries.}

No monitored pre-change alarms occur in any cell. For AK all eight policies have 0/500 alarms at each budget. For no change, Full has 1/500 alarms, shared exactly across budgets, and all other policies have 0/500. Under the benign AK shift, C and Hybrid label 99.976\%--99.984\% of post-change rounds, and 100\% in the final 100,000 rounds. Nothing bounds this post-change cost, and this task has no reset variant.

\begin{table}[ht]
\centering
\small
\caption{AK post-change label counts through 200,000 rounds and Hybrid minus C, with paired 95\% intervals. Near-zero differences do not imply low absolute labeling cost.}
\label{tab:tt-safe-cost}
\setlength{\tabcolsep}{4pt}
\renewcommand{\arraystretch}{1.1}
\begin{tabular}{@{}lrrr@{}}
\toprule
\textbf{$b$} & \textbf{C labels} & \textbf{Hybrid labels} & \textbf{Difference [95\% CI]} \\ \midrule
1\% & 199,968.4 & 199,968.1 & -0.320 [-0.640, -0.064] \\
0.5\% & 199,960.2 & 199,960.1 & -0.160 [-0.400,0.000] \\
0.2\% & 199,952.4 & 199,950.9 & -1.496 [-2.202, -0.880] \\
\bottomrule
\end{tabular}
\end{table}

\begin{figure}[H]
\centering
\includegraphics[width=\linewidth]{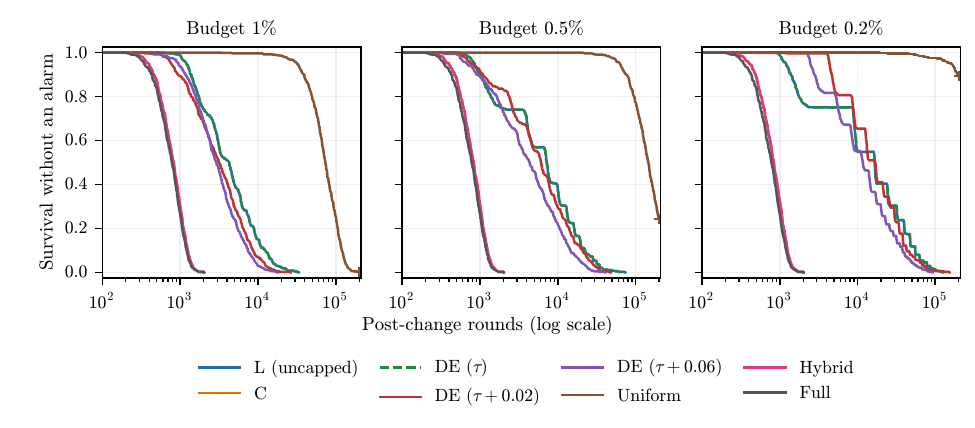}
\caption{TravelTime no-alarm survival for the first locked harmful target NE. All eight fixed policies use the same CUSUM; Uniform retains substantial administrative censoring at low budgets.}
\label{fig:tt-harmful-km}
\end{figure}

\begin{figure}[H]
\centering
\includegraphics[width=\linewidth]{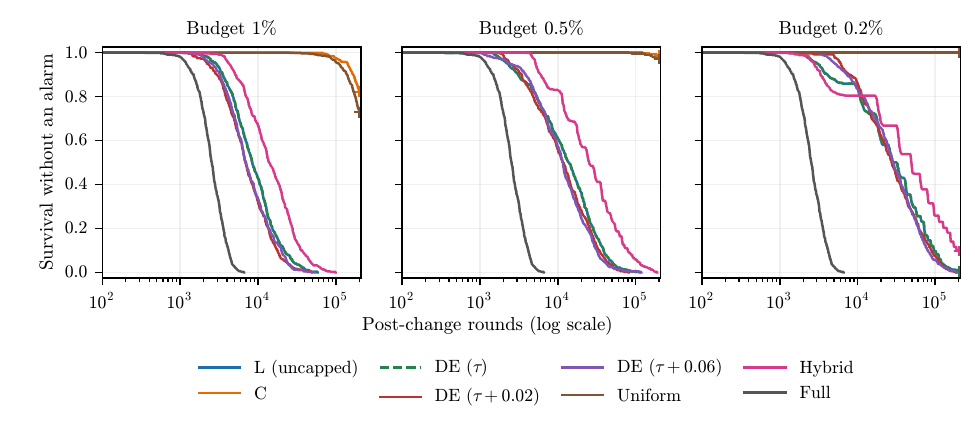}
\caption{TravelTime input-preserving flip survival. C loses power as its unchanged-input trigger becomes rare; L and Hybrid restore detection, with 49/500 Hybrid paths still censored at the lowest budget.}
\label{fig:tt-flip-km}
\end{figure}

\FloatBarrier
\begin{table}[ht]
\centering
\small
\caption{Income, all three harmful targets at $A=10^6$. Ranges run over SD, WV and MT. Best DE is the post hoc minimum over the three designs of Table~\ref{tab:de-shared}. Uniform uses the common detector and a 200,000-round censoring horizon.}
\label{tab:income-summary}
\setlength{\tabcolsep}{4pt}
\renewcommand{\arraystretch}{1.1}
\begin{tabular}{@{}lrrr@{}}
\toprule
\textbf{Budget} & \textbf{C delay} & \textbf{Best DE/C} & \textbf{Uniform censored} \\ \midrule
1\% & 1,063--2,030 & 3.44--3.57 & 0.0--38.0\% \\
0.5\% & 1,131--2,118 & 5.65--5.88 & 26.0--86.4\% \\
0.2\% & 1,245--2,230 & 12.15--12.92 & 92.0--99.4\% \\
\bottomrule
\end{tabular}
\end{table}
\begin{table}[H]
\centering
\small
\caption{TravelTime at $A=10^6$: restricted mean post-change rounds through 200,000 rounds, with 500 paired paths per cell. Best DE is the post hoc minimum over the three fixed designs. Cens. gives L/C/Hybrid counts; all paths survive pre-change time. Uniform and every fixed DE, KM and cost result are in Appendix~\ref{app:traveltime}.}
\label{tab:traveltime-summary}
\setlength{\tabcolsep}{4pt}
\renewcommand{\arraystretch}{1.1}
\begin{tabularx}{\linewidth}{@{}llYYYYYY@{}}
\toprule
\textbf{$b$} & \textbf{Target} & \textbf{L} & \textbf{C} & \textbf{Best DE} & \textbf{Hybrid} & \textbf{Full} & \textbf{Cens.} \\ \midrule
1\% & NE & 5,623 & 838 & 3,715 & 841 & 803 & 0/0/0 \\
1\% & KS & 6,619 & 1,029 & 4,739 & 1,031 & 985 & 0/0/0 \\
1\% & ND & 5,896 & 904 & 4,096 & 907 & 867 & 0/0/0 \\
1\% & Flip & 11,052 & 191,466 & 9,120 & 18,634 & 2,700 & 0/411/0 \\
0.5\% & NE & 9,883 & 847 & 6,420 & 848 & 803 & 0/0/0 \\
0.5\% & KS & 11,056 & 1,042 & 7,799 & 1,043 & 985 & 0/0/0 \\
0.5\% & ND & 9,362 & 915 & 6,959 & 917 & 867 & 0/0/0 \\
0.5\% & Flip & 18,211 & 199,508 & 15,403 & 36,372 & 2,700 & 0/493/0 \\
0.2\% & NE & 22,963 & 858 & 16,682 & 859 & 803 & 0/0/0 \\
0.2\% & KS & 25,859 & 1,051 & 19,289 & 1,053 & 985 & 0/0/0 \\
0.2\% & ND & 23,688 & 930 & 17,890 & 931 & 867 & 1/0/0 \\
0.2\% & Flip & 42,502 & 200,000 & 38,346 & 70,236 & 2,700 & 2/500/49 \\
\bottomrule
\end{tabularx}
\end{table}
\FloatBarrier
\section{Reproducibility and AI assistance}\label{app:repro}

\FloatBarrier
\subsection{Experimental record}\label{app:experimental-record}

The TravelTime study in Appendix~\ref{app:traveltime} comprises 18 frozen scenario--budget cells with 500 paired paths and eight policies each (72{,}000 policy--path records) at $A=10^6$. A source-only amendment, made before any detection run, fixes the flip risk at $\tau+0.05$. The archive contains the original and amended protocols, numerical certificates, fixed target gates, the code/data freeze, all 18 raw files, paired summaries, cost diagnostics and Kaplan--Meier checks. Primary L is uncapped; the physical DE cap is $10$ at unit $10^{-4}$. Code hashes and seed substreams are recorded. All 144 Kaplan--Meier areas equal the corresponding restricted means, and Full paths agree across budgets.

The accompanying artifacts contain records of data acquisition and filtering; disjoint training, calibration and evaluation splits; fitted source models; query policies; detector configurations; raw trajectories; early alarms; censoring indicators; and paired summary tables. The review artifact contains a SHA256 manifest of all package files and the recorded software versions; per-run command logs and the original freeze manifests are retained in the authors' development archive. Independent trajectory replays and summary recomputations cover 180 ACS main cells, 504 extension cells and 24 Hybrid cells. The empirical estimand is uniform finite-pool resampling, and bootstrap intervals condition on the fixed upstream pipeline.

Each experimental stage fixed its protocol before observing its detection outcomes. The CUSUM evaluations, confirmation resets, label perturbations and Hybrid were added after the initial protocol; in particular, Hybrid was introduced after the covariate-only controller's weakness under label perturbations was observed. A source-drift calibration issue in a proposed label controller was repaired before any ACS detection run, and the full calibration record is included with the artifacts.

\paragraph{Protocol stages.}
The table records, for each stage, what was known when its decisions
were fixed:
\begin{center}
\small
\setlength{\tabcolsep}{4pt}
\renewcommand{\arraystretch}{1.08}
\begin{tabularx}{\linewidth}{@{}p{0.20\linewidth}p{0.28\linewidth}X@{}}
\toprule
\textbf{Stage} & \textbf{Information available} & \textbf{Timing of the decisions} \\ \midrule
Calibration repair & Source calibration & Completed before any ACS detection run. \\
CUSUM, resets and flips & Additions to the initial protocol & Each stage fixed before its own detection outcomes. \\
Hybrid & C's weakness under label flips & Fallback introduced, then its evaluation stage fixed. \\
Rounded and shared controls & Original ACS results & Separate extensions fixed before their new detection runs (Appendices~\ref{sec:rounded-uncapped}, \ref{app:shared-baselines}). \\
PublicCoverage & Feasibility screening & Stopped at protocol gates before detection. \\
TravelTime & Source risk and fixed screening & Source-only flip amendment before any detection run; the AK stopping gate was retained. \\
\bottomrule
\end{tabularx}
\end{center}

For continuous synthetic controllers, the uncapped nominal-excursion calculation gives $EU=\kappa_0 E\tau_-$ with $\kappa_0=\mathrm{kl}(r_0,r_d)$; with sleep $\lceil U/\mu\rceil$, the ideal renewal rate lies between $\mu/(\kappa_0+2\mu)$ and $\mu/(\kappa_0+\mu)$. The all-12 comparison uses these approximately calibrated settings, and DEfull supplies the separately calibrated exact-budget family. The matched-risk checkpoint is reported in Appendix~\ref{app:matched-risk}.

The shared-detector study in Appendix~\ref{app:shared-baselines} has 18 paired scenario--budget cells, 500 streams per cell, eight acquisition rules and two thresholds. Its frozen protocol, 144{,}000 policy--threshold records, replay checks and a second summary recalculation are included. The survival curves reuse these streams. The PublicCoverage attempt and its stopping checks are documented in Appendix~\ref{app:coverage-feasibility}.

The nearest-rounded, uncapped study in Appendix~\ref{sec:rounded-uncapped} was first implemented in C++ and then regenerated in the original numba framework with the two additional controllers; all 14 cells and 500 streams per cell match exactly in the recorded alarm and label-count fields. The regeneration protocol, environment, source/input hashes, per-cell raw outputs and comparisons are included. This study was added after the initial protocol, and its rounded score uses the adjustment-coefficient argument given in that appendix. Realized source-pool budget overages are reported alongside the results. The regeneration used Python 3.14.3, numba 0.68.0, NumPy 2.5.3 and SciPy 1.18.1.

\paragraph{Reported configuration constants.}
For reproducibility, the configuration values reported in this manuscript are
\(\delta=2.2225546893\) (Appendix~\ref{app:original-synthetic}), \(\tau=0.2012022486\) and
\(\widehat r_{\mathrm{source,eval}}=0.1714665372\) (Appendix~\ref{app:income-design}),
covariate threshold \(21.092832\) (Appendix~\ref{app:income-flip}), run-length thresholds
\(27.13228\) (L) and \(22.78044\) (C) (Table~\ref{tab:8}), and logistic
coefficients \(-4.2961009736\) and \(0.0797286115\) (Appendix~\ref{app:matched-risk}).
The constructed label-perturbation risk is derived as
\(r_{\mathrm{flip}}=0.15+0.7\widehat r_{\mathrm{source,eval}}\approx0.2700265760\).
For TravelTime, the certified enclosure of the uncapped mean source
excursion is \([15.360461242238593,15.360461253838490]\)
(Appendix~\ref{app:traveltime}).

\FloatBarrier
\subsection{Use of AI tools}\label{app:ai-assistance}

The authors used large language models to translate research literature, identify research gaps, develop proof ideas (including for Theorem~\ref{thm:additive} and Lemma~\ref{lem:audit}), draft an initial outline of the manuscript, and polish the language. The authors verified every step of all proofs in the paper and its appendices, including the request-epoch, local-calibration and boundary arguments and Appendices~\ref{app:audit}--\ref{app:fixed-audit}.

The authors wrote the experimental code, with AI tools suggesting improvements; the code for calibrating label budgets and detection thresholds was generated with AI tools. The authors independently reproduced all reported experiments, in addition to the automated trajectory replays and summary recomputations described in Appendix~\ref{app:experimental-record}.

The authors take full responsibility for the accuracy, originality, attribution and integrity of this paper, including AI-assisted material.

\section{Detailed related-work and model comparisons}\label{app:related-details}

Sequential risk monitoring distinguishes changes in model performance from changes in the input distribution. \citet{PR22} track deployed-model risk, including delayed labels. \citet{TVNN25} use testing by betting to detect bounded conditional-risk violations under unknown shifts, with time-uniform false-alarm control for each fixed decision threshold. Their protocol reveals a label and loss after each decision; we study adaptive acquisition under a nominal prefix budget and Lorden delay with full-null finite-window validity. \citet{AB24} study label-free monitoring with a learned error proxy, and \citet{schirmer2025tta} extend label-free risk monitoring to models updated by test-time adaptation; PPRM \citep{ZCYS26} combines synthetic and true labels to bound running risk, and WATCH \citep{PHLS25} uses weighted-conformal martingales.

\citet{GZZ22} use feature signals and requested expert labels to
monitor deployment performance. Their loss combines average query cost
with accuracy-estimation or threshold-decision error under gradual
distribution drift. \citet{YWP18} use unlabeled screening followed by
requested-label verification for concept drift. \citet{LYEOM26} use drift evidence to size uniform window audits for
an anytime-valid upper risk certificate. The certificate gates
predictions and fallback under a finite-horizon labeling budget. These
works establish the acquisition idea. Our results concern a common
source-prefix allowance, false-alarm validity over the full risk null,
and Lorden delay at the critical budget. A wide upper certificate can
activate fallback even at a safe risk, so that gating guarantee does
not supply the local false-alarm condition studied here.

Active labeling connects this problem to prediction-powered inference. \citet{XKM24} combine query policies, predictors, and anytime-valid risk control; \citet{SW26} analyze active sequential prediction-powered mean estimation and the role of input-dependent queries. Our augmented loss uses this established correction principle. The constructive controllers instead choose an on/off state before the current input, which allows zero acquisition probability without evaluating an inverse-probability correction. The resulting alarm uses the SR/CUSUM e-detector framework of \citet{SRR23}. General growth-optimal e-variable and KL-projection foundations come from Safe Testing \citep{GdHK23}, numeraire theory \citep{LRR25}, and constraint-generated testing \citep{LRR26}. Appendix~\ref{app:fixed-audit} specializes these growth-optimal constructions to fixed audit policies.

\citet{PK08} formulate sleep/wake observation control as a Bayesian
Markov decision problem with costs for false alarms, delay and sensor
observations. Our sampling/sleep structure belongs to this observation-control
tradition; what is specific here is the source-prefix budget and the
worst-history Lorden criterion.

The closest observation-control precedents are the DE-CuSum procedure of \citet{BV12}, its composite-post-change extension \citep{BV14}, and the observation scheduling studied by \citet{RJS17}. They already establish acquisition controlled by accumulated evidence and fixed-budget first-order efficiency. \citet{LB25} consider experiments with different costs and information, obtaining delay within a threshold-independent constant of the highest-quality experiment. Their robust extension \citep{LB26} handles unknown post-change experiment laws through least-favorable members. Our label-history controller is a baseline from this line of work. \citet{tsopelakos2023anomaly} study sequential identification of anomalous data sources under a sampling constraint expressed as the ratio of the expected total number of samples to the expected stopping time. Their objective is terminal identification rather than detecting an unknown change time, and their stopping-time resource constraint differs from our common nominal allowance on every deterministic prefix.

Conditional stopped-trajectory KL decompositions are established tools
in bandit change detection \citep[Appendix~A, Lemma~1]{GSL21}; our
information step applies this principle with a predictably switched
safe reference. Resource-dependent leading-order overheads have also been established
in multi-stream sampling control \citep{XMM21,XM23}. Our contribution
concerns a different setting: a composite risk null, a common
source-prefix allowance, and a policy-uniform additive converse. Nonstationary
information-growth converses \citep{LTV22} describe delay through an
inverse information clock under growth and tail assumptions, which an
expected budget does not supply uniformly over acquisition rules
(Appendix~\ref{app:clock-example} gives an example).

Adaptive sparse-sampling obstructions and vanishing sampling fractions
also arise in transient detection \citep[Theorems 4--5]{CT17}, under a
transient model, timing uncertainty and stopped sampling criteria. Our
result concerns a persistent change and an expected-prefix budget.
Technically, our converse rests on the joint use of a budget-driven
waiting comparison and a safe random-start reference.

For requests chosen before the current input, the problem embeds in
the two-experiment framework $E_0=X$, $E_1=(X,L)$, so free inputs alone
do not separate it from controlled sensing. Table~\ref{tab:9}
summarizes the remaining differences.

\Needspace{22\baselineskip}
\begin{table}[H]
\centering
\small
\caption{Observation and resource assumptions in cost-constrained experiment design and in the present model.}
\label{tab:9}
\setlength{\tabcolsep}{4pt}
\renewcommand{\arraystretch}{1.15}
\begin{tabularx}{\linewidth}{@{}lZZ@{}}
\toprule
\textbf{Feature} & \textbf{Cost-constrained design} & \textbf{Present model} \\ \midrule
Null & Specified nominal experiment laws & Conditional risk at most $\tau$, including safe input changes \\
Observation & $E_0=X$, $E_1=(X,L)$ embeds past-based requests & Also permits current-input-dependent requests \\
Resource comparison & Pre-change experiment fractions & Source-prefix allowance; one prefix in a stated range suffices at the critical budget \\
Relevant conclusion & Constructive delay comparisons & Policy-uniform additive converse and critical-budget separation \\
\bottomrule
\end{tabularx}
\end{table}

\FloatBarrier

\paragraph{Nominal safety.}

The nominal source is safe: \(r_{\mathrm{src}}:=E_{P_0}L\le\tau\). Since \(P_0\) is IID, \(E_{P_0}[L_t\mid\mathcal F_{t-1}]=r_{\mathrm{src}}\), so \(P_0\in\mathcal H_0\). This supplies the safe pre-change segment of the reference processes in Appendix~\ref{app:additive}; their post-change segments are safe by construction. The theoretical label-history controller additionally uses a design bound \(r_{\mathrm{src}}\le r_0<\tau\) (Section~\ref{sec:controllers}).

\end{document}